\documentclass{article}

    \PassOptionsToPackage{numbers, compress}{natbib}

    \usepackage[final]{neurips_2024}

\usepackage{amsmath,amsfonts,bm}
\usepackage{mathrsfs}

\def\eqref#1{equation~\ref{#1}}

\def\1{\bm{1}}

\DeclareMathAlphabet{\mathsfit}{\encodingdefault}{\sfdefault}{m}{sl}
\SetMathAlphabet{\mathsfit}{bold}{\encodingdefault}{\sfdefault}{bx}{n}

\newcommand{\norm}[1]{\left\lVert#1\right\rVert}
\newcommand{\abs}[1]{\left|#1\right|}

\usepackage{tikz}

\newcommand{\inner}[2]{\langle#1,#2\rangle}

\DeclareMathOperator*{\argmax}{arg\,max}
\DeclareMathOperator*{\argmin}{arg\,min}

\newenvironment{myproof}[2]{\paragraph{\textit{Proof of {#1} {#2}. }}}{\hfill$\blacksquare$}

\usepackage{xcolor}

\definecolor{brightmaroon}{rgb}{0.76, 0.13, 0.28}
\definecolor{brown(web)}{rgb}{0.65, 0.16, 0.16}

\newcommand{\pagoda}{{\color{brightmaroon}{PaGoDA}}}

\def\eqref#1{(\ref{#1})}

\def\eqref#1{(\ref{#1})}

\def\eqref#1{(\ref{#1})}

\usepackage{svg}
\usepackage[utf8]{inputenc} 
\usepackage[T1]{fontenc}    
\usepackage{hyperref}       
\usepackage{url}            
\usepackage{booktabs}       
\usepackage{amsfonts}       
\usepackage{nicefrac}       
\usepackage{microtype}      
\usepackage{xcolor, color, colortbl}         
\usepackage{tikz}
\usepackage[export]{adjustbox}
\usepackage{mathtools}

\usepackage{amsmath}
\usepackage{amssymb}
\usepackage{mathtools}
\usepackage{amsthm}
\usepackage{bm}
\usepackage{pifont}
\usepackage{xcolor}
\usepackage{calc}
\usepackage{enumerate}
\usepackage[shortlabels]{enumitem}
\usepackage[normalem]{ulem}
\usepackage{makecell}

\usepackage{mathtools}
\usepackage{fontawesome}
\usepackage{subcaption}
\usepackage{graphics}
\usepackage{graphicx}
\usepackage{caption}
\usepackage{wrapfig}
\usepackage{dblfloatfix}
\usepackage{tabu}
\usepackage{adjustbox}
\usepackage{algorithm}
\usepackage{algpseudocode}
\usepackage{multirow}
\usepackage{threeparttable}
\usepackage{arydshln}
\usepackage{minitoc}
\usepackage{kotex}
\usepackage[most]{tcolorbox} 

\newtheorem{theorem}{Theorem}[section]
\newtheorem*{theorem*}{Theorem}

\newtheorem{lemma}[theorem]{Lemma}

\theoremstyle{definition}
\newtheorem{definition}{Definition}[section]

\newtheorem{assumpA}{Assumption}
\newtheorem{assumpB}{Assumption}
\newtheorem{assumpC}{Assumption}

\theoremstyle{remark}

\newtheorem*{remark*}{Remark}
\allowdisplaybreaks

\newcommand*\diff{\mathop{}\!\mathrm{d}}
\newsavebox{\leftbox}
\newsavebox{\rightbox}
\definecolor{Gray}{gray}{0.9}

\newcommand{\cc}[1]{\cellcolor{gray!#1}}
\hypersetup{colorlinks}

\makeatletter
\def\adl@drawiv#1#2#3{%
        \hskip.5\tabcolsep
        \xleaders#3{#2.5\@tempdimb #1{1}#2.5\@tempdimb}%
                #2\z@ plus1fil minus1fil\relax
        \hskip.5\tabcolsep}
\newcommand{\cdashlinelr}[1]{%
  \noalign{\vskip\aboverulesep
           \global\let\@dashdrawstore\adl@draw
           \global\let\adl@draw\adl@drawiv}
  \cdashline{#1}
  \noalign{\global\let\adl@draw\@dashdrawstore
           \vskip\belowrulesep}}
\makeatother

\newlength\myheight
\newlength\mydepth
\settototalheight\myheight{Xygp}
\title{\pagoda \,{\includegraphics[height=3\myheight]{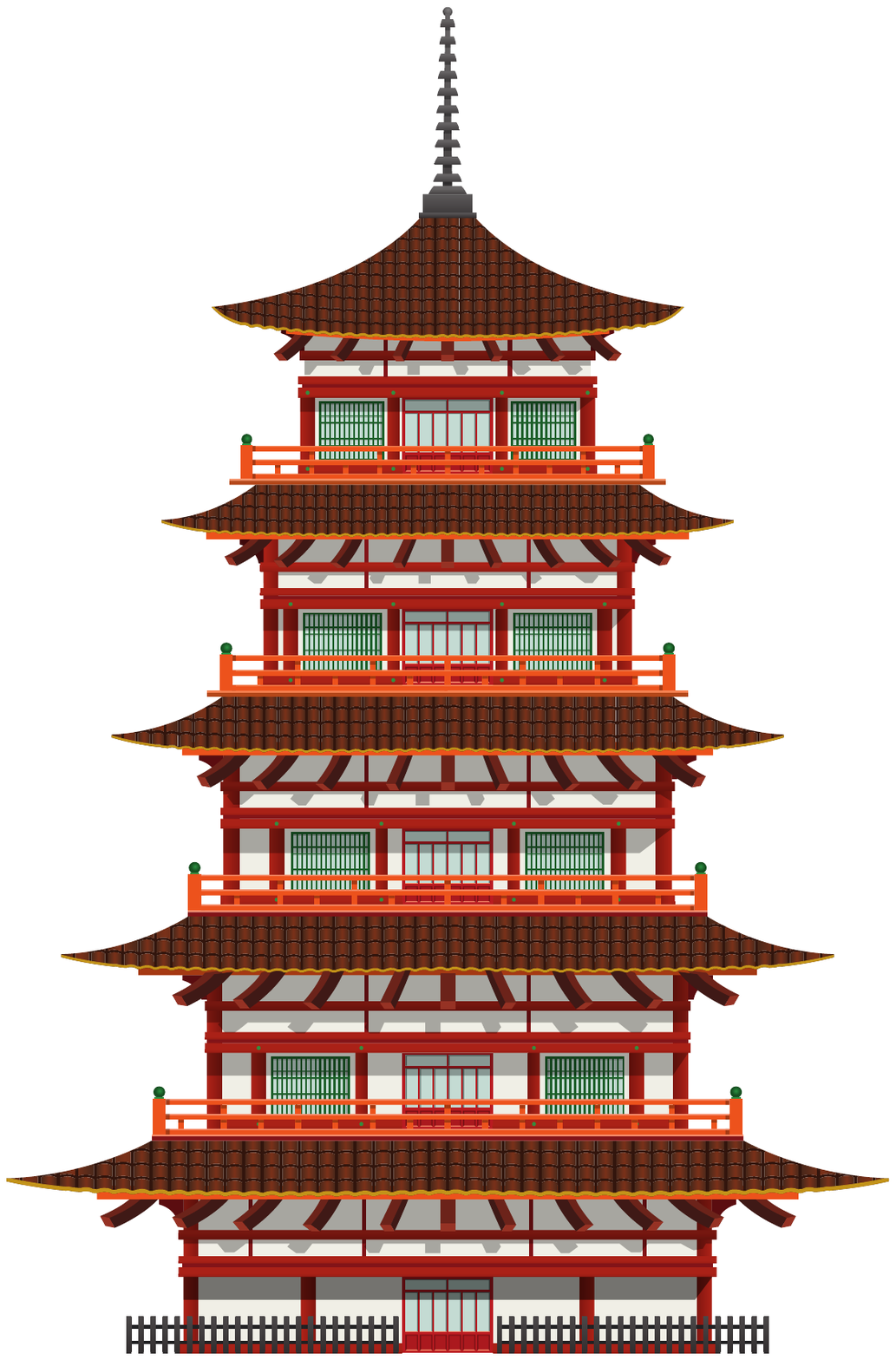}}: Progressive Growing of a One-Step Generator from a Low-Resolution Diffusion Teacher
}

\author{%
  Dongjun Kim\thanks{Equal contribution}~~\thanks{This work was partially done during an internship at Sony AI.} \\
  Stanford University \\
  CA, USA \\
  \texttt{dongjun@stanford.edu} \\
  \And
  Chieh-Hsin Lai$^{*}$ \\
  Sony AI \\
  Tokyo, Japan \\
  \texttt{chieh-hsin.lai@sony.com} \\
  \AND
  Wei-Hsiang Liao \\
  Sony AI \\
  \And
  Yuhta Takida \\
  Sony AI \\
  \And
  Naoki Murata \\
  Sony AI \\
  \And
  Toshimitsu Uesaka \\
  Sony AI \\
  \AND
  Yuki Mitsufuji \\
  Sony AI, Sony Group Corporation \\
  \And
  Stefano Ermon \\
  Stanford University
}

\begin{document}

\maketitle

\begin{abstract}
The diffusion model performs remarkable in generating high-dimensional content but is computationally intensive, especially during training. We propose \textbf{P}rogressive \textbf{G}rowing \textbf{o}f \textbf{D}iffusion \textbf{A}utoencoder (\textbf{\pagoda}), a novel pipeline that reduces the training costs through three stages: training diffusion on downsampled data, distilling the pretrained diffusion, and progressive super-resolution. With the proposed pipeline, PaGoDA achieves a $64\times$ reduced cost in training its diffusion model on $8\times$ downsampled data; while at the inference, with the single-step, it performs state-of-the-art on ImageNet across all resolutions from $64\times64$ to $512\times512$, and text-to-image. PaGoDA's pipeline can be applied directly in the latent space, adding compression alongside the pre-trained autoencoder in Latent Diffusion Models (e.g., Stable Diffusion). The code is available at \url{https://github.com/sony/pagoda}.
\end{abstract}

\section{Introduction}

Diffusion Models (DM)~\citep{ho2020denoising,song2020score}, which generate content through gradual denoising, have recently achieved high fidelity in high-dimensional generation~\citep{rombach2022high,dhariwal2021diffusion}. While slow sampling has been improved by distilling trained DMs into single-step generators~\citep{song2023consistency,luhman2021knowledge,kim2023consistency}, DMs remain computationally intensive, especially at high resolutions, requiring substantial data and GPU resources, thereby limiting large-scale training to a few organizations~\citep{esser2024scaling,flux2024}. This highlights the need for a more efficient pipeline to reduce both training and inference costs while maintaining the quality.

To address these challenges, we present \textbf{P}rogressive \textbf{G}rowing \textbf{o}f \textbf{D}iffusion \textbf{A}utoencoder (\textbf{\pagoda}), a novel pipeline that significantly reduces costs while achieving competitive quality with one-step sampling. PaGoDA is built on a simple yet effective idea: while diffusion distillation~\citep{luhman2021knowledge} is typically treated as a final stage of the whole pipeline, we explore to have one more stage for the super-resolution after diffusion distillation. This approach led us to design PaGoDA with three distinct stages as below.
\begin{tcolorbox}[colback=gray!15, colframe=gray!20, boxrule=0pt, arc=0pt, outer arc=0pt]
\textbf{\pagoda's Proposed Training Pipeline}
\begin{enumerate}[label=\textbf{Stage \arabic*.}, topsep=5pt, leftmargin=1.8cm] 
    \item (Pretraining) Train a DM on downsampled data. 
    \item (Distillation) Distill the trained DM with DDIM inversion to a one-step generator.
    \item (Super-Resolution) Progressively expand the generator for resolution upsampling.
\end{enumerate}
\end{tcolorbox}

\begin{figure*}[t]
	\centering
 \includegraphics[width=\linewidth]{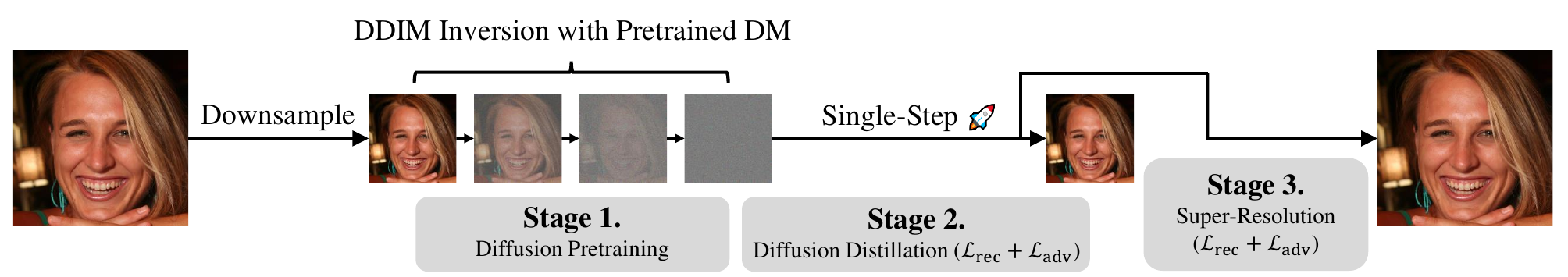}
 \vskip 0.1in
 \caption{Pipeline overview. PaGoDA deterministically encodes with downsampling followed by DDIM inversion, and constructs its decoder in a progressively growing manner.}
    \label{fig:PaGoDA_architecture}
    \vskip -0.1in
\end{figure*}

By adding Stage 3 for the super-resolution after the distillation phase, our approach gains a key advantage: training DM on a low-dimensional, downsampled space rather than directly in the desired high-dimensional space. This dimensional reduction substantially lowers the computational demands of diffusion pretraining by orders of magnitude. For example, an $8\times 8$ downsampling rate reduces the training computation by a factor of $64\times$. Moreover, the computational costs for the distillation and super-resolution stages are relatively minimal compared to the initial diffusion pretraining, making our pipeline highly efficient in terms of overall computation.

\begin{figure*}[ht]
\vskip -0.1in
\centering
\includegraphics[width=0.85\linewidth]{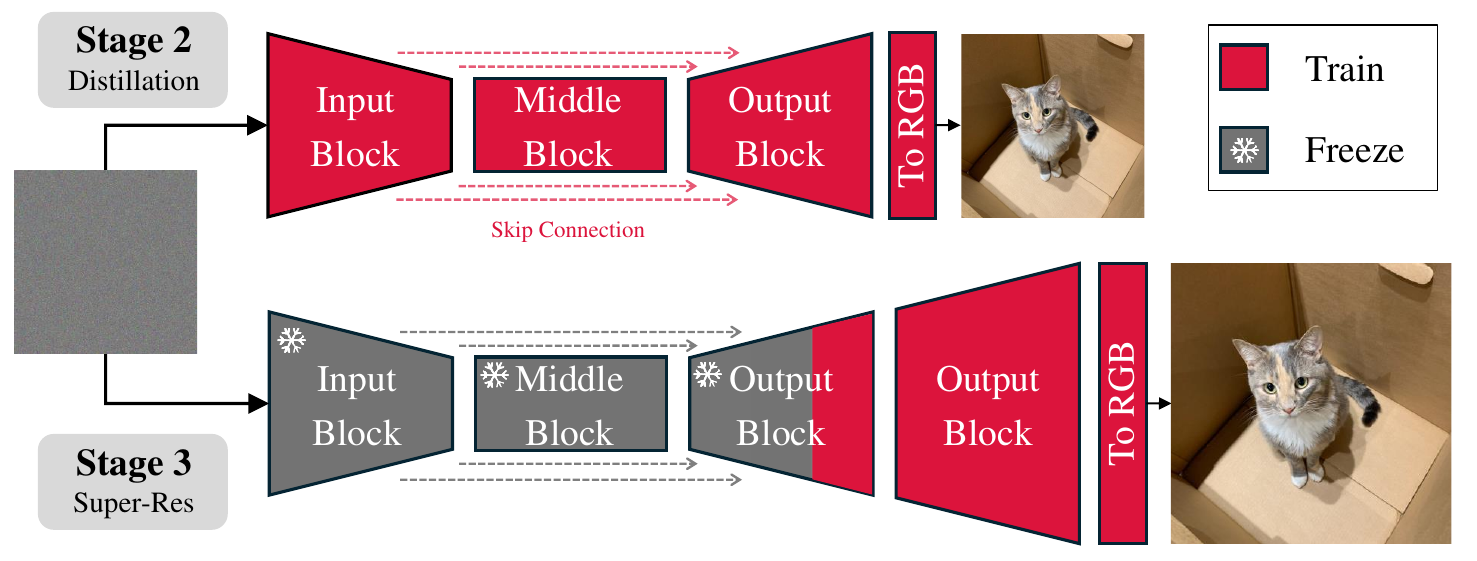}
 \caption{(Top) At Stage 2, PaGoDA learns the one-step generator at a base resolution. (Down) At Stage 3, PaGoDA progressively learns for super-resolution by adding additional network blocks.}
 \label{fig:PaGoDA_super_resolution}
 \vskip -0.1in
 \end{figure*}

Figure~\ref{fig:PaGoDA_architecture} provides an overview of our pipeline. We begin with DM trained at base resolution, and generate a dataset of base-resolution data-latent pairs $(\mathbf{x},\mathbf{z})$, where $\mathbf{x}$ is real data and $\mathbf{z}$ is the latent representation of $\mathbf{x}$, obtained by DDIM inversion~\citep{song2020denoising}. In Stage 2, we train a decoder to map $\mathbf{z}$ back to $\mathbf{x}$, completing the diffusion distillation~\citep{luhman2021knowledge}. In Stage 3, we add ResNet blocks~\citep{he2016deep} to enhance sample resolution and progressively train these newly added upscaling networks, as visualized in Figure~\ref{fig:PaGoDA_super_resolution}. The novel use of DDIM inversion in the distillation process, first introduced in PaGoDA, enables the decoder to be trained with the high-frequency signal from the real data at Stage 3. This integration of DDIM inversion establishes strong connections across stages, creating a cohesive and unified framework.

In our experiments, we employed the progressively growing generator to upsample from the pre-trained diffusion model's $64\times64$ resolution to generate samples at $512\times512$ resolution. Notably, PaGoDA achieved state-of-the-art (SOTA) Fréchet Inception Distances (FID)~\citep{heusel2017gans} on ImageNet across all resolutions from $64\times64$ to $512\times512$. Additionally, we demonstrated PaGoDA's effectiveness in addressing inverse problems and facilitating controllable generation. However, PaGoDA's potential extends beyond its current application. As PaGoDA being a dimensional reduction technique that operates independently of Latent Diffusion Models (LDM)~\citep{rombach2022high}, PaGoDA could be directly applied into the latent space as-is, offering the possibility of further gain on training computes. We leave this exploration as a promising avenue for future research.

\section{Preliminary}\label{sec:preliminary}

DM~\citep{ho2020denoising} samples from the data distribution $p_{\text{\text{data}}}$ through an iterative denoising process, beginning from a Gaussian prior distribution $p_{\text{\text{prior}}}$. This denoising process attempts to reverse~\citep{song2020score} a forward diffusion process. If the forward process is defined by $\diff\mathbf{x}_{t}=\sqrt{2t}\diff\mathbf{w}_{t}$~\citep{karras2022elucidating}, the deterministic counterpart of the denoising (generation) process, known as the probability flow ordinary differential equation (PF-ODE)~\citep{song2020score}, or DDIM~\citep{song2020denoising}, is expressed as 
\begin{align*}
    \frac{\diff \mathbf{x}_t}{\diff t} =  -t \nabla \log p_t (\mathbf{x}_t) \approx - t s_{{\bm{\phi}}_0}(\mathbf{x}_t,t) ,
\end{align*}
where $s_{{\bm{\phi}}_0}(\mathbf{x}_t,t)$ is a neural approximation of $ \nabla \log p_t (\mathbf{x}_t)$. Consequently, (deterministic) sample generation from DM is equivalent to solving the PF-ODE (or DDIM) along the trajectory, formally,
\begin{align*}
    \mathbf{x}_{0}^{\text{DDIM}}(\mathbf{x}_{T})=\mathbf{x}_{T}-\int_{T}^{0}ts_{\bm{\phi}_{0}}(\mathbf{x}_{t},t)\diff t, \quad\mathbf{x}_T \sim p_{\text{\text{prior}}}.
\end{align*}
Modern solvers of the PF-ODE~\cite{song2020denoising,lu2022dpm} have significantly accelerated sampling speed, reducing the required network evaluations from hundreds to tens. To further speed up sampling, DMs are distilled with a student model~\citep{luhman2021knowledge} $G_{\bm{\theta}}:\mathbb{R}^{d}\rightarrow\mathbb{R}^{d}$ to map from $\mathbf{x}_{T}$ to $\mathbf{x}_{0}^{\text{DDIM}}(\mathbf{x}_{T})$ by minimizing
\begin{align}\label{eq:dstl}
    \mathcal{L}_{\text{dstl}}(G_{\bm{\theta}})=\mathbb{E}_{p_{\text{prior}}(\mathbf{x}_{T})}\Big[\big\Vert 
\mathbf{x}_{0}^{\text{DDIM}}(\mathbf{x}_{T})-G_{\bm{\theta}}(\mathbf{x}_{T}) \big\Vert_{2}^{2}\Big].
\end{align}
We call this DDIM-based approach as the \emph{noise-to-data} distillation.

\section{Progressive Growing of Diffusion Autoencoder}\label{sec:PaGoDA}

\subsection{Stage 1: Diffusion Models Trained on Downsampled Data}

Training DMs for high-dimensional data generation is primarily feasible for a limited number of well-resourced organizations, largely due to two factors: access to large-scale datasets and substantial computational resources. This centralization of model development underscores the urgent need to democratize access by significantly reducing resource demands during diffusion training. While several strategies~\citep{pernias2023wurstchen,rombach2022high} have been proposed, our approach, PaGoDA, introduces a paradigm shift by training the DM at a downsampled resolution in Stage 1, rather than at the original full resolution. For instance, training on a $d$-dimensional downsampled resolution requires approximately $4^{n}$ times less computational budget compared to training in the full $4^{n}d$-dimensional space. In practical terms, when $n=3$, this translates to training in an 8$\times$8 downsampled space, effectively reducing training costs by a factor of 64$\times$, thus making large-scale diffusion training more accessible to a broader range of researchers.

Although this paper does not extend PaGoDA's application to the LDM such as SD, training on a (say) 4$\times$4 downsampled latent space could theoretically reduce the computational cost by 16$\times$ compared to full-resolution latent training, further emphasizing PaGoDA's potential for widespread adoption. In the case of generating 1024$\times$1024 images, PaGoDA requires training the diffusion model at only 32$\times$32 resolution, with Stage 3 subsequently upscaling it to the full 128$\times$128 latent space of conventional approaches~\citep{esser2024scaling,flux2024}. This progressive approach illustrates PaGoDA’s effectiveness in maintaining model quality while lowering the barriers to high-resolution diffusion training.

\subsection{Stage 2: Diffusion Distillation on Downsampled Data with DDIM Inversion}\label{sec:pagoda_objective}

After pretraining DM on the downsampled space, PaGoDA distills DM to a one-step generator. For distillation, PaGoDA introduces a new loss specifically designed for later usage in super-resolution at Stage 3. In particular, we propose the reconstruction loss (compare it with $\mathcal{L}_{\text{dstl}}$ in Eq.~\ref{eq:dstl} of Section~\ref{sec:preliminary})
\begin{align}\label{eq:recon_loss}
    \mathcal{L}_{\text{rec}}(G_{\bm{\theta}}):=\mathbb{E}_{p_{\text{data}}(\mathbf{x}_{0})}\Big[\big\Vert \mathbf{x}_{0}-G_{\bm{\theta}}\big(\mathbf{x}_{T}^{\text{DDIM}^{-1}}(\mathbf{x}_{0})\big) \big\Vert_{2}^{2}\Big],
\end{align}
where $\mathbf{x}_{T}^{\text{DDIM}^{-1}}(\mathbf{x}_{0})$ is now the latent representation of $\mathbf{x}_{0}$, obtained from DDIM inversion, not from DDIM, i.e., the solution at time $T$ of the PF-ODE starting from $\mathbf{x}_{0}$ in time forward, defined by
\begin{align*}
    \mathbf{x}_{T}^{\text{DDIM}^{-1}}(\mathbf{x}_{0}):=\mathbf{x}_{0}-\int_{0}^{T}ts_{\bm{\phi}_{0}}(\mathbf{x}_{t},t)\diff t.
\end{align*}

Distillation using $\mathcal{L}_{\text{rec}}$ maps latent representations to real data, following a \textit{data-to-noise} distillation approach. While this method has the potential to improve real data alignment compared to the traditional noise-to-data approach in Eq.~\ref{eq:dstl}, we observe a decline in generation quality over iterations. This decline stems from the prior hole problem~\citep{aneja2021contrastive}, where the generator's input, $\mathbf{x}_{T}^{\text{DDIM}^{-1}}$, derived from limited real data, fails to cover the entire prior manifold, leaving certain regions unexplored.

A straightforward strategy like early stopping could alleviate this issue, but it restricts the use of Exponential Moving Average (EMA) in Stage 2. To fundamentally resolve this problem, we propose a solution that maintains generation quality even during prolonged training. In Section~\ref{sec:guarantee}, we provide optimal and stability analysis of Stage 2, guaranteeing that our proposal is stable across training iterations. The key challenge is effectively covering the prior manifold, which we address by introducing an auxiliary adversarial loss, as defined below:
\begin{align}\label{eq:adv_loss}
    \mathcal{L}_{\text{adv}}(G_{\bm{\theta}},D_{\bm{\psi}}):=\mathbb{E}_{p_{\text{\text{data}}}(\mathbf{x})}\big[\log{D_{\bm{\psi}}(\mathbf{x})}\big]+\mathbb{E}_{p_{\text{\text{prior}}}(\mathbf{z})}\Big[\log{\Big(1-D_{\bm{\psi}}\big(G_{\bm{\theta}}(\mathbf{z})\big)\Big)}\Big],
\end{align}
Here, $D_{\bm{\psi}}$ is a discriminator that classifies the real and fake samples by maximizing the adversarial loss, and $p_{\text{\text{prior}}}(\mathbf{z})$ is the prior distribution.

The second term in $\mathcal{L}_{\text{adv}}$, which involves $G_{\bm{\theta}}(\mathbf{z})$ with $\mathbf{z}$ sampled from the prior, ensures that the decoder is exposed to the entire support of the prior distribution during training. Overall, we train PaGoDA with the mini-max optimization of the following combined objective:
\begin{align}\label{loss:pagoda}
    \min_{G_{\bm{\theta}}}\max_{D_{\bm{\psi}}}\mathcal{L}_{\text{PaGoDA}}(G_{\bm{\theta}},D_{\bm{\psi}}):=\min_{G_{\bm{\theta}}}\Big[\mathcal{L}_{\text{\text{rec}}}(G_{\bm{\theta}})+ \lambda\max_{D_{\bm{\psi}}}\mathcal{L}_{\text{adv}}(G_{\bm{\theta}},D_{\bm{\psi}})\Big].
\end{align}
While PaGoDA incorporates the adversarial loss, the reconstruction loss simultaneously guides the decoder to accurately reconstruct the entire training data. This combination allows the adversarial loss to address underrepresented regions in the prior distribution effectively without compromising sample diversity. In our ImageNet experiments, we found that updating the reconstruction loss with as little as $1\%$ of the data-latent pairs did not affect sample quality and diversity. Exploring the impact of varying the number of data-latent pairs is left as a future work.

\subsection{Stage 3: Progressively Growing Decoder for Super-Resolution}

\begin{figure*}[t]
\centering
\begin{subfigure}{0.48\linewidth}
		\centering
		\includegraphics[width=\linewidth]{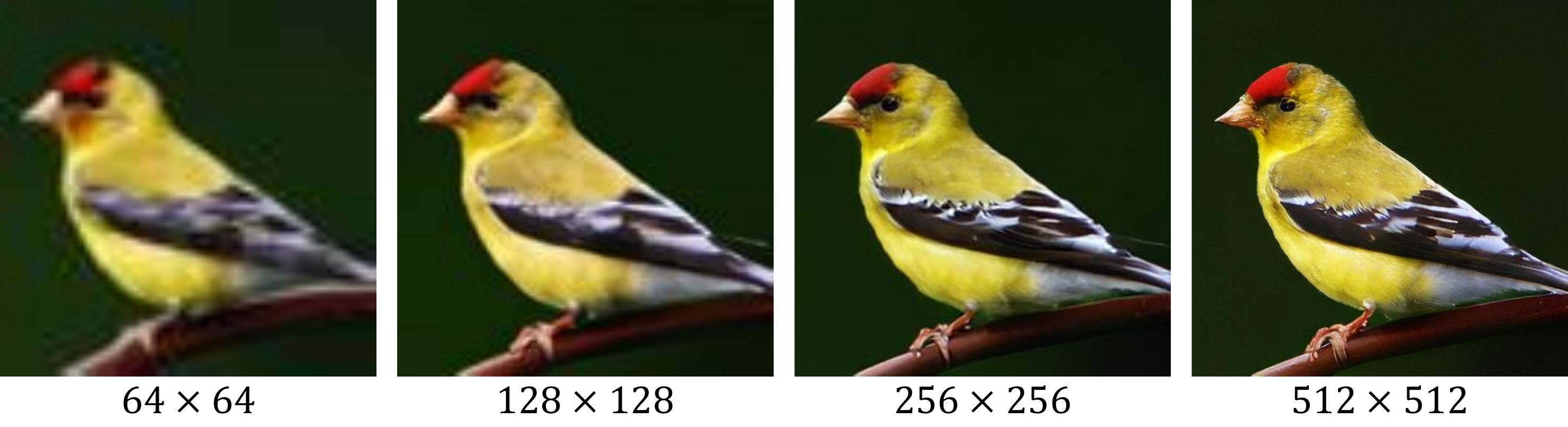}
  \vskip -0.05in
		\subcaption{Resolution Jump w/ Recon Loss}
	\end{subfigure}	
     \begin{subfigure}{0.48\linewidth}
		\centering
		\includegraphics[width=\linewidth]{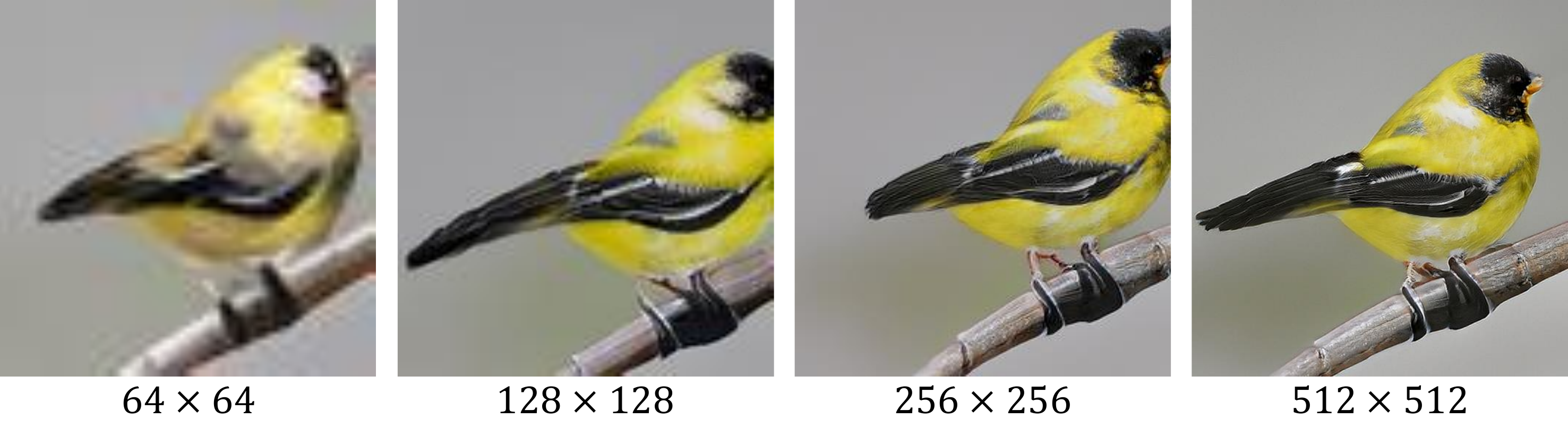}
  \vskip -0.05in
		\subcaption{Resolution Jump w/o Recon Loss}
	\end{subfigure}
 \caption{Effect of the reconstruction loss in Stage 3. Without the reconstruction loss, the object moves at each resolution jump.}
	\label{fig:upsampling}
\end{figure*}

Stage 3 trains the super-resolution to generate higher-dimensional data from the downsampled resolution learned in the previous stages. As illustrated in Figure~\ref{fig:PaGoDA_super_resolution}, the resolution jump from $\mathbb{R}^{d}$ to $\mathbb{R}^{4^{n}d}$ is achieved by freezing most parameters of the distilled model from Stage 2 and training only the final layers, which is augmented with an additional upscaler network (of ResNet blocks~\citep{zagoruyko2016wide}). In other words, within the base-resolution U-Net~\citep{ronneberger2015u}, we freeze its input, middle, and output blocks except for the last few layers (previously highest resolution block) during Stage 3 training. Consequently, the unfrozen latter part of the network is trained for super-resolution. We suggest to progressively increasing the resolution by a factor of 2$\times$, though larger jumps by factors of 4$\times$ or 8$\times$ yield comparable performance.

Additionally, the last layer typically converts multi-channel (usually 128 or 256 channels) features to 3-channel RGB output. However, to minimize information loss, we retain these features and pass them directly to the next output block without converting them to 3 channels. This architectural choice, along with progressive training, is heavily inspired by Progressive Growing GAN~\citep{karras2017progressive}.

In Stage 3, the reconstruction loss from Stage 2 is adapted as
\begin{align*}
\mathbb{E}_{p_{\text{data}}(\mathbf{x}_{\text{high}})}\Big[\big\Vert\mathbf{x}_{\text{high}}-G_{\bm{\theta}}(\mathbf{x}_{T}^{\text{DDIM}^{-1}}(\mathbf{x}_{0}))\big\Vert_{2}^{2}\Big],
\end{align*}
where $\mathbf{x}_{0}\in\mathbb{R}^{d}$ is the downsampled counterpart of $\mathbf{x}_{\text{high}}\in\mathbb{R}^{4^{n}d}$. The adversarial loss in this stage is
\begin{align*}
\mathbb{E}_{p_{\text{data}}(\mathbf{x}_{\text{high}})}\big[\log{D_{\bm{\psi}}(\mathbf{x}_{\text{high}})}\big]+\mathbb{E}_{p_{\text{prior}}(\mathbf{z})}\Big[\log{\Big( 1-D_{\bm{\psi}}\big(G_{\bm{\theta}}(\mathbf{z})\big) \Big)}\Big].
\end{align*}
Overall, both the reconstruction and adversarial losses are combined to guide training.

\begin{wrapfigure}{r}{0.35\textwidth}
\centering
  \vskip -0.2in
		\includegraphics[width=\linewidth]{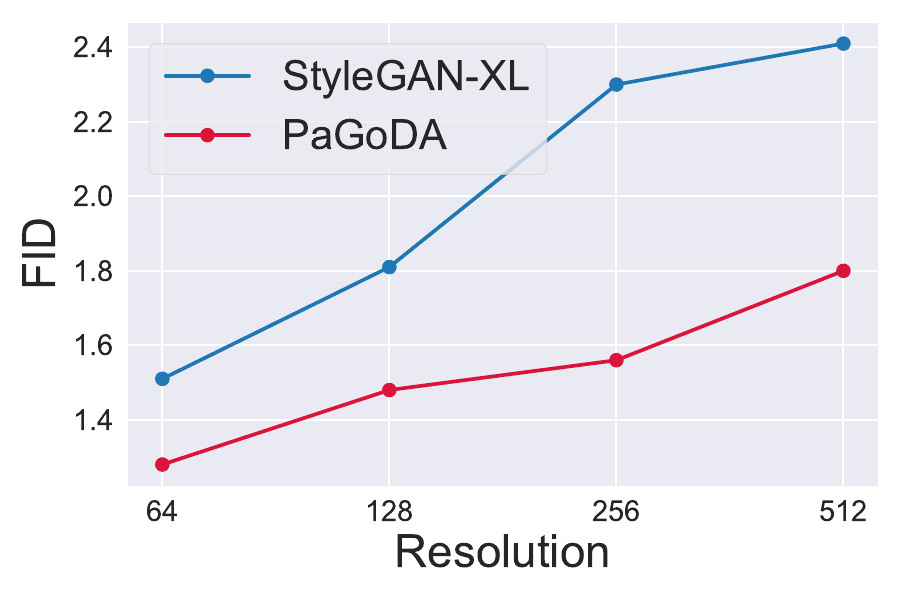}
  \vskip -0.07in
		\caption{The adversarial loss makes PaGoDA competitive with GAN-based super-resolution models in Stage 3.}
  \label{fig:effect_to_adversarial}
  \vskip -0.2in
  \end{wrapfigure}
Stage 3 employs two key mechanisms to effectively capture high-frequency details while maintaining training stability. First, the reconstruction loss is applied directly to high-dimensional real data, which was not feasible with earlier noise-to-data distillation methods with Eq.~\ref{eq:dstl}. As illustrated in Figure~\ref{fig:upsampling}, $\mathcal{L}_{\text{rec}}$ stabilizes the upscaling process by preventing objects from shifting across resolutions, allowing the added neural network to focus solely on upsampling. Second, the adversarial loss operates directly in high-dimensional space, enabled by the one-step generator trained in Stage 2. This generator is critical; without it, adversarial training in Stage 3 would be infeasible. As shown in Figure~\ref{fig:effect_to_adversarial} tested on ImageNet, the adversarial loss is pivotal for achieving effective upscaling performance.

\subsection{Optimality Guarantee and Training Stability of PaGoDA Pipeline}\label{sec:guarantee}

\begin{wrapfigure}{r}{0.35\textwidth}
\centering
\vskip -0.23in
		\includegraphics[width=\linewidth]{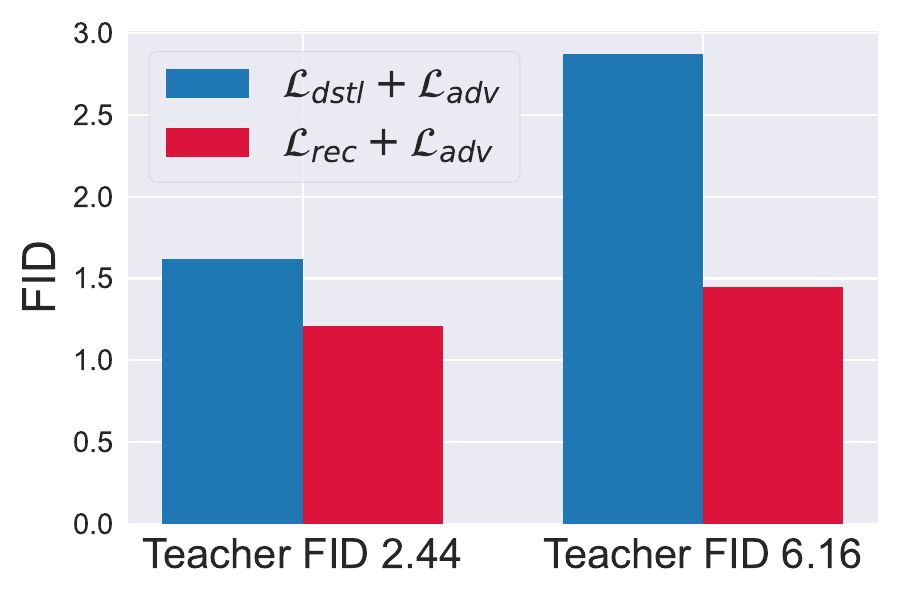}
  \vskip -0.07in
		\caption{Comparison of $\mathcal{L}_{\text{dstl}}$ and $\mathcal{L}_{\text{rec}}$, both combined with $\mathcal{L}_{\text{adv}}$, using identical hyperparameters. $\mathcal{L}_{\text{rec}}$ shows the robust performance, also supported by Theorem~\ref{thm:optimality}.}
  \label{fig:teacher_ablation}
  \vskip -0.2in
\end{wrapfigure}
When using the conventional $\mathcal{L}_{\text{dstl}}$ for distillation, the optimal student becomes $G_{\bm{\theta}^{*}}(\mathbf{x}_{T})=\mathbf{x}_{0}^{\text{DDIM}}(\mathbf{x}_{T})$, meaning that the student's samples replicate those of DDIM. As a result, the student's performance is heavily dependent on the teacher's performance. Consequently, the student's generative distribution may diverge from the real data distribution, even when $\mathcal{L}_{\text{dstl}}$ is combined with adversarial loss. In contrast, by using the DDIM inversion-based reconstruction loss proposed in Stage 2, we mathematically prove in Theorem~\ref{thm:optimality} that the optimal student's generative distribution aligns with the real data distribution. As visualized in Figure~\ref{fig:teacher_ablation}, our PaGoDA Stage 2 (red) achieves robust performance even with a weaker teacher, unlike traditional noise-to-data distillation loss $\mathcal{L}_{\text{dst}}$ of Eq.~\ref{eq:dstl}, which struggles despite the use of adversarial loss.
\begin{theorem}\label{thm:optimality} Let $\lambda>0$. Suppose $D^{*}(G)\in\argmax_{D}\mathcal{L}_{\text{adv}}(G,D)$. If both PaGoDA's reconstruction loss and adversarial loss share a common minimizer $G^*$, then $p_{G^*}(\mathbf{x})=p_{\text{\text{data}}}(\mathbf{x})$. Here,  $p_{G^*}$ is the generative distribution learned by optimizing Eq.~\eqref{loss:pagoda}.
\end{theorem}

Additionally, Theorem~\ref{thm:stability} shows that PaGoDA's training is stable with the help of reconstruction loss, even with adversarial training. We empirically observe that PaGoDA can be trained effectively without many of the techniques typically used to stabilize GANs~\cite{karras2019style,mescheder2018training}.
\begin{theorem}\label{thm:stability}\textup{[Informal]} Let $E$ be a fixed deterministic encoder. Suppose that at the generator's equilibria $G^*$ of Eq.~\eqref{loss:pagoda}, $p_{G^*}(\mathbf{x})=p_{\text{\text{data}}}(\mathbf{x})$, and  $\mathbf{x}=G^*(E(\mathbf{x}))$. Then, under conditions similar to those found in the  stability literature for improving GAN~\cite{nagarajan2017gradient,mescheder2018training}, training with Eq.~\eqref{loss:pagoda} is stable (gradient descent locally converges to its equilibria).
\end{theorem}
We refer to Theorems~\ref{th:optimality} and \ref{th:stable_pagoda} for rigorous and extended versions of Theorems~\ref{thm:optimality} and \ref{thm:stability}, respectively. All proofs can be found in Appendix~\ref{sec:theory_proof}.

\section{PaGoDA with Classifier-Free Guidance}\label{sec:classifier_free}

In this section, we integrate Classifier-Free Guidance (CFG)~\citep{ho2021classifier,dhariwal2021diffusion} into PaGoDA for Text-to-Any generation, with a focus on Text-2-Image. Incorporating CFG alters the sample distribution, necessitating adjustments to the loss functions for Stages 2 and 3. Since previous GAN literature~\citep{kang2023scaling, sauer2023stylegan, sauer2023adversarial, sauer2024fast} has not addressed CFG integration, we introduce the classifier-free guided adversarial loss to accommodate this adaptation.

CFG guides the denoising process by adjusting the conditional score gradient $\nabla\log{p_{t}(\mathbf{x}_{t}\vert\mathbf{c})}$ into a guided score $\nabla\log{p_{t}(\mathbf{x}_{t}\vert\mathbf{c})}+(\omega-1)\nabla\log{p(\mathbf{c}\vert\mathbf{x}_{t})}$. This adjustment leads our distillation learning target from $p_{\text{data}}(\mathbf{x}\vert\mathbf{c})$ to $p_{\text{data}}(\mathbf{x}\vert\mathbf{c},\omega)$, defined by
\begin{align*}
    p_{\text{\text{data}}}(\mathbf{x}\vert\mathbf{c},\omega)\propto p_{\text{\text{data}}}(\mathbf{x}\vert\mathbf{c})^{\omega}p_{\text{\text{data}}}(\mathbf{x})^{1-\omega},
\end{align*}
reflecting the influence of guidance strength $\omega$.

\subsection{Classifier-Free Guided Adversarial Loss}

To describe the classifier-free adversarial loss, we first consider the loss:
\begin{align*}
    \mathcal{L}_{\text{adv}}^{\mathbf{c},\omega}(G_{\bm{\theta}},D_{\bm{\psi}}):= \mathbb{E}_{p_{\text{\text{data}}}(\mathbf{x}\vert\mathbf{c},\omega)}\Big[\log{D_{\bm{\psi}}(\mathbf{x},\mathbf{c},\omega)}\Big] + \mathbb{E}_{p_{G_{\bm{\theta}}}(\mathbf{x}\vert\mathbf{c},\omega)}\Big[ \log{\Big(1-D_{\bm{\psi}}(\mathbf{x},\mathbf{c},\omega)\Big)} \Big],
\end{align*}
where now both generator and discriminator incorporates $\omega$ as an additional condition~\citep{meng2023distillation}, see Eq.~\eqref{eq:adv_loss} for the comparison. From the standard GAN argument~\citep{goodfellow2014generative}, this GAN loss guarantees the optimal generator to match to the data distribution, i.e., $p_{G^{*}}(\mathbf{x}\vert\mathbf{c},\omega)=p_{\text{\text{data}}}(\mathbf{x}\vert\mathbf{c},\omega)$. Hence, the classifier-free adversarial loss could be defined by
\begin{align*}
    &\mathcal{L}_{\text{adv}}^{\text{CFG}}(G_{\bm{\theta}},D_{\bm{\psi}}):=\mathbb{E}_{p_{\text{data}}(\mathbf{c})\pi(\omega)}\big[ \mathcal{L}_{\text{adv}}^{\mathbf{c},\omega}(G_{\bm{\theta}},D_{\bm{\psi}})\big]\\
    &\quad=\mathbb{E}_{p_{\text{data}}(\mathbf{c})\pi(\omega)p_{\text{data}}(\mathbf{x}\vert\mathbf{c},\omega)}\Big[\log{D_{\bm{\psi}}\big(\mathbf{x},\mathbf{c},\omega\big)}\Big]+\mathbb{E}_{p_{\text{\text{data}}}(\mathbf{c})\pi(\omega)p_{G_{\bm{\theta}}}(\mathbf{x}\vert\mathbf{c},\omega)}\Big[\log{\big(1-D_{\bm{\psi}}\big(\mathbf{x},\mathbf{c},\omega\big)\big)}\Big].
\end{align*}

A key challenge with $\mathcal{L}_{\text{adv}}^{\mathbf{c},\omega}$ is that sampling from $p_{\text{\text{data}}}(\mathbf{x}\vert\mathbf{c},\omega)$ is generally infeasible, making it difficult to compute the first term of $\mathcal{L}_{\text{adv}}^{\text{CFG}}$. To address this issue, we leverage the Bayes formula
\begin{align*}
    p_{\text{\text{data}}}(\mathbf{c})\pi(\omega)p_{\text{\text{data}}}(\mathbf{x}\vert\mathbf{c},\omega)=p_{\text{\text{data}}}(\mathbf{x},\mathbf{c})p(\omega\vert\mathbf{x},\mathbf{c}),
\end{align*}
where both representations are two different ways to decompose the joint distribution over $(\mathbf{x},\mathbf{c},\omega)$, with $\pi(\omega)$ being the prior distribution of the CFG scale $\omega$. From this formula, if we could predict the guidance weight $\omega$ by observing $\mathbf{x}$ and $\mathbf{c}$, i.e., if we know $p(\omega\vert\mathbf{x},\mathbf{c})$, then sampling $(\mathbf{x},\mathbf{c},\omega)$ from $p_{\text{\text{data}}}(\mathbf{c})\pi(\omega)p_{\text{\text{data}}}(\mathbf{x}\vert\mathbf{c},\omega)$ can be alternatively achieved by: 1) sampling $(\mathbf{x},\mathbf{c})$ from $p_{\text{data}}(\mathbf{x},\mathbf{c})$, and 2) predicting most likely $\omega$ using $p(\omega\vert\mathbf{x},\mathbf{c})$.

We approximate $p(\omega\vert\mathbf{x},\mathbf{c})$ with a U-Net encoder network with 1-dimensional output, called \textit{CFG weight estimator} $\omega_{\bm{\phi}}$. The input of $\omega_{\bm{\phi}}$ network is a single-channel matrix with $(i,j)$-th value as the multiplication of the $i$/$j$-th values of $\mathbf{x}$/$\mathbf{c}$ CLIP embeddings, respectively. As this matrix is high-dimensional, we input the downsampled $64\times 64\times 1$ matrix to the U-Net encoder. These CLIP embeddings are also used to condition the network. With DM pretrained at Stage 1, which is supposed to be sufficiently close to the data distribution, we train the CFG weight estimator by minimizing $\mathbb{E}_{p_{\text{\text{prior}}}(\mathbf{z})p_{\text{\text{data}}}(\mathbf{c})\pi(\omega)}[\Vert \omega - \omega_{\bm{\phi}}(\mathbf{\hat{x}}(\mathbf{z},\mathbf{c},\omega),\mathbf{c}) \Vert_{2}^{2}]$, where $\mathbf{\hat{x}}(\mathbf{z},\mathbf{c},\omega)$ is a clean base-resolution sample drawn the teacher diffusion. Then, $\omega_{\bm{\phi}}(\mathbf{x},\mathbf{c})$-value becomes the point estimation of $p(\omega\vert\mathbf{x},\mathbf{c})$.

\subsection{PaGoDA Pipeline with Classifier-Free Guidance}\label{sec:cfg_loss}

We replace the adversarial loss in Stages 2 and 3 with the proposed classifier-free guided adversarial loss. In Stage 3, we shift the focus from $\mathbf{x}\in\mathbb{R}^{d}$ to $\mathbf{x}_{\text{high}}\in\mathbb{R}^{4^{n}d}$ to effectively capture high-frequency details. Additionally, in both Stages 2 and 3, we replace the input of the generator in the reconstruction loss to be classifier-free guided DDIM inversion noise. To enhance text-sample alignment, we further regularize training with CLIP~\citep{radford2021learning} similarity. For training, we use the ViT-L/14~\citep{dosovitskiy2020image} CLIP model pretrained on YFCC100M~\citep{thomee2016yfcc100m}, while for evaluation, we use the ViT-g/14 CLIP model pretrained on LAION-2B~\citep{schuhmann2022laion}, minimizing the risk of overfitting.

\section{Experiments}

\subsection{PaGoDA Tested on ImageNet without CFG}

We conduct experiments on ImageNet using PaGoDA without CFG to validate the core pipeline described in Section~\ref{sec:PaGoDA}, utilizing the discrete time diffusion scheduling proposed by EDM~\citep{karras2022elucidating}. Before training, we collect DDIM inversion latent representations for all ImageNet data using the Heun method~\citep{karras2022elucidating} with 40 timesteps (79 NFE). Throughout the experiments, we maintain the batch size to be 256 for both $\mathcal{L}_{\text{\text{rec}}}$ and $\mathcal{L}_{\text{adv}}$ in Stages 2 and 3. We initialize our base resolution generator with the pre-trained diffusion U-Net. Following CTM~\citep{kim2023consistency}, we implement adaptive weighting~\citep{esser2021taming} with $\lambda=0.2\frac{\Vert\nabla_{\bm{\theta}^{l}}\mathcal{L}_{\text{\text{rec}}}\Vert_{2}^{2}}{\Vert\nabla_{\bm{\theta}^{l}}\mathcal{L}_{\text{adv}}\Vert_{2}^{2}}$, where $\bm{\theta}^{l}$ represents the last layer of the generator.

\begin{figure*}[!t]
	\centering
 \begin{subfigure}{0.47\linewidth}
		\centering
		\includegraphics[width=\linewidth]{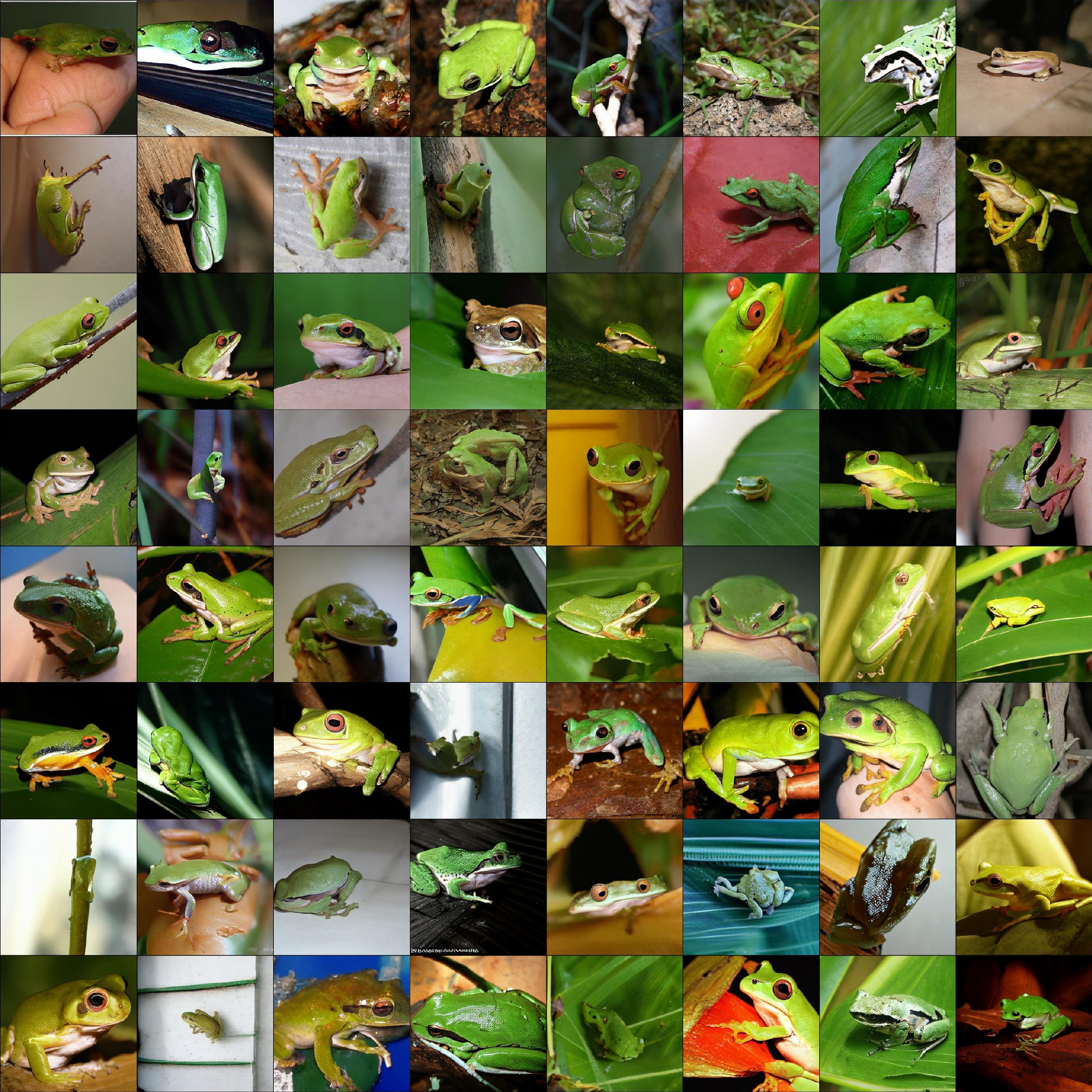}
	\end{subfigure}	
 \begin{subfigure}{0.47\linewidth}
		\centering
		\includegraphics[width=\linewidth]{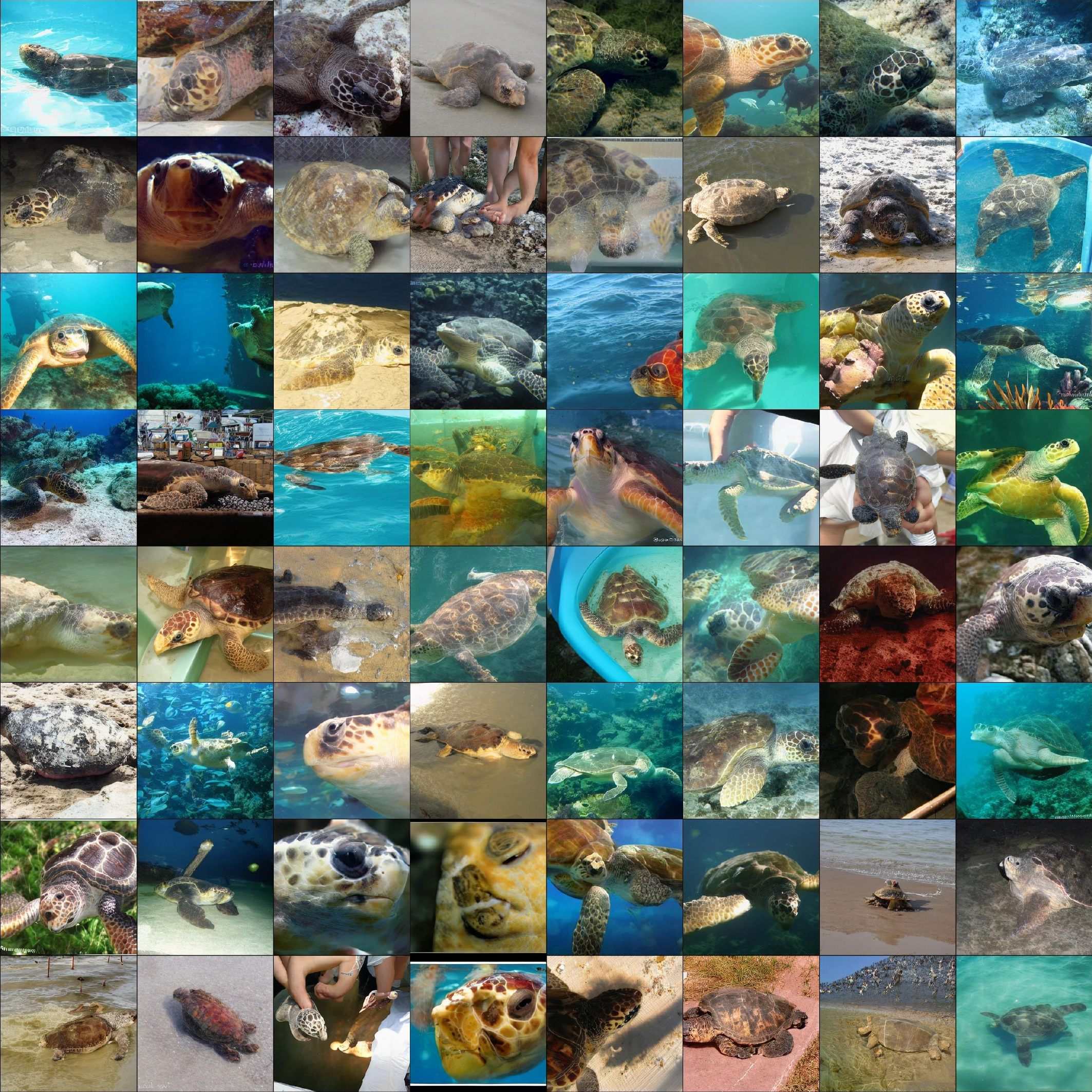}
	\end{subfigure}	
	\caption{Uncurated samples generated by PaGoDA  at resolution $512\times 512$ \textit{without CFG}. Left: class 31 (tree frog); Right: class 33 (loggerhead turtle).}
    \label{fig:imagenet_512}
	 \vskip -0.2in
\end{figure*}

For higher resolution generation, we double the previous resolution by adding two auxiliary ResNet blocks followed by one upsampler ResNet block. The previously trained generator remains frozen, except for the highest-resolution blocks, which are unfrozen. We then train these newly added blocks along with the unfrozen parts, using a fixed GAN weight of $\lambda=1.0$. Appendix~\ref{sec:imagenet_exp_details} provides additional details. By freezing part of the trained generator, we achieve greater stability in super-resolution training without adaptive weighting. See Figure~\ref{fig:imagenet_512} for uncurated $512\times 512$ random samples of ImageNet without CFG.

\subsubsection{Quantitative Results}

Table~\ref{tab:imagenet_comprehensive} presents the performance of PaGoDA. Our model consistently outperforms all existing models across all resolutions, achieving SOTA FIDs without the need of CFG and any other stabilization tricks for GAN. Remarkably, PaGoDA's Inception Score (IS)~\citep{salimans2016improved} is on par with other diffusion and GAN models that employed CFG, which implies that PaGoDA samples are as distinctive as CFG samples. Also, PaGoDA generates samples as diverse as the real data distribution, evidenced by diversity recall metric~\citep{kynkaanniemi2019improved}, where the PaGoDA reports 0.63 for $64\times 64$ resolution (data's recall is 0.67). In contrast, StyleGAN-XL is far behind of PaGoDA in terms of the diversity metric, reporting 0.52 for $64\times 64$ resolution. Note that we used StyleGAN-XL's discriminator in PaGoDA training, implying that the reconstruction loss significantly improves the sample diversity.

\begin{table}[ht]
\vskip -0.1in
	\caption[Experimental results of PaGoDA on ImageNet.]{Experimental results of PaGoDA on ImageNet.}
 \vskip 0.01in
	\label{tab:imagenet_comprehensive}
	\scriptsize
	\centering
 \setlength{\tabcolsep}{3pt}
	\begin{tabular}{lccccccccccccc}
		\toprule
        \multirow{2}{*}{Model} & \multirow{2}{*}{\shortstack{Sampling\\NFE}} & \multicolumn{3}{c}{Without CFG} & \multicolumn{3}{c}{With CFG} & \multicolumn{3}{c}{Without CFG} & \multicolumn{3}{c}{With CFG} \\
        && FID $\downarrow$ & IS $\uparrow$ & Recall $\uparrow$ & FID & IS & Recall & FID & IS & Recall & FID & IS & Recall \\\\
        &&\multicolumn{6}{l}{\textbf{$\mathbf{64\times 64}$ resolution}} & \multicolumn{6}{l}{\textbf{$\mathbf{128\times 128}$ resolution}}\\\midrule
        RIN~\citep{jabri2022scalable} & 250 & 1.23 & 66.5 & - & - & - & - & 2.75 & 144.1 & - & - & - & - \\
        simple Diffusion~\citep{hoogeboom2023simple} & 250 & - & - & - & - & - & - & 1.91 & 171.9 & - & 2.05 & 189.9 & - \\
        VDM++~\citep{kingma2023understanding} & 79 & 1.43 & 63.7 & - & - & - & - & 1.75 & 171.1 & - & 1.78 & 190.5 & - \\
        StyleGAN-XL~\citep{sauer2022stylegan} & 1 & - & - & - & 1.51 & \textbf{82.35} & 0.52 & - & - & - & 1.81 & \textbf{200.55} & 0.55 \\
        CTM~\citep{kim2023consistency} & 1 & 1.92 & 70.38 & 0.57 & - & - & - & - & - & - & - & - & - \\
        \cc{15}PaGoDA (ours) & \cc{15}1 & \cc{15}\textbf{1.21} & \cc{15}76.47 & \cc{15}\textbf{0.63} &\cc{15}- &\cc{15}- &\cc{15}- & \cc{15}\textbf{1.48} & \cc{15}174.36 & \cc{15}\textbf{0.61} & \cc{15}- & \cc{15}- & \cc{15}- \\\\
&&\multicolumn{6}{l}{\textbf{$\mathbf{256\times 256}$ resolution}} & \multicolumn{6}{l}{\textbf{$\mathbf{512\times 512}$ resolution}}\\\midrule
DiT-XL~\citep{peebles2023scalable} & 250 & 9.62 & 121.5 & - & 2.27 & 278.2 & - & 12.03 & 105.3 & - & 3.04 & 240.8 & - \\      
simple Diffusion~\citep{hoogeboom2023simple} & 250 & 2.77 & 211.8 & - & 2.44 & 256.3 & - & 3.54 & 205.3 & - & 3.02 & 248.7 & - \\
VDM++~\citep{kingma2023understanding} & 250 & 2.40 & 225.3 & - & 2.12 & 267.7 & - & 2.99 & 232.2 & - & 2.65 & 278.1 & - \\
EDM2-XXL~\citep{karras2023analyzing} & 63 & - & - & - & - & - & - & 1.91 & - & - & 1.81 & - & - \\
StyleGAN-XL~\citep{sauer2022stylegan} & 1 & - & - & - & 2.30 & \textbf{265.12} & 0.53 & - & - & - & 2.41 & \textbf{267.75} & 0.52 \\
\cc{15}PaGoDA (ours) & \cc{15}1 & \cc{15}\textbf{1.56} & \cc{15}259.61 & \cc{15}\textbf{0.59} &\cc{15}- &\cc{15}- &\cc{15}- & \cc{15}\textbf{1.80} & \cc{15}251.31 & \cc{15}\textbf{0.58} & \cc{15}- & \cc{15}- & \cc{15}- \\
                    \bottomrule
	\end{tabular}
 \end{table}

\subsubsection{Discussion on Base Resolution}

When applying PaGoDA pipeline, the choice of downsampled base resolution in Stage 1 will be primarily determined by available computational resources. Thus, we investigate the impact of the base resolution at this section. To understand the impact, we conducted experiments at $32\times 32$ and $64\times 64$ resolutions, as summarized in Table~\ref{tab:implementation}. Starting at resolutions below $32\times 32$ imposes excessive complicacy on the upscaling network, while higher resolutions significantly increase the computational costs at the Stage 1. Therefore, our analysis focuses on these two resolutions, balancing between computational efficiency and upscaling feasibility.

\begin{table}[ht]
\scriptsize
\vskip -0.1in
 \caption{Ablation of base resolution.}
 \vskip 0.02in
 \label{tab:implementation}
	\centering
 \setlength{\tabcolsep}{3pt}
 \begin{tabular}{lcccccccccc}
		\toprule
        Model & Base Res & Upscaled Res & NFE & FID & Base Res & Upscaled Res & NFE & FID & Speed [s] & Params \\\midrule
        Teacher Diffusion & $32\times 32$ & $32\times 32$ & 79 & 1.75 & $64\times 64$ & $64\times 64$ & 79 & 2.44 & 3.16s & 296M \\\midrule
        \multirow{5}{*}{\shortstack[l]{PaGoDA}} & $32\times 32$ & $32\times 32$ & 1 & 0.79 \\
        & $32\times 32$ & $64\times 64$ & 1 & 1.34 & $64\times 64$ & $64\times 64$ & 1 & 1.21 & 0.040s & 296M \\
        & $32\times 32$ & $128\times 128$ & 1 & 1.61 & $64\times 64$ & $128\times 128$ & 1 & 1.48 & 0.041s & 299M \\
        & $32\times 32$ & $256\times 256$ & 1 & 1.83 & $64\times 64$ & $256\times 256$ & 1 & 1.56 & 0.044s & 301M \\
        & &&&& $64\times 64$ & $512\times 512$ & 1 & 1.80 & 0.046s & 302M \\
                    \bottomrule
	\end{tabular}
 \end{table}%

We utilized only 1 H100 node with 8 GPUs for diffusion training on $32\times 32$ with 4096 batch size. Also, for $64\times 64$ diffusion, we borrow a pretrained checkpoint \cite{song2023consistency}, which used $\ge 32$\footnote{This is an estimate.} A100 GPUs to train with 4096 batch size. Results in Table~\ref{tab:implementation} demonstrate that the diffusion model trained in Stage 1 maintains robust performance across both resolutions. Interestingly, the one-step generator distilled in Stage 2 consistently outperforms the teacher model, likely benefiting from the effectiveness of StyleGAN-XL~\citep{sauer2022stylegan}, combined with the reconstrcution loss. In Stage 3, the degree of upscaling from the base resolution emerges as the most influential factor for the quality, with upscaling up to 8x showing minimal performance degradation across both tested resolutions.

The upscaler in PaGoDA refines coarse samples generated at lower resolutions, making the pipeline inherently aligned with the scaling laws of smaller resolutions. This design is advantageous, as scaling laws typically worsen with increasing resolution~\citep{henighan2020scaling}, while PaGoDA leverages the more favorable scaling dynamics at lower resolutions to maintain efficiency. Furthermore, the lightweight upscaling module introduces minimal additional latency, keeping inference times nearly identical to those at the base resolution. This practical efficiency makes PaGoDA a promising solution for scalable diffusion model training across various computational settings.

\subsubsection{Discussion on Upscaling Capability}

In Stage 3, we train the super-resolution module using a combination of reconstruction and adversarial losses. As shown in Figures~\ref{fig:upsampling} and \ref{fig:effect_to_adversarial}, we compare PaGoDA’s performance to that of StyleGAN-XL. The comparison reveals key insights: 1) PaGoDA maintains consistent object alignment across resolution jumps, largely due to the reconstruction loss, and 2) its performance is strongly influenced by the GAN component, which plays a crucial role in capturing high-frequency details.

\begin{wraptable}{r}{0.4\textwidth}
\vskip -0.1in
	\caption{Comparison on upsampling.}
 \vskip 0.01in
	\label{tab:comparison_to_sd}
	\scriptsize
	\centering
 \setlength{\tabcolsep}{3pt}
	\begin{tabular}{lcccc}
		\toprule
        Model & Resolution & Params & NFE & FID \\\midrule
        \multirow{2}{*}{\shortstack[l]{EDM2}} &
        $64^2$ DM & 1.1B & 63 & 1.33 \\
        & $512^2$ LDM & 1.1B & 63+1 & 1.96 \\\midrule
        \multirow{3}{*}[-0.4em]{PaGoDA} & $64^2$ DM (teacher) & 0.3B & 79 & 2.44 \\\cmidrule(lr){2-5}
        & $64^2\rightarrow 64^2$ & 0.3B & 1 & 1.21 \\
        & $64^2\rightarrow 512^2$ & 0.3B & 1 & 1.80 \\
                    \bottomrule
	\end{tabular}
 \end{wraptable}
Other upsampling methods, such as SD and Cascaded Diffusion Models (CDM)~\citep{ho2022cascaded} also target high-quality upscaling. While PaGoDA, CDM, and SD share the same goal, they adopt different approaches, making them complementary rather than competing solutions. In fact, their strengths can be combined to enhance overall compression and upscaling performance. For instance, CDM or PaGoDA can be applied to the latent space of SD, integrating their techniques for better results. Despite their compatibility, it is still essential to assess how these methods compare in terms of their upscaling effectiveness. In the following analysis, we break down the upscaling capabilities of PaGoDA, CDM, and SD to understand their respective strengths and potential limitations.

Since PaGoDA is experimented based on EDM~\citep{karras2022elucidating}, we adapted the experimental results from EDM2~\citep{karras2023analyzing} to facilitate a direct comparison with PaGoDA in the upscaling performance. EDM2 presents results for both pixel DM and latent DM. In latent diffusion, a $512\times 512\times 3$ image is compressed into a $64\times 64\times 4$ latent space for training DM, while pixel diffusion operates directly on $64\times 64\times 3$ images, sharing the identical network architecture used in its latent DM. As reported in Table~\ref{tab:comparison_to_sd}, EDM2 shows a minor performance decline from 1.33 to 1.96.

\begin{wrapfigure}{r}{0.35\textwidth}
 \vskip -0.2in
  \centering
		\includegraphics[width=\linewidth]{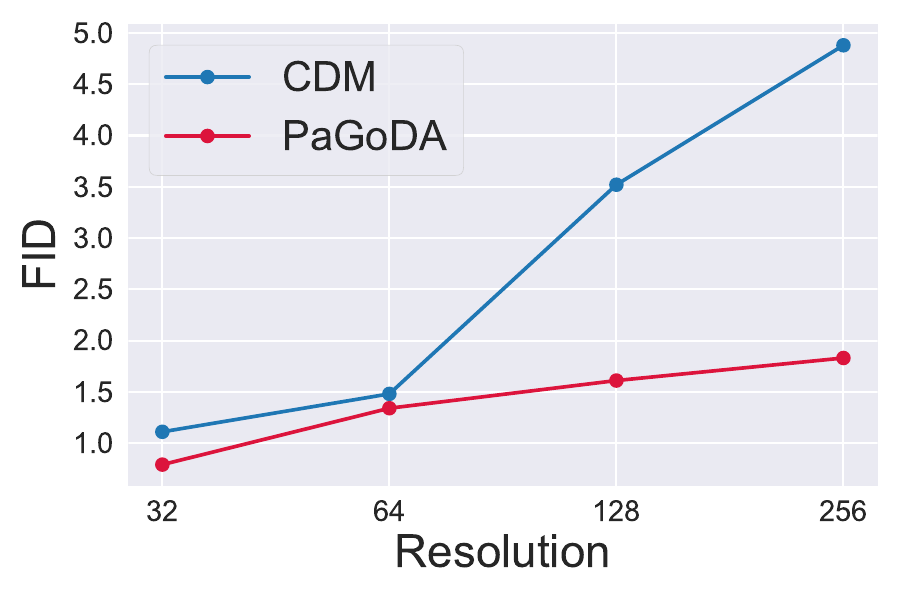}
  \vskip -0.05in
		\caption{Comparison between PaGoDA and CDM.}
  \label{fig:pagoda_vs_cdm}
  \vskip -0.2in
\end{wrapfigure}
Similarly, PaGoDA exhibits a comparable performance drop from 1.21 to 1.80 when upscaling from $64\times 64$ to $512\times 512$. This similarity suggests that PaGoDA’s upscaling capacity aligns closely with that of the LDM framework, indicating minimal performance differences even when handling high-resolution data.

Lastly, when comparing PaGoDA to CDM, we observe in Figure~\ref{fig:pagoda_vs_cdm} that CDM encounters significant performance drops beyond certain dimensional thresholds ($128\times 128$), while PaGoDA maintains consistent performance across varying resolutions. This robustness makes PaGoDA a reliable option for high-resolution generation, with its performance remaining steady even as resolution increases.

\subsection{Discussion on Controllability}

\begin{figure*}[t]
	\centering
 \begin{subfigure}{0.3\linewidth}
		\centering
		\includegraphics[width=\linewidth]{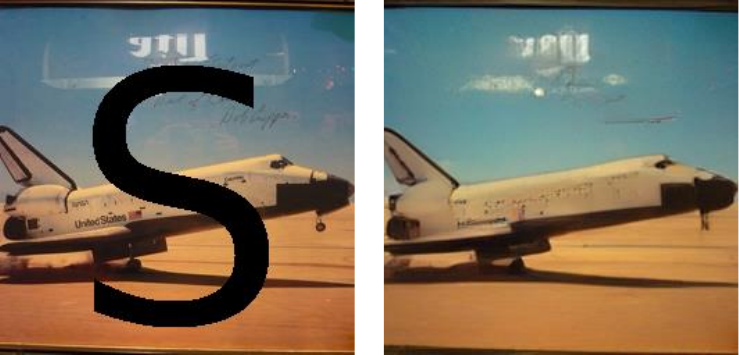}
		\subcaption{Inpainting}
	\end{subfigure}
 \hfil
 \begin{subfigure}{0.3\linewidth}
		\centering
		\includegraphics[width=\linewidth]{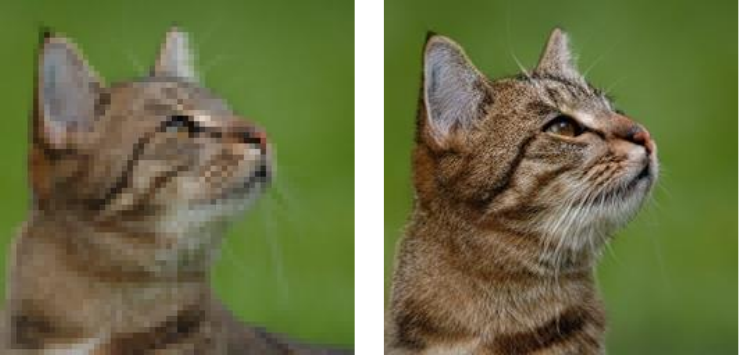}
		\subcaption{Super-resolution}
	\end{subfigure}
 \hfil
 \begin{subfigure}{0.3\linewidth}
		\centering
		\includegraphics[width=\linewidth]{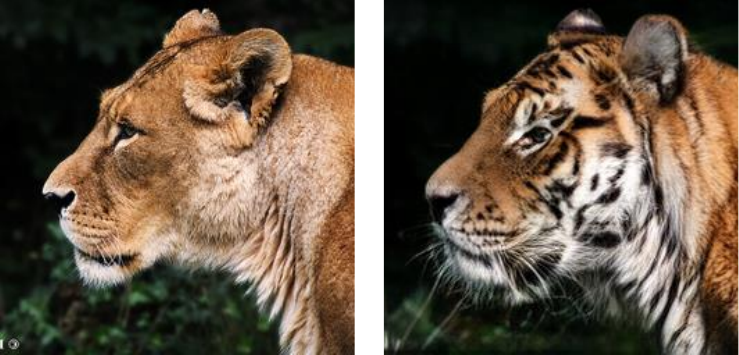}
		\subcaption{Class transfer}
	\end{subfigure}
 \begin{subfigure}{0.96\linewidth}
		\centering
		\includegraphics[width=\linewidth]{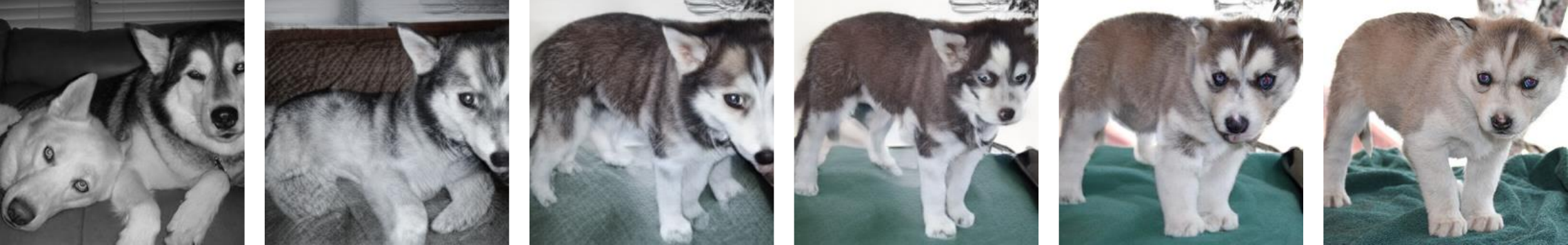}
		\subcaption{Latent interpolation}
	\end{subfigure}
 \vskip -0.05in
	\caption{Controllable generation of PaGoDA with various tasks.}
    \label{fig:inverse_problem}
	  \vskip -0.1in
\end{figure*}

Once we have a trained PaGoDA generator $G_{\bm{\theta}_{0}}$, we can utilize it for solving inverse problems~\citep{chung2022diffusion} and for controllable generation~\citep{zhang2023adding} in a training-free manner~\citep{he2023manifold}.

\textbf{Latent Optimization} We consider the inverse problem: $\mathbf{y}=\mathcal{A}(\mathbf{x})+\eta$, where 
 $\mathbf{y}$ represents the observation, and $\mathcal{A}:\mathbb{R}^{d}\rightarrow\mathbb{R}^{m}$ with $d\ge m$ is a known operator. The restored data $\mathbf{x}$ can be reconstructed by optimizing the latent. Specifically, if $\mathbf{z}^{*}\in\argmin_{\mathbf{z}}\Vert \mathbf{y}-\mathcal{A}(G_{\bm{\theta}_{0}}(\mathbf{z},\mathbf{c})) \Vert_{2}^{2}$, then $G_{\bm{\theta}_{0}}(\mathbf{z}^{*},\mathbf{c})$ is the best possible estimate of the solution for the inverse problem. Figure~\ref{fig:inverse_problem}-(a) displays the outcomes of an inpainting task where latent optimization is employed with Adam optimizer~\citep{kingma2014adam}.

\textbf{DDIM Inversion} Specific tasks, such as super-resolution illustrated in Figure~\ref{fig:inverse_problem}-(b) and class transfer depicted in Figure~\ref{fig:inverse_problem}-(c), can be effectively addressed without relying on latent optimization. For these tasks, we apply DDIM inversion to the downsampled observations, then map the DDIM latent back to RGB pixel by feeding the latent into the decoder. Generally, using DDIM inversion yields superior outcomes compared to latent optimization for these types of tasks.

\textbf{Latent Interpolation} Building on techniques from GAN research, we also explored latent interpolation for style mixing. Despite our model's latent dimension being larger than the typical 512-dimensional style vector used in GAN, our observations indicate that latent mixing by slerp operation~\citep{shoemake1985animating, karras2019style} achieves effective results, as demonstrated in Figure~\ref{fig:inverse_problem}-(d).

\subsection{Text-to-Image Generation}

We collect the data-latent pairs on the CC12M dataset~\citep{changpinyo2021cc12m} through DDIM inversion and utilize the filtered COYO-700M~\citep{kakaobrain2022coyo-700m} dataset for adversarial training. The filtering criteria include only data with CLIP score (measured by ViT-B/32~\citep{Radford2021LearningTV}) higher than 32, and aesthetic score-v2~\citep{schuhmann2022laion} higher than 5.0. Due to concerns regarding sensitive content in the open-sourced LAION dataset~\citep{schuhmann2022laion}, we were unable to conduct large-scale diffusion training for Stage 1. This constraint led us to focus primarily on stages 2 and 3, leveraging pretrained open-source checkpoints. For the pretrained teacher diffusion, we used the DeepFloyd-IF model~\citep{deepfloyd}, trained on $64\times 64$ pixel space. For further experimental details, see Appendix~\ref{sec:t2i_exp_details}.

\begin{wrapfigure}{r}{0.35\textwidth}
 \vskip -0.2in
  \centering
		\includegraphics[width=\linewidth]{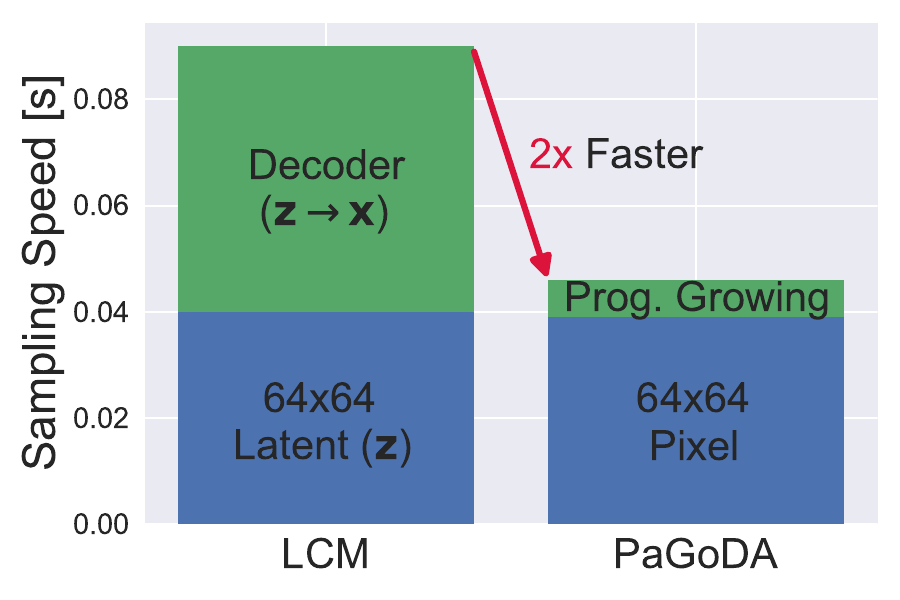}
  \vskip -0.05in
		\caption{PaGoDA offers faster inference than the one-step LCM.}
  \label{fig:pagoda_bar_figure}
  \vskip -0.2in
\end{wrapfigure}
Table~\ref{tab:t2i_coco_2017} compares our PaGoDA mainly with the distilled models from SD v1.5 on $512\times 512$. One notable observation from the table is that, even though the latent distilled model generates the latent representation in a single step, additional time is required for decoding this latent into image. In contrast, PaGoDA (on pixel teacher) eliminates such decoding step, thereby overcoming the time constraints associated with distilling SD models. For a more detailed breakdown of the time taken by each component, see Figure~\ref{fig:pagoda_bar_figure}.

Returning to the performance results in the table, PaGoDA achieves performance comparable to that of the teacher model. This superior performance is also observed on a different test set as shown in Table~\ref{tab:t2i_coco_2014}, further demonstrating PaGoDA’s scalability on text-to-image tasks.

\begin{table}[t]
\vskip -0.05in
\begin{minipage}[t]{.52\linewidth}
\scriptsize
 \caption[Experimental results on T2I. FID-5K is measured on MSCOCO-2017 validation dataset. CLIP score is measured by the ViT-g/14 backbone.]{Experimental results on T2I. FID-5K is measured on MSCOCO-2017~\citep{lin2014microsoft} validation data. CLIP score is measured by the ViT-g/14 backbone. Our model uses DeepFloy-IF as the pre-trained diffusion.}
 \label{tab:t2i_coco_2017}
 \resizebox{.95\linewidth}{!}{
	\centering
 \setlength{\tabcolsep}{3pt}
 \begin{tabular}{lccccc}
		\toprule
        Model & Params & NFE & Speed [s] & FID $\downarrow$ & CLIP $\uparrow$ \\\midrule
        SD1.5~\citep{rombach2022high} & 0.9B & 50+1 & 2.59s & 19.1 & 31.3 \\
        DeepFloyd-IF~\citep{deepfloyd} & 0.9B & 27 & 2.95s & 22.3 & 28.1 \\
        \midrule
        \multicolumn{5}{l}{\textbf{Latent Distillation Models based on SD1.5~\citep{rombach2022high}}}\\
        CAD~\citep{meng2023distillation} & 0.9B & 8+1 & 0.34s & 24.2 & 30.0 \\
        PD~\citep{salimans2021progressive} & 0.9B & 4+1 & 0.21s & 26.4 & 30.0 \\
        LCM~\citep{luo2023latent} & 0.9B & 2+1 & 0.13s & 30.4 & 29.3 \\
        InstaFlow~\citep{liu2023instaflow} & 0.9B & 1+1 & 0.09s & 23.4 & 30.4 \\
        UFOGen~\citep{xu2023ufogen} & 0.9B & 1+1 & 0.09s & 22.5 & 31.1 \\
        Scott~\citep{liu2024scott} & 0.9B & 1+1 & 0.09s & 21.9 & 31.2 \\
        ADD~\citep{sauer2023adversarial} & 0.9B & 1+1 & 0.09s & 19.7 & 32.6 \\\midrule
        \multicolumn{5}{l}{\textbf{Pixel Distillation Model based on DeepFloyd-IF~\citep{deepfloyd}}}\\
        \cc{15}PaGoDA (ours) & \cc{15}0.9B & \cc{15}1 & \cc{15}0.05s & \cc{15}20.4 & \cc{15}31.2\\
                    \bottomrule
	\end{tabular}}
 \end{minipage}%
 \hfil
\begin{minipage}[t]{.43\linewidth}
	\caption[Experimental results on T2I. FID-30K is measured on MS COCO-2014 validation data.]{Experimental results on T2I. FID-30K is based on MSCOCO-2014~\citep{lin2014microsoft} validation data. Speed is measured on A100.}
	\label{tab:t2i_coco_2014}
	\scriptsize
 \resizebox{.94\linewidth}{!}{
	\centering
 \setlength{\tabcolsep}{3pt}
	\begin{tabular}{lcccc}
 \toprule
        Model & Params & Speed [s] & FID $\downarrow$ \\\midrule
        eDiff-I~\citep{balaji2022ediff} & 9.1B & 32.0s & 6.95 \\
        LDM~\citep{rombach2022high} & 1.5B & 9.4s & 12.63 \\
        Imagen~\citep{saharia2022photorealistic} & 3.0B & 9.1s & 7.27 \\
        SD1.5~\citep{rombach2022high} & 0.9B & 2.9s & 9.62 \\
        PixArt-$\alpha$~\citep{chen2023pixart} & 0.6B & - & 10.65 \\
        Scott~\citep{liu2024scott} & 0.9B & 0.13s & 12.22 \\
        GigaGAN~\citep{kang2023scaling} & 1.0B & 0.13s & 9.09 \\
        StyleGAN-T~\citep{sauer2023stylegan} & 1.0B & 0.10s & 13.90 \\
        InstaFlow~\citep{liu2023instaflow} & 0.9B & 0.09s & 13.10 \\
        UFOGen~\citep{xu2023ufogen} & 0.9B & 0.09s & 12.78 \\
        DMD~\citep{yin2023one} & 0.9B & 0.09s & 11.49 \\
        LAFITE~\citep{zhou2022towards} & 75M & 0.02s & 26.94 \\
        \cc{15}PaGoDA (ours) & \cc{15}0.9B & \cc{15}0.05s & \cc{15}10.23 \\
                    \bottomrule
	\end{tabular}}
 \end{minipage}
\end{table}

\section{Conclusion}
PaGoDA introduces a training pipeline that can democratize the diffusion training by cutting training budget with orders of magnitudes. The pipeline is consisted of three stages: 1) we pretrain the diffusion models on the downsampled data, 2) we distill the teacher diffusion into a one-step generator on the downsampled data, and 3) we train an upsampler module until we reach to the desired resolution.

\section*{Acknowledgement}
This project was supported by Sony, ARO (W911NF-21-1-0125), ONR (N00014-23-1-2159), and the CZ Biohub. Computational resource of AI Bridging Cloud Infrastructure (ABCI) provided by National Institute of Advanced Industrial Science and Technology (AIST) was used. We extend our special thanks to our colleagues Takashi Shibuya from Sony AI and Yutong He from Carnegie Mellon University for their invaluable feedback.

\clearpage
\newpage

\bibliography{dgm}
\bibliographystyle{unsrt}

\newpage
\section*{NeurIPS Paper Checklist}

\begin{enumerate}

\item {\bf Claims}
    \item[] Question: Do the main claims made in the abstract and introduction accurately reflect the paper's contributions and scope?
    \item[] Answer: \answerYes{} 
    \item[] Justification: We have made our abstract and introduction to accurately reflect the core contribution of the paper.

\item {\bf Limitations}
    \item[] Question: Does the paper discuss the limitations of the work performed by the authors?
    \item[] Answer: \answerYes{} 
    \item[] Justification: We have created a separate ``Limitations and Broader Impacts'' section in the appendix to enumerate potential limitations of our methodology, including the algorithmic, theoretical, and experimental limitations. 

\item {\bf Theory Assumptions and Proofs}
    \item[] Question: For each theoretical result, does the paper provide the full set of assumptions and a complete (and correct) proof?
    \item[] Answer: \answerYes{} 
    \item[] Justification: Although we omit some of assumptions in the main paper mainly due to page limit, we provide full details of assumptions and complete proof in the appendix.

    \item {\bf Experimental Result Reproducibility}
    \item[] Question: Does the paper fully disclose all the information needed to reproduce the main experimental results of the paper to the extent that it affects the main claims and/or conclusions of the paper (regardless of whether the code and data are provided or not)?
    \item[] Answer: \answerYes{} 
    \item[] Justification: We disclose all experimental details in the main paper and the appendix, including the hyperaparameters used and the datasets used with their filetering methodologies. For further reproducibility, we plan to release our code upon acceptance.

\item {\bf Open access to data and code}
    \item[] Question: Does the paper provide open access to the data and code, with sufficient instructions to faithfully reproduce the main experimental results, as described in supplemental material?
    \item[] Answer: \answerYes{} 
    \item[] Justification: In the reviewing process, we release our code to the reviewers to regenerate our experimental results. After the acceptance, we plan to release the code to the public.

\item {\bf Experimental Setting/Details}
    \item[] Question: Does the paper specify all the training and test details (e.g., data splits, hyperparameters, how they were chosen, type of optimizer, etc.) necessary to understand the results?
    \item[] Answer: \answerYes{} 
    \item[] Justification: We faithfully release our hyperparameters and experimental details in the appendix and the main paper.

\item {\bf Experiment Statistical Significance}
    \item[] Question: Does the paper report error bars suitably and correctly defined or other appropriate information about the statistical significance of the experiments?
    \item[] Answer: \answerNo{} 
    \item[] Justification: We have not reported error bars mainly due to the lack of computational resources.

\item {\bf Experiments Compute Resources}
    \item[] Question: For each experiment, does the paper provide sufficient information on the computer resources (type of compute workers, memory, time of execution) needed to reproduce the experiments?
    \item[] Answer: \answerYes{} 
    \item[] Justification: We explain how much compute resources we used for experiments in the appendix.
    
\item {\bf Code Of Ethics}
    \item[] Question: Does the research conducted in the paper conform, in every respect, with the NeurIPS Code of Ethics \url{https://neurips.cc/public/EthicsGuidelines}?
    \item[] Answer: \answerYes{} 
    \item[] Justification: We faithfully follow the code of ethics, suggested by the link above.

\item {\bf Broader Impacts}
    \item[] Question: Does the paper discuss both potential positive societal impacts and negative societal impacts of the work performed?
    \item[] Answer: \answerYes{} 
    \item[] Justification: We discuss the broader impacts as a separate section in the ``Limitations and Broader Impacts'' of the Appendix~\ref{app:limitation}.
    
\item {\bf Safeguards}
    \item[] Question: Does the paper describe safeguards that have been put in place for responsible release of data or models that have a high risk for misuse (e.g., pretrained language models, image generators, or scraped datasets)?
    \item[] Answer: \answerYes{} 
    \item[] Justification: For the T2I checkpoint release, we plan to use HuggingFace to enroll every users to the system so to control the downloaded user list. Additionally, we prohibited using the LAION dataset~\citep{schuhmann2022laion}, which includes NSFW contents. Instead, we used the COYO-700M~\citep{kakaobrain2022coyo-700m} dataset, a large-scale text-to-image dataset that removes NSFW images by NSFW image detectors~\citep{opennsfw,man} and texts that contain NSFW words~\citep{nsfw_text_1, nsfw_text_2, nsfw_text_3}. 

\item {\bf Licenses for existing assets}
    \item[] Question: Are the creators or original owners of assets (e.g., code, data, models), used in the paper, properly credited and are the license and terms of use explicitly mentioned and properly respected?
    \item[] Answer: \answerYes{} 
    \item[] Justification: We have properly credited the original owners of assets by citing them. In the code release, we comply the license and terms of the assets.

\item {\bf New Assets}
    \item[] Question: Are new assets introduced in the paper well documented and is the documentation provided alongside the assets?
    \item[] Answer: \answerYes{} 
    \item[] Justification: In the supplementary, we include the the details of the dataset/code/model via structured templates.

\item {\bf Crowdsourcing and Research with Human Subjects}
    \item[] Question: For crowdsourcing experiments and research with human subjects, does the paper include the full text of instructions given to participants and screenshots, if applicable, as well as details about compensation (if any)? 
    \item[] Answer: \answerNA{} 
    \item[] Justification: The paper does not involve crowdsourcing nor research with human subjects.

\item {\bf Institutional Review Board (IRB) Approvals or Equivalent for Research with Human Subjects}
    \item[] Question: Does the paper describe potential risks incurred by study participants, whether such risks were disclosed to the subjects, and whether Institutional Review Board (IRB) approvals (or an equivalent approval/review based on the requirements of your country or institution) were obtained?
    \item[] Answer: \answerNA{} 
    \item[] Justification: The paper does not involve crowdsourcing nor research with human subjects.

\end{enumerate}

\clearpage
\newpage
\appendix

\tableofcontents
	\newpage
	\parttoc

\begin{figure*}[!t]
	\centering
 \includegraphics[width=\linewidth]{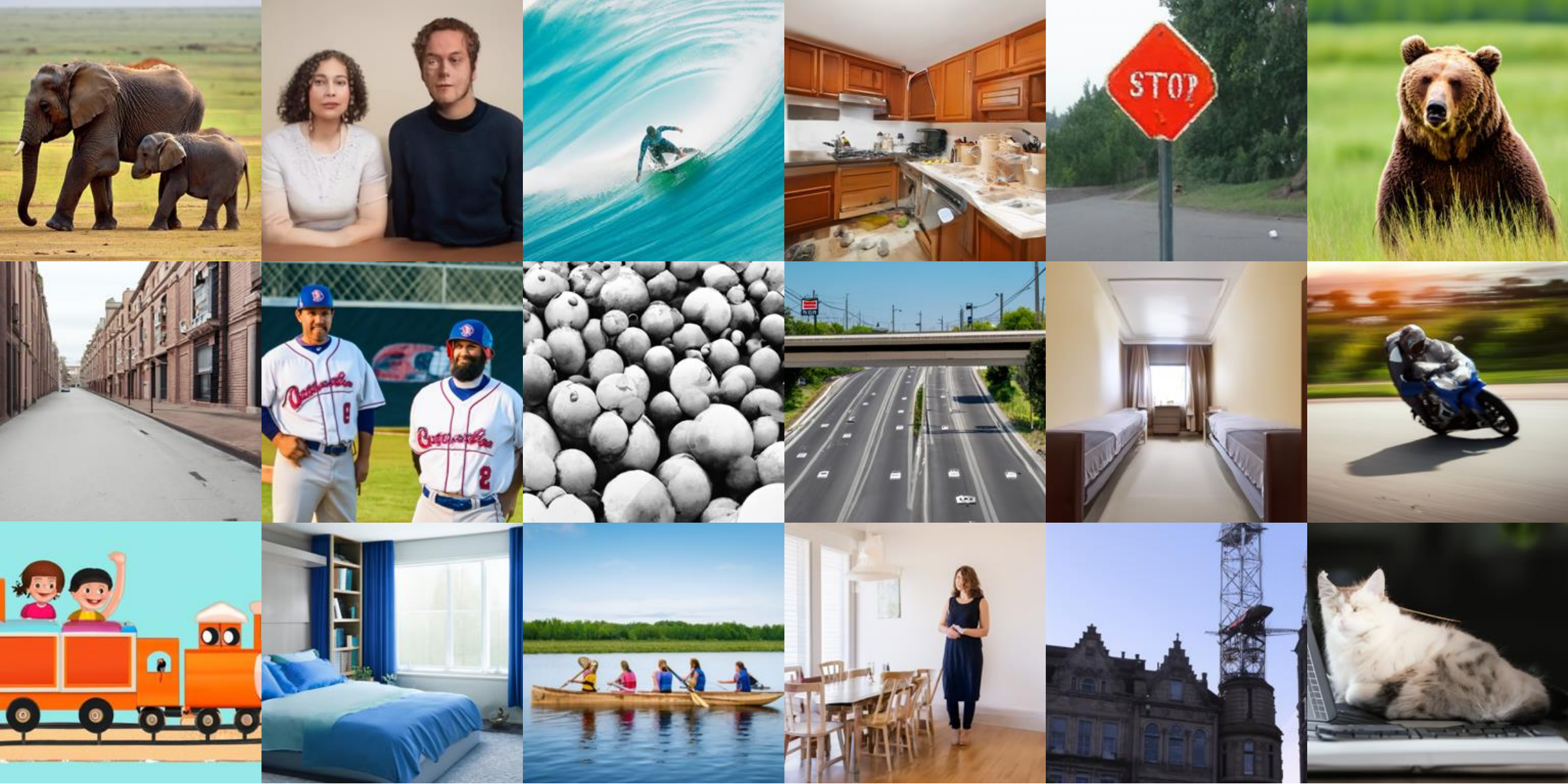}
	\caption{Text-to-image samples from PaGoDA.}
    \label{fig:t2i_many}
\end{figure*}

\section{Experimental Details}

\subsection{Conditional Generation with ImageNet}\label{sec:imagenet_exp_details}

Throughout the experiments, we omit the class condition $\mathbf{c}$ otherwise mentioned for notational simplicity.

\textbf{Dataset Construction.} We loaded ImageNet2014\footnote{\url{https://www.image-net.org/index.php}} dataset using center cropping and downsampling using the bicubic algorithm from the PIL python package. To augment the data, we applied a horizontal random flip, and obtained each of latent representations by solving the EDM's 2nd-order ODE sampler (Heun's method)~\cite{karras2022elucidating} with their suggested diffusion time scheduling and timestep selection. Consequently, in total, we processed approximately 2.5 million data instances forward in time using the PF-ODE to prepare for training. This computational cost of constructing the training dataset is comparable to sampling an equivalent volume of sample from a pre-trained diffusion model.

\textbf{GAN Details.} We adopted the discriminator architecture from StyleGAN-XL. Initially, We loaded DeiT-base~\citep{touvron2021training} and EfficientNet-lite~\citep{tan2019efficientnet} as feature extractors, in line with StyleGAN-XL's setup. When processing real or fake data through the discriminator, we first applied differentiable augmentation (DiffAugment)~\citep{zhao2020differentiable}, incorporating three transformations: \textit{Translation}, \textit{Cutout}, and \textit{Color}. Interestingly, we observed no performance differences between the \textit{unconditional} and \textit{conditional} discriminators. We hypothesize that this lack of disparity arises because the discriminator primarily updates the generator to refine high-frequency details, while preserving the low-frequency global semantics due to the reconstruction power. Additionally, we opted not to use additional techniques to tame the GAN training, such as R1 regularization~\citep{mescheder2018training} or path length regularization~\citep{karras2019style} in our GAN training. PaGoDA's training generally remains stable due to its reconstruction loss, which is consistent with our theoretical expectation (Theorem~\ref{th:stable_pagoda}).

We conducted tests on GANs under two distinct scenarios. Initially, following the approach used in Stable Diffusion's VAE training, we introduced both the real data $\mathbf{x}$ and the reconstructed sample $\tilde{\mathbf{x}}=G_{\bm{\theta}}(E(\mathbf{x}))$ to the discriminator, training it to differentiate between the two while updating the generator to maximize $\log{D_{\bm{\psi}}(\tilde{\mathbf{x}})}$. In this setup, as the reconstruction only utilizes the latent representation $E(\mathbf{x})$, the generation quality is not improved.

In the second scenario, adhering to the traditional GAN framework, we trained the discriminator using randomly sampled real data alongside randomly generated fake data 
$\tilde{\mathbf{x}}=G_{\bm{\theta}}(\mathbf{z})$ from $\mathbf{z}\sim p_{\text{\text{prior}}}(\mathbf{z})$. Then, the endeavor of maximizing $\log{D_{\bm{\psi}}(\tilde{\mathbf{x}})}$ now significantly improves the generation quality. Overall, we observed no performance degradation when both types of GAN training were applied to the generator. However, given our limited budget and the goal to develop a generative model rather than a compression model, we opted to proceed solely with the second type of GAN setup.

\textbf{Reconstruction Details.} For the reconstruction loss, we train the generator $G_{\bm{\theta}}$ by comparing the original data $\mathbf{x}\sim p_{\text{\text{data}}}(\mathbf{x})$ and its reconstructed counterpart $G_{\bm{\theta}}(E(\mathbf{x},\mathbf{c}),\mathbf{c})$ at the data's resolution, where $E(\mathbf{x},\mathbf{c})$ is the solution of the DDIM inversion. Since our training occurs in pixel space, we conduct this comparison in the feature space using the Learned Perceptual Image Patch Similarity (LPIPS) metric, and there is no need to develop a new feature extractor in latent space. We experimented with features extracted from DeiT-base~\citep{touvron2021training} and EfficientNet-lite~\citep{tan2019efficientnet}; however, we observed no notable improvement from using LPIPS.

For the training, we use the RAdam~\citep{liu2019variance} with learning rate of 8e-6 for the decoder and 2e-3 for the discriminator, and without weight decay. We use the EMA of 0.999, and all reported FIDs are based on the EMA checkpoint. Until $256^2$ resolution, we use only 1 H100 node (with 80Gb memory) to train, and we use 8 A100 nodes (with 40Gb memory, in total $8\times8=64$ GPUs) to train the $512^2$ model. Throughout the experiments, we use the batch size of 256. 

For the concerns on the overfitting, we provide additional results in Tables 6 and 7. 

\begin{table}[t]
\centering
\scriptsize
\label{tab:fd-64x64}
\vskip -0.1in
 \caption{Comparison on ImageNet $64\times64$. We evaluate scores, including Fr\'echet distance on DINOv2 features~\citep{oquab2023dinov2}, based on the statistics released by EDM2. The validation scores are measured by comparing 50k samples and 50k ImageNet validation data.}
 \vskip 0.1in
	\centering
 \begin{tabular}{lcccccc}
		\toprule
        \multirow{2}{*}{$64\times 64$} & \multirow{2}{*}{Architecture} & \multirow{2}{*}{NFE} & \multicolumn{2}{c}{$\text{FID}_{\text{InceptionV3}}$} & \multicolumn{2}{c}{$\text{FD}_{\text{DINOv2}}$} \\
        & & & vs. Train & vs. Val & vs. Train & vs. Val \\\toprule
        \multicolumn{3}{l}{Val Data} & 1.05 & - & 13.86 & - \\\midrule
        StyleGAN-XL & GAN & 1 & 2.64 & 3.52 & 214.59 & 220.47 \\
        CTM & ADM & 1 & 1.69 & 2.88 & 159.67 & 165.96 \\
        EDM & ADM & 79 & 2.51 & 2.93 & 112.17 & 120.58 \\
        EDM2-XL & EDM2 & 63 & 1.38 & 2.29 & 70.31 & 80.53 \\
        \cc{15}PaGoDA & \cc{15}ADM & \cc{15}1 & \cc{15}1.01 & \cc{15}2.10 & \cc{15}70.04 & \cc{15}78.44 \\
                    \bottomrule
	\end{tabular}
 \end{table}%

\begin{table}[t]
\vskip -0.1in
 \scriptsize
\label{tab:fd-512x512}
\caption{Comparison on ImageNet $512\times512$. From this result, it would be interesting to experiment PaGoDA on EDM2 architecture for better performance. * the results of ImageNet $256\times256$.}
 \vskip 0.1in
 	\centering
 \begin{tabular}{lccccccc}
		\toprule
        $512\times 512$ & Arch & DM & $\#$Params & NFE & $\text{FID}_{\text{InceptionV3}}$ & $\text{FD}_{\text{DINOv2}}$ \\\toprule
        Val Data & & & & & 1.58 & 14.13 \\\midrule
        StyleGAN-XL & GAN & - & 0.2B & 1 & 2.41 & 214.88* \\
        EDM w/o CFG (teacher) & ADM & Latent & 0.3B & 63 & 7.24 & 204.10 \\
        EDM2-S w/o CFG & EDM2 & Latent & 0.3B & 63 & 2.56 & 68.64 \\
        EDM2-S w/ CFG & EDM2 & Latent & 0.3B & 63 & 2.23 & 55.23 \\
        EDM2-XXL w/ CFG & EDM2 & Latent & 1.1B & 63 & 1.81 & 33.09 \\
        \cc{15}PaGoDA w/o CFG & \cc{15}ADM & \cc{15}Pixel & \cc{15}0.3B & \cc{15}1 & \cc{15}1.80 & \cc{15}96.77 \\
                    \bottomrule
	\end{tabular}
\end{table}

\begin{figure*}[!t]
	\centering
 \includegraphics[width=\linewidth]{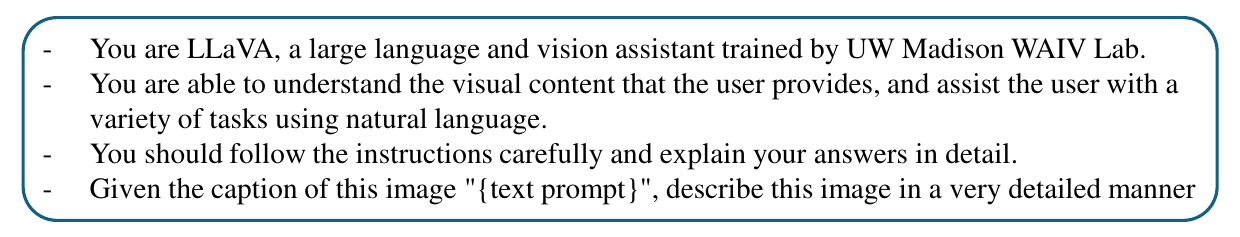}
	\caption{Input prompt of LLaVA to recaption the text-image paired data.}
    \label{fig:LLaVA}
	 \vskip -0.1in
\end{figure*}

\begin{figure*}[!t]
	\centering
 \includegraphics[width=\linewidth]{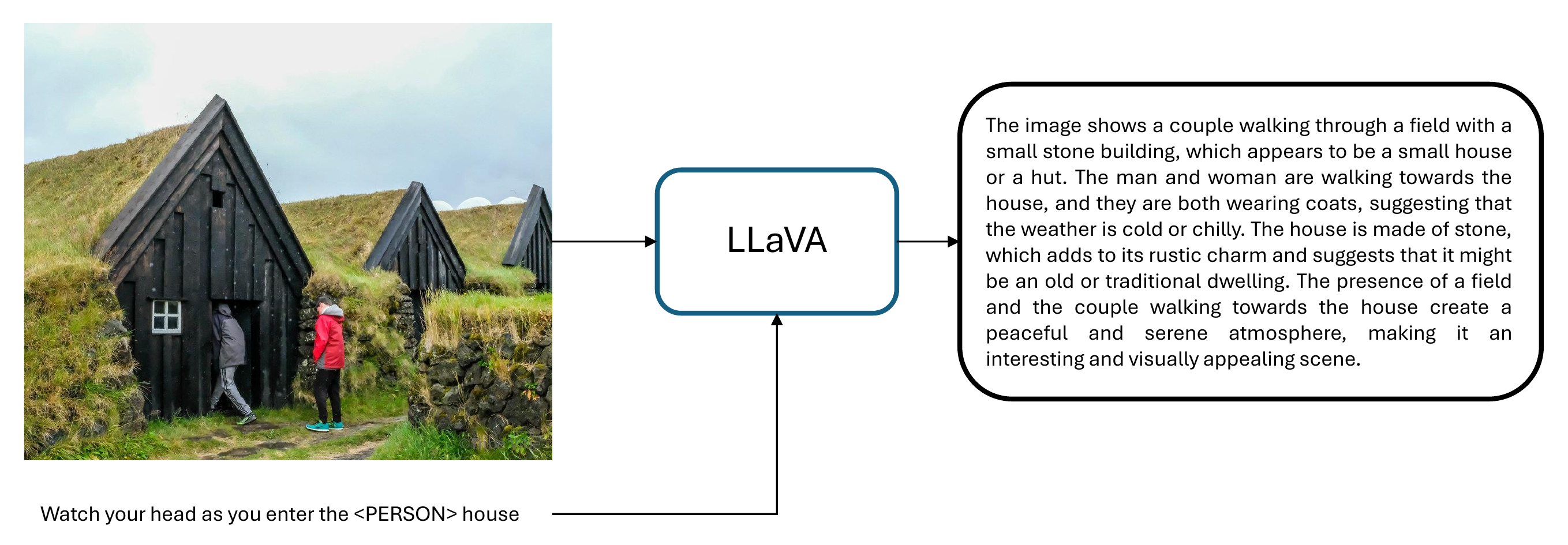}
 \vskip -0.1in
	\caption{Example of recaptioned image-text pair.}
    \label{fig:recaption}
	 \vskip -0.1in
\end{figure*}

\begin{figure*}[!t]
	\centering
 \begin{subfigure}{0.48\linewidth}
		\centering
		\includegraphics[width=\linewidth]{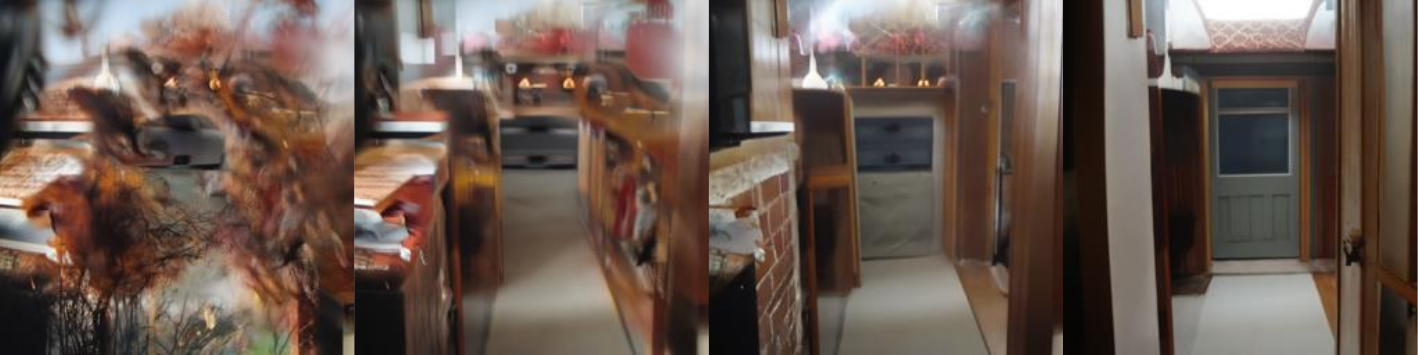}
   \vskip -0.05in
		\subcaption{DeepFloyd-IF Caption Generation}
	\end{subfigure}	
 \begin{subfigure}{0.48\linewidth}
		\centering
		\includegraphics[width=\linewidth]{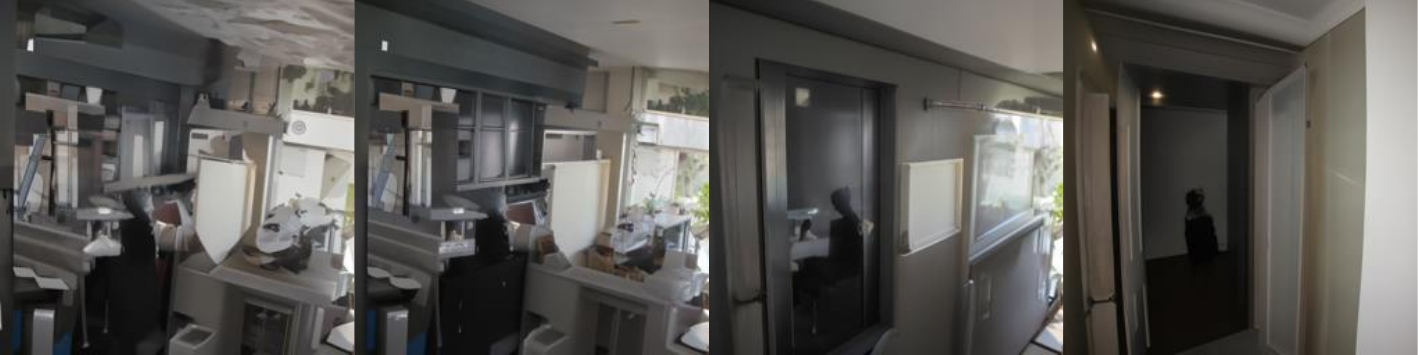}
   \vskip -0.05in
		\subcaption{PaGoDA Caption Generation}
	\end{subfigure}	
 \begin{subfigure}{0.48\linewidth}
		\centering
		\includegraphics[width=\linewidth]{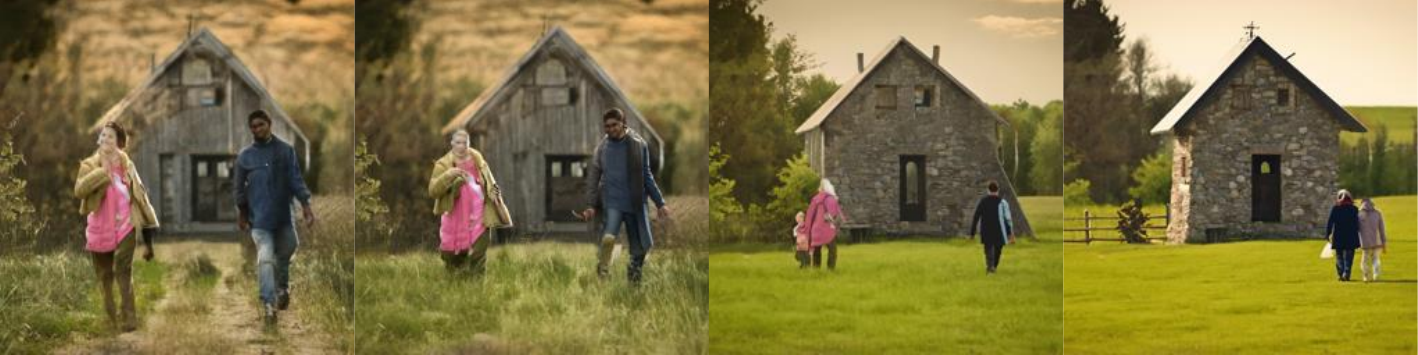}
   \vskip -0.05in
		\subcaption{DeepFloyd-IF Recaption Generation}
	\end{subfigure}	
 \begin{subfigure}{0.48\linewidth}
		\centering
		\includegraphics[width=\linewidth]{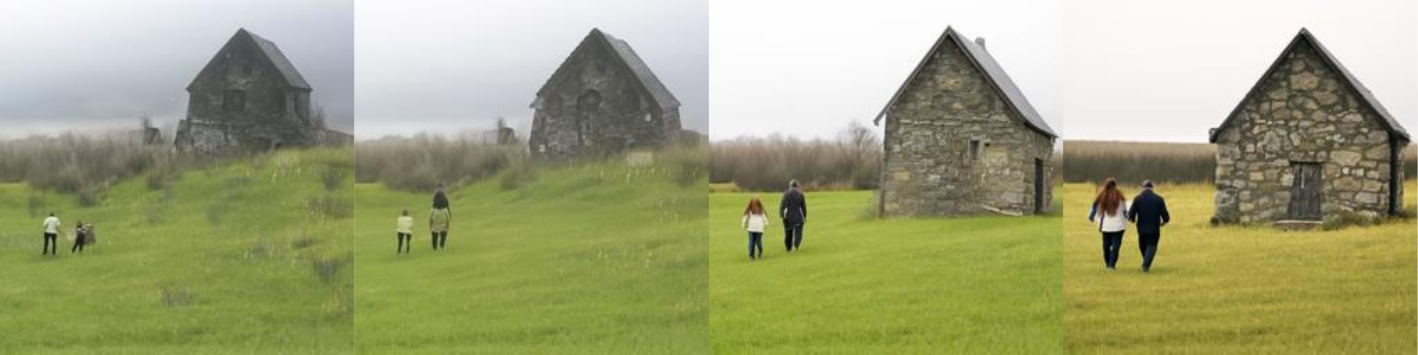}
   \vskip -0.05in
		\subcaption{PaGoDA Recaption Generation}
	\end{subfigure}	
	\caption{Caption vs. Recaption. From left to right, CFG scale increases. The caption and its corresponding recaption are given by the exemplary case in Figure~\ref{fig:recaption}.}
    \label{fig:caption_recaption}
	 \vskip -0.2in
\end{figure*}

\subsection{Text-to-Image Generation}\label{sec:t2i_exp_details}

\textbf{Dataset Construction.} Due to the presence of inappropriate contents (CSAM) in the LAION dataset~\citep{schuhmann2022laion}, we have decided to discontinue its use. Instead, we are now training our model using the CC12M~\citep{changpinyo2021cc12m} and a filtered version of COYO-700M~\citep{kakaobrain2022coyo-700m} datasets. For COYO-700M, we apply filters to select only those text-image pairs that meet specific criteria: a CLIP score (measured by ViT-B/32~\citep{Radford2021LearningTV}) above 32.0 and an aesthetic score-v2~\citep{schuhmann2022laion} higher than 5.0. Additionally, we are enhancing the dataset quality by recaptioning the original text prompts from CC12M, adopting practices similar to those used in DallE-3~\citep{betker2023improving} and PixArt-$\alpha$~\citep{chen2023pixart}. Specifically, we employ LLaVA-7B~\citep{liu2024visual}, a language model with vision assistance, to generate descriptions of the images based on the text prompts, thereby ensuring more relevant and accurate text-image pairings.

The input prompt of LLaVA is depicted in Figure~\ref{fig:LLaVA}, where we put text prompt to $\{\texttt{text prompt}\}$. The output from this recaptioning process adheres to a consistent format, typically beginning with phrases like ``\texttt{This image features ...}'' or ``\texttt{This image shows ...}''. To provide clear demonstration, Figure~\ref{fig:caption_recaption} displays several examples of original captions alongside their recaptioned counterparts.

\begin{wrapfigure}{r}{0.4\textwidth}
		\centering
		\includegraphics[width=\linewidth]{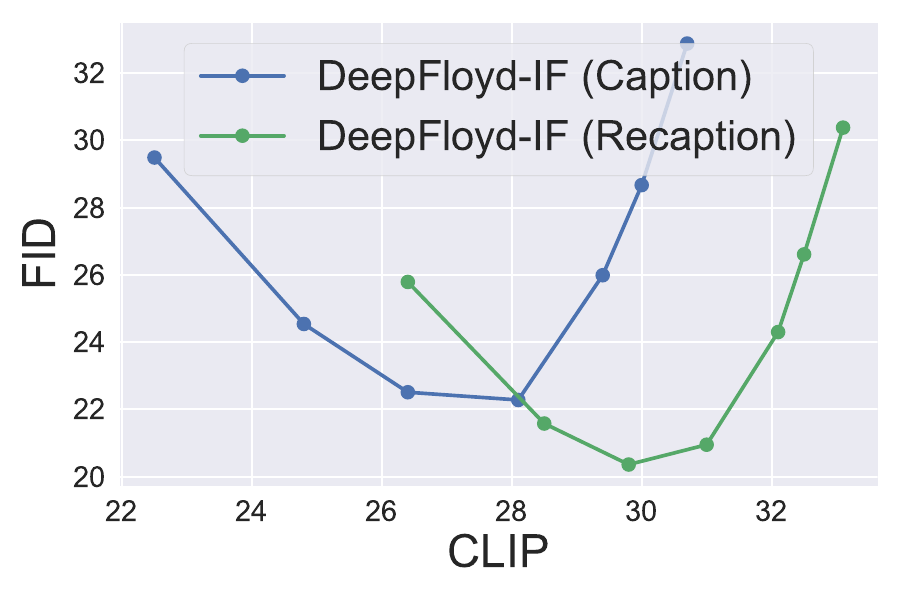}
 \vskip -0.05in
 \caption{Effect of recaptioning.}
	\label{fig:effect_recaption}
 \vskip-0.1in
\end{wrapfigure}
Interestingly, the recaptioned samples generally outperform the original caption samples. Notably, the recaptioned samples exhibit sufficient quality, particularly when the CFG scale is small, as shown in Figure~\ref{fig:effect_recaption}. Therefore, to ensure balanced generation performance across varying CFG scales, we generate samples from the original captions with the CFG scale uniformly sampled from the range $[2,10]$. For the recaptioned text, we use a CFG scale that follows a truncated Gaussian distribution on the range $[1,10]$, centered at 2 with a scale of 3. Overall, incorporating these recaptioned texts into the PaGoDA training results in only a marginal improvement in performance metrics such as FID and CLIP. However, it significantly enhances the actual quality of generation, particularly at smaller CFG scales, because the recaptioning provides better-aligned training data.

Using LLaVA, we recaption $\tilde{\mathbf{c}}(\mathbf{x},\mathbf{c})$ and obtain the DDIM latent representation, $E(\mathbf{x},\tilde{\mathbf{c}}(\mathbf{x},\mathbf{c}))$, on the entire CC12M dataset. Then, for the original text $\mathbf{c}$, we have a triplet of $(\text{image},\text{text},\text{latent})$ of $(\mathbf{x},\mathbf{c}, E(\mathbf{x},\mathbf{c}))$ for one set, and another triplet $(\mathbf{x},\mathbf{\tilde{c}}(\mathbf{x},\mathbf{c}),E(\mathbf{x},\tilde{\mathbf{c}}(\mathbf{x},\mathbf{c}))$ for recaptioned dataset. When computing $\mathcal{L}_{\text{\text{rec}}}$, we mix these triplets and randomly sample from this mixed dataset.

\begin{figure*}[t]
\vskip-0.1in
		\centering
		\includegraphics[width=\linewidth]{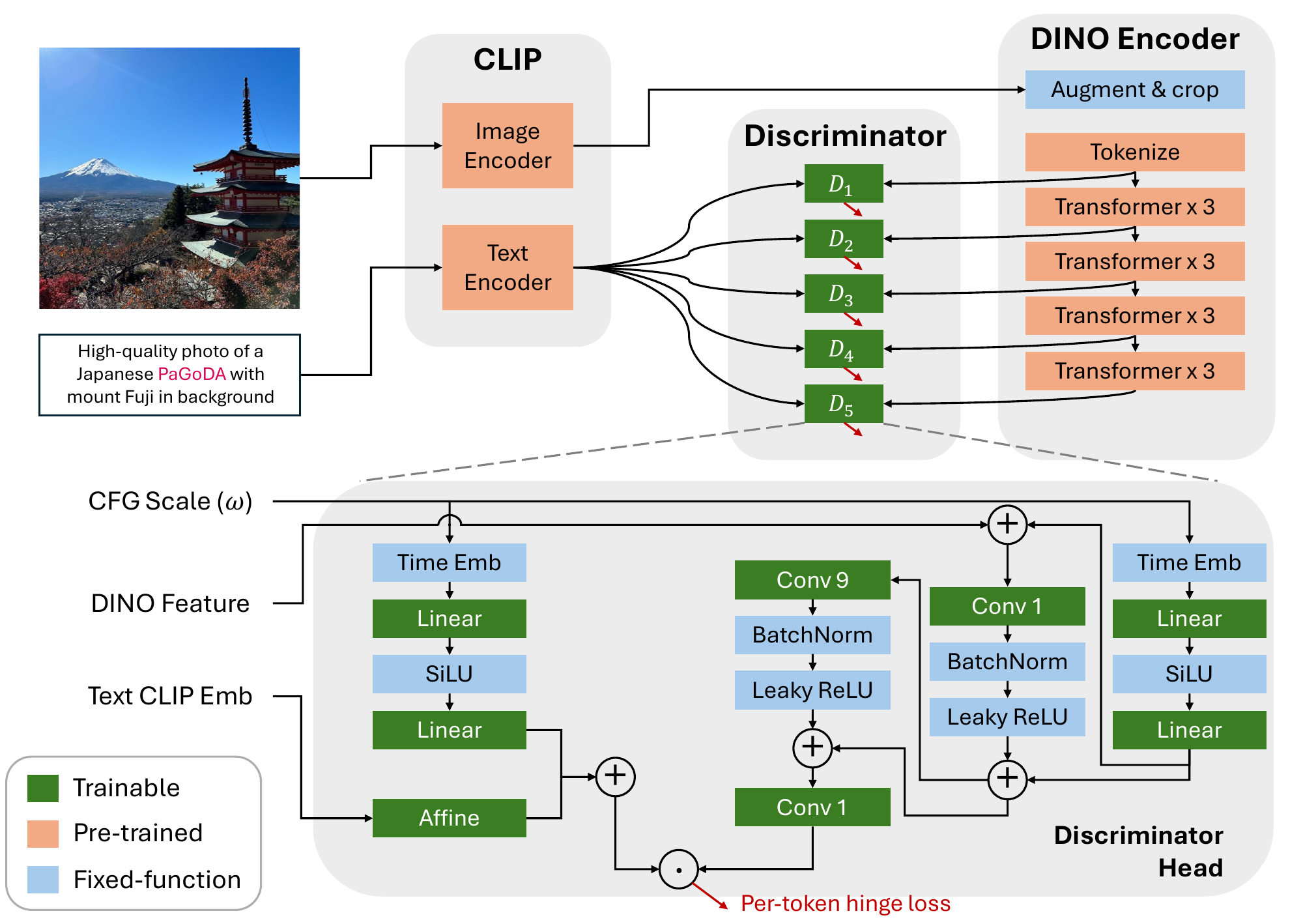}
 \vskip -0.05in
 \caption{Discriminator architecture.}
	\label{fig:discriminator_architecture}
 \vskip-0.1in
\end{figure*}
\textbf{GAN Details.} Similar to the ImageNet case, we have adopted the discriminator architecture from StyleGAN-T. In line with StyleGAN-T, we utilize the DINO ViT-S/16~\citep{caron2021emerging} as the feature extractor and apply DiffAugment~\citep{zhao2020differentiable}, incorporating \textit{Translation}, \textit{Cutout}, and \textit{Color} transformations. Building upon this architecture, we integrate a $\omega$ condition into each discriminator head, as illustrated in Figure~\ref{fig:discriminator_architecture}. The inputs for each discriminator head include the DINO feature, text CLIP embedding, and the CFG scale $\omega$, which is scaled by a factor of 100. We handle the CFG scale similarly to the time variable in traditional diffusion U-Net models, incorporating the output CFG embedding into the existing components of the StyleGAN-T discriminator head. We assume both image $\mathbf{x}$ and text $\mathbf{c}$ are related with the CFG scale, thus we designed the discriminator to incorporate $\omega$ information into both modules, enhancing the relevance and contextuality of the discrimination process.

\textbf{Reconstruction Details.} In our text-to-image training, we largely adhere to the protocols established for ImageNet training. However, a notable modification involves the decoder network, which now incorporates a $\omega$ condition as an auxiliary input. Crucially, this $\omega$ condition is processed in decoder in the same way as the time condition in diffusion models. We achieve this by scaling $\omega$ by a factor of 100, thus aligning it with the existing time ranges. This method ensures a consistent treatment of the $\omega$ parameter, integrating it smoothly into the established model architecture.

\begin{wrapfigure}{r}{0.4\textwidth}
\vskip-0.2in
		\centering
		\includegraphics[width=\linewidth]{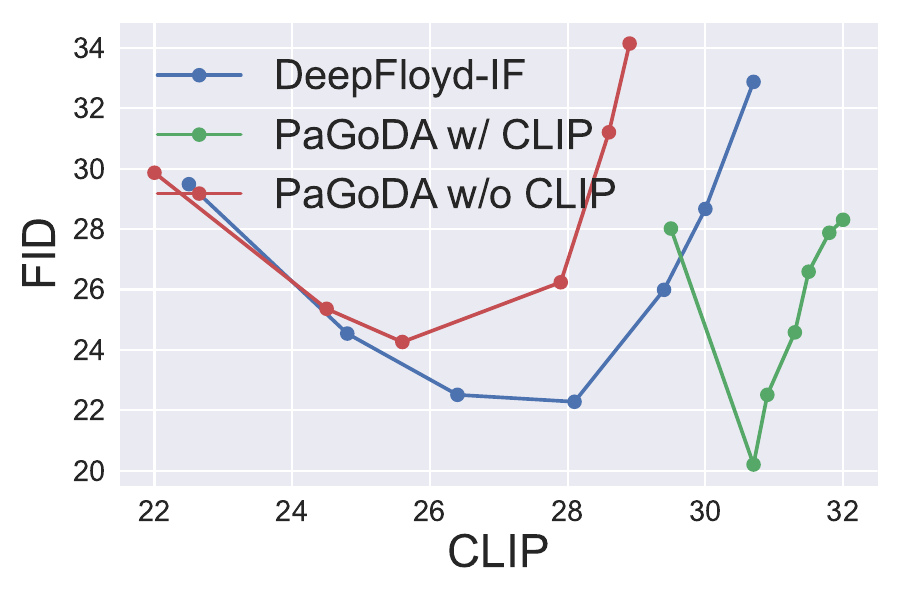}
 \vskip -0.05in
 \caption{Effect of CLIP regularization.}
	\label{fig:CLIP}
 \vskip -0.5in
\end{wrapfigure}

\textbf{CLIP Details.} Neither the reconstruction loss nor the GAN loss directly models or maximizes the text-image correlation. To address this, we introduce an additional text-image alignment metric to train our model. Specifically, we employ ViT-L/14~\citep{Radford2021LearningTV} to assess the CLIP value. This regularization significantly enhances PaGoDA's performance, as evidenced in Figure~\ref{fig:CLIP} by improving both FID and CLIP scores. These enhancements suggest that not only is the sample quality improving, but also the alignment between text and images is becoming more accurate.

For the training, we use the AdamW8bit optimizer \citep{dettmers20218} to minimize the required memory with learning rate of 1e-5 for both decoder and discriminator. Similar to the ImageNet experiment, we do not apply the weight decay. In this text-to-image experiment, we do not use EMA, following previous works~\citep{saharia2022photorealistic}. In the base resolution, we use the adaptive weighting with $\lambda=4\frac{\Vert \nabla_{\bm{\theta}^{l}}\mathcal{L}_{\text{\text{rec}}} \Vert_{2}^{2}}{\Vert\nabla_{\bm{\theta}^{l}}\mathcal{L}_{\text{adv}}\Vert_{2}^{2}}$. Overall, we use the DeepFloyd-IF-I model with 0.9B number of parameters.

\begin{figure*}[!t]
	\centering
 \begin{subfigure}{0.4\linewidth}
		\centering
 \includegraphics[width=\linewidth]{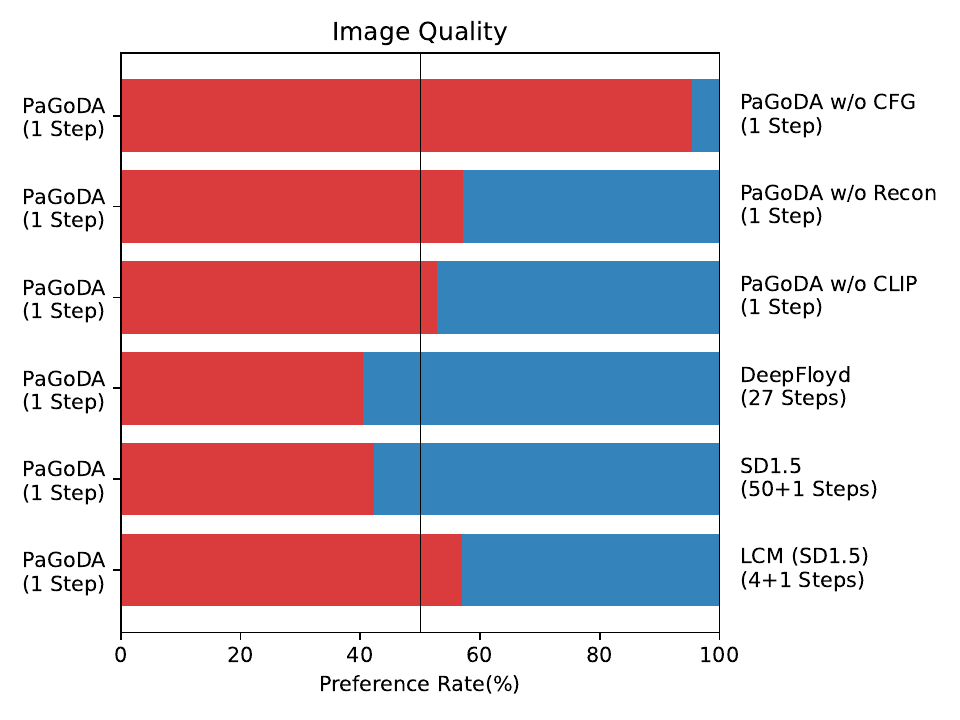}
 \end{subfigure}
 \begin{subfigure}{0.4\linewidth}
		\centering
 \includegraphics[width=\linewidth]{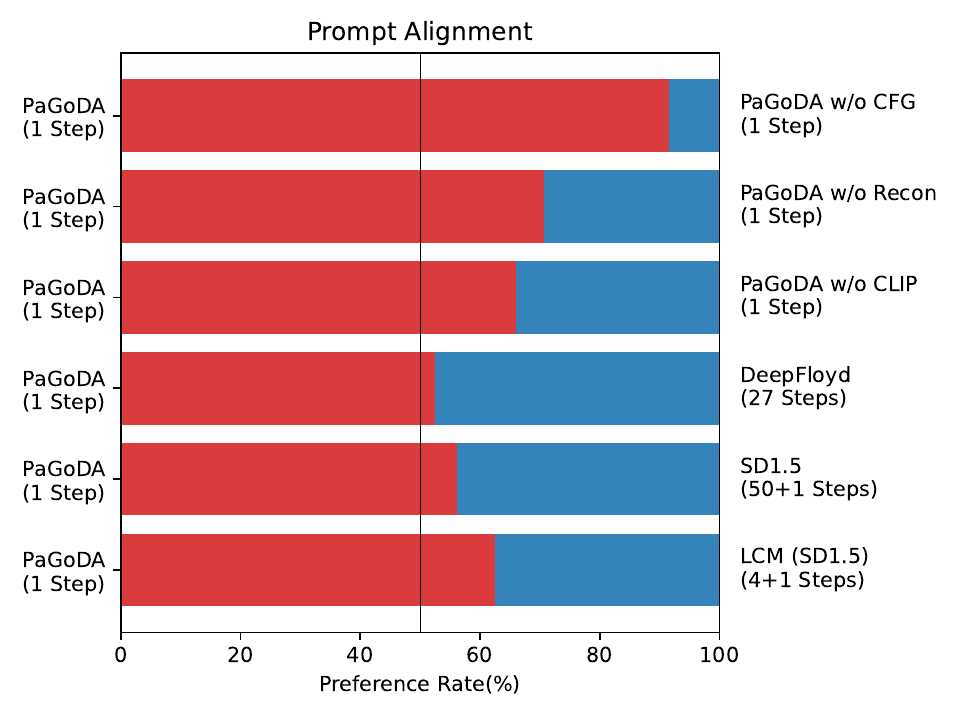}
 \end{subfigure}
	 \caption{Human evaluation result on T2I with CFG set to be 7 across models.}
  \label{fig:human_eval}
\end{figure*}

Figure~\ref{fig:human_eval} compares PaGoDA with the existing baselines.

\section{Theoretical Analysis}\label{sec:theory_proof}
In this section, we present rigorous statements and proofs of all theorems. The theorems are shown for the unconditional generation case (i.e., without the condition $\mathbf{c}$), but the analysis can be extended to the conditional scenario.

\subsection{Convergence with PaGoDA's Reconstruction Loss}\label{sec:conv_recon}
In Section~\ref{subsec:pre_conv}, we introduce the necessary notations and preliminaries. In Sections~\ref{subsec:W_2_conv_analysis} and \ref{subsec:W_1_conv_analysis}, we demonstrate that the Wasserstein-1 and Wasserstein-2 discrepancies of the learned density (with PaGoDA's reconstruction loss) from $p_{\text{\text{data}}}$ are upper bounded by PaGoDA's reconstruction loss and the pre-trained DM's training error. All results are proved for unconditional generation (i.e., without $\mathbf{c}$ as an input), but they can be easily generalized to the conditional case.

\subsubsection{Preliminaries of Convergence Analysis}\label{subsec:pre_conv}
Consider OU process for $t\in[0,T]$, where $T>0$:
\begin{align*}
    \diff\mathbf{x}_t = -f(t)\mathbf{x}_t\diff t + g(t) \diff\mathbf{w}_t
\end{align*}
Its associated PF-ODE is
\begin{align*}
    \diff\mathbf{x}_t = \big[-f(t)\mathbf{x}_t - \frac{1}{2}g^2(t)\nabla\log p_t (\mathbf{x}_t)\big]\diff t.
\end{align*}

We consider $f(t)\equiv 1$ and $g(t)\equiv\sqrt{2}$ for simplicity. That is,
\begin{equation}\label{eq:OU_simple}
    \diff\mathbf{x}_t = -\mathbf{x}_t\diff t + \sqrt{2} \diff\mathbf{w}_t.
\end{equation}

We recall that PaGoDA's reconstruction loss (unconditional case) is defined as:
\begin{equation*}
    \mathcal{L}_{\text{\text{rec}}}({\bm{\theta}};{\bm{\phi}_0}):=\mathbb{E}_{p_{\text{\text{data}}}(\mathbf{x})p_{{\bm{\phi}_0}}(\mathbf{z}\vert\mathbf{x})}\Big[\big\Vert \mathbf{x}-G_{\bm{\theta}}^{T\rightarrow0}(\mathbf{z}) \big\Vert_{2}^{2}\Big],
\end{equation*}
Here, we use $p_{\bm{\phi}_0}(\mathbf{z} \vert \mathbf{x})$ to denote the density obtained by solving the pre-trained teacher DM's empirical PF-ODE forward in time from $t=0$ to $t=T$:
\begin{align*}
\diff \mathbf{x}_t = \big[-f(t)\mathbf{x}_t - \frac{1}{2}g^2(t) \mathbf{s}_{{\bm{\phi}_0}} (\mathbf{x}_t, t)\big] \diff t,
\end{align*}
where $\mathbf{s}_{\phi} (\mathbf{x}_t, t)$ indicates the pre-trained DM. We remark that $p_{{\bm{\phi}_0}}(\mathbf{z} \vert \mathbf{x})$ defines a deterministic process.

We take $p_{\text{\text{prior}}}:=\mathcal{N}\big(\mathbf{0}, (1-e^{-2T})\mathbf{I}\big)$ as the prior distribution, and define $p_{T, {{\bm{\phi}_0}}}:=G_{{\bm{\phi}_0}}^{0\rightarrow T}\sharp p_{\text{\text{data}}}$ as the distribution obtained by solving the teacher-determined empirical PF-ODE forward in time. Let us consider the density obtained by sampling from PaGoDA (trained without GAN) $p_{0,{\bm{\theta}}}:=G_{\bm{\theta}}^{T\rightarrow0} \sharp p_{\text{\text{prior}}}$. We also let $G^{T\rightarrow0}$ denote the ground truth transition map from $T$ to $0$, defined by the PF-ODE.

Conceptually, Theorems~\ref{th:w_2_convergence} and \ref{th:w_1_convergence_rough} demonstrate that 
\begin{align*}
    W_p(p_{0,{\bm{\theta}}}, p_{\text{\text{data}}}) \lesssim \mathcal{L}_{\text{\text{rec}}}({\bm{\theta}};{{\bm{\phi}_0}}) + \epsilon_{\text{DM}}, \quad p=1,2.
\end{align*}
This implies that training with PaGoDA's reconstruction loss ensures the learned density $p_{0,\bm{\theta}} = G_{\bm{\theta}}^{T \rightarrow 0} \sharp p_{\text{\text{prior}}}$ is close to $p_{\text{\text{data}}}$ in Wasserstein distance sense. Moreover, improving the teacher DM to reduce the error $\epsilon_{\text{DM}}$ is a way to further decrease the discrepancy between $p_{0,\bm{\theta}}$ and $p_{\text{\text{data}}}$.

We remark that the differences between the two theorems primarily lie in the distinct smoothness assumptions on $p_{\text{\text{data}}}$.


\subsubsection{$W_2$ Bound with PaGoDA's Reconstruction Loss}\label{subsec:W_2_conv_analysis}

\begin{assumpA}\label{th:assumption}
    \begin{enumerate}[(i)]
        \item $m^2:=\mathbb{E}_{p_{\text{\text{data}}}(\mathbf{x})}\norm{\mathbf{x}}_2^2<\infty$;
        \item There is a $\epsilon_{\text{DSM}}>0$ so that $\sup_{\mathbf{x}, t}\norm{\mathbf{s}_{{\bm{\phi}_0}}(\mathbf{x}, t) - \nabla\log 
        p_t(\mathbf{x})}_2^2\leq \epsilon_{\text{DM}}^2$;
        \item $G_{\bm{\theta}}^{T\rightarrow 0}$ is Lipschitz in $\mathbf{x}$: 
        \begin{align*}
            \Lambda:=\sup_{\mathbf{x}\neq\mathbf{y}} \frac{\norm{G_{\bm{\theta}}^{T\rightarrow 0}(\mathbf{x})-G_{\bm{\theta}}^{T\rightarrow 0}(\mathbf{y})}_2}{\norm{\mathbf{x}-\mathbf{y}}_2}<\infty,
        \end{align*}
        for all $\bm{\theta}$ and $T$.
        \item $\log p_{\text{\text{data}}}$ is $\gamma$-strongly concave with $\gamma > 3/2$:
        \begin{align*}
            \inner{\mathbf{x}- \mathbf{y}}{\nabla \log p_{\text{\text{data}}}(\mathbf{x}) - \nabla \log p_{\text{\text{data}}}(\mathbf{y})} \leq - \gamma \norm{\mathbf{x}- \mathbf{y}}_2^2,
        \end{align*}
        for all $\mathbf{x}$ and $\mathbf{y}$.
    \end{enumerate}
\end{assumpA}


\begin{theorem}\label{th:w_2_convergence} Given that Assumption~\ref{th:assumption} holds, suppose $\delta$ is a positive constant such that $\delta < \frac{e^{-2T}}{3-e^{-2T}}$, and let $h(\gamma, T):= \frac{\gamma}{e^{-2T}+\gamma(1-e^{-2T})} -(1+\delta)$, where it is noted that $h(\gamma, T)$ is also positive. Then
    \begin{align*}
        W_2(p_{0,{\bm{\theta}}}, p_{\text{\text{data}}})      \leq &~\mathcal{L}_{\text{\text{rec}}}({\bm{\theta}};{{\bm{\phi}_0}}) +\Big[\mathbb{E}_{p_{\text{\text{data}}}(\mathbf{x})p_{{\bm{\phi}_0}}(\mathbf{z}\vert\mathbf{x})}\big\Vert \mathbf{x}-G^{T\rightarrow0}(\mathbf{z}) \big\Vert_{2}^{2}\Big]^{\frac{1}{2}}
    \\ &+ \big(\Lambda +e^{-\frac{1}{2}h(\gamma, T) T}\big) W_2\big(p_{T}, p_{T,{{\bm{\phi}_0}}} \big)
    \\ &+ \frac{\epsilon_{\text{DM}}}{\sqrt{2\delta h(\gamma,T)}} \big(1-e^{-h(\gamma, T)T}\big)^{\frac{1}{2}} + e^{-\frac{T}{2}} m\Lambda.
    \end{align*}
In particular, if we assume Assumption~\ref{th:assumption} (iii) holds also for $G^{T\rightarrow 0}$, 
\begin{align*}
    W_2(p_{0,{\bm{\theta}}}, p_{\text{\text{data}}}) \lesssim \mathcal{L}_{\text{\text{rec}}}({\bm{\theta}};{{\bm{\phi}_0}}) + \epsilon_{\text{DM}} + e^{-\frac{T}{2}} m\Lambda.
\end{align*}
Here, we use $\lesssim$ to absorb the dependence on the constants $T$ and $\gamma$ into the inequality.
\end{theorem}

We present an inequality which is essential for the proof of Theorem~\ref{th:w_2_convergence}.

\begin{lemma}[Proposition 3.5. in \cite{saumard2014log}]\label{th:strong_concave}
    Let $P$ and $Q$ be two distributions on $\mathbb{R}^D$. Suppose that $\log P$ is $\gamma_P$-concave and $\log Q$ is $\gamma_Q$-concave. Then the convolution of $\log P*Q$ is a $(1/\gamma_P+1/\gamma_Q)^{-1}$-concave distribution.
\end{lemma}

\begin{myproof}{Theorem}{\ref{th:w_2_convergence}}
The proof of the theorem is inspired by \citep{tang2024contractive,lyu2023sampling}.
Define $p_{T, {{\bm{\phi}_0}}}:=G_{{\bm{\phi}_0}}^{0\rightarrow T}\sharp p_{\text{\text{data}}}$, and $p_{0, {{\bm{\phi}_0}}}:=G^{T\rightarrow0} \sharp p_{T, {{\bm{\phi}_0}}}$. From the triangle inequality, we have
\begin{align*}
    W_2(p_{0,{\bm{\theta}}}, p_{\text{\text{data}}}) \leq \underset{(A)}{\underbrace{W_2(p_{0,{\bm{\theta}}}, p_{0,{{\bm{\phi}_0}}}) }}+ \underset{(B)}{\underbrace{W_2(p_{0,{{\bm{\phi}_0}}}, p_{\text{\text{data}}})}}.
\end{align*}
For (A), let $\pi(\mathbf{y}, \mathbf{z})\in\Pi\big(p_{\text{\text{prior}}}, p_{T,{{\bm{\phi}_0}}} \big)$  be a coupling of $p_{\text{\text{prior}}}$ and $p_{T,{{\bm{\phi}_0}}}$. Then
\begin{align*}
    (A) &= W_2\big(G_{\bm{\theta}}^{T\rightarrow0} \sharp p_{\text{\text{prior}}},  G^{T\rightarrow0} \sharp p_{T, {{\bm{\phi}_0}}}\big)
    \\ &\leq \Big(\mathbb{E}_{(\mathbf{y}, \mathbf{z})\sim\pi}\norm{G_{\bm{\theta}}^{T\rightarrow0}(\mathbf{y})-G^{T\rightarrow0}(\mathbf{z})}_2^2 \Big)^{\frac{1}{2}}
    \\ &\leq \underset{(A.1)}{\underbrace{\Big(\mathbb{E}_{(\mathbf{y}, \mathbf{z})\sim\pi}\norm{G_{\bm{\theta}}^{T\rightarrow0}(\mathbf{y})-G_{\bm{\theta}}^{T\rightarrow0}(\mathbf{z})}_2^2 \Big)^{\frac{1}{2}}}} +  \underset{(A.2)}{\underbrace{\Big(\mathbb{E}_{(\mathbf{y}, \mathbf{z})\sim\pi}\norm{G_{{\bm{\theta}}}^{T\rightarrow0}(\mathbf{z})-G^{T\rightarrow0}(\mathbf{z})}_2^2 \Big)^{\frac{1}{2}}}}.
\end{align*}
For (A.1), we can yield
\begin{align}\label{eq:A.1_result}
    (A.1) &\leq\Lambda\min_{\pi\in\Pi\big(p_{\text{\text{prior}}}, p_{T,{{\bm{\phi}_0}}}  \big)}\Big(\mathbb{E}_{(\mathbf{y}, \mathbf{z})\sim\pi}\norm{\mathbf{y}-\mathbf{z}}_2^2 \Big)^{\frac{1}{2}} \nonumber
    \\ &=\Lambda W_2\big(p_{\text{\text{prior}}}, p_{T,{{\bm{\phi}_0}}} \big) \nonumber
    \\ & \leq \Lambda W_2\big(p_{\text{\text{prior}}}, p_{T} \big) + \Lambda W_2\big(p_{T}, p_{T,{{\bm{\phi}_0}}} \big)\nonumber
    \\ & \leq e^{-\frac{T}{2}} \big(\mathbb{E}_{p_{\text{\text{data}}}(\mathbf{x}_0)}\norm{\mathbf{x}_0}_2^2\big)^{\frac{1}{2}}\Lambda  + \Lambda W_2\big(p_{T}, p_{T,{{\bm{\phi}_0}}} \big).
\end{align}
Here, the last inequality is a consequence of the following bound
\begin{align*}
    W_2(p_{\text{\text{prior}}}, p_{T})  \leq e^{-\frac{T}{2}} \big(\mathbb{E}_{p_{\text{\text{data}}}(\mathbf{x}_0)}\norm{\mathbf{x}_0}_2^2\big)^{\frac{1}{2}},
\end{align*}
which holds because $p_{\text{\text{prior}}}$ is taken as $\mathcal{N}\big(\mathbf{0}, (1-e^{-2T})\mathbf{I}\big)$, and $\mathbf{x}_T\sim p_T$ governed by Eq.~\eqref{eq:OU_simple} admits the expression
\begin{align*}
    \mathbf{x}_T = e^{-T}\mathbf{x}_0 + \int_0^T e^{-(T-s)} \sqrt{2} \diff\mathbf{w}_s = e^{-T}\mathbf{x}_0 + \mathbf{z}, \quad \mathbf{z}\sim\mathcal{N}\big(\mathbf{0}, (1-e^{-2T})\mathbf{I}\big). 
\end{align*}
For (A.2), since $p_{T,{{\bm{\phi}_0}}}(\mathbf{z})=\int p_{{\bm{\phi}_0}}(\mathbf{z}|\mathbf{x})p_{\text{\text{data}}}(\mathbf{x})\diff\mathbf{x}$, by applying Minkowski inequality we have
\begin{align}\label{eq:A.2_result}
    (A.2) &=\Big(\mathbb{E}_{(\mathbf{y}, \mathbf{z})\sim\pi}\norm{G_{{\bm{\theta}}}^{T\rightarrow0}(\mathbf{z})-G^{T\rightarrow0}(\mathbf{z})}_2^2 \Big)^{\frac{1}{2}}  \nonumber
    \\& = \Big(\mathbb{E}_{ \mathbf{z}\sim \nonumber p_{T,{{\bm{\phi}_0}}}(\mathbf{z})}\norm{G_{{\bm{\theta}}}^{T\rightarrow0}(\mathbf{z})-G^{T\rightarrow0}(\mathbf{z})}_2^2 \Big)^{\frac{1}{2}} 
   \\ &\leq \Big(\mathbb{E}_{ p_{\text{\text{data}}}(\mathbf{x}) \nonumber p_{{\bm{\phi}_0}}(\mathbf{z}|\mathbf{x})}\norm{G_{{\bm{\theta}}}^{T\rightarrow0}(\mathbf{z})-\mathbf{x}}_2^2 \Big)^{\frac{1}{2}} + \Big(\mathbb{E}_{ p_{\text{\text{data}}}(\mathbf{x}) p_{{\bm{\phi}_0}}(\mathbf{z}|\mathbf{x})}\norm{\mathbf{x}-G^{T\rightarrow0}(\mathbf{z})}_2^2 \Big)^{\frac{1}{2}} 
   \\ &= \mathcal{L}_{\text{\text{rec}}}({\bm{\theta}};{{\bm{\phi}_0}})+\Big[\mathbb{E}_{p_{\text{\text{data}}}(\mathbf{x})p_{{\bm{\phi}_0}}(\mathbf{z}\vert\mathbf{x})}\big\Vert \mathbf{x}-G^{T\rightarrow0}(\mathbf{z}) \big\Vert_{2}^{2}\Big]^{\frac{1}{2}}.
\end{align}
The proof for (B) is motivated by \citep{tang2024contractive}. Consider the following two reverse time PF-ODEs on the interval $[0, T]$
\begin{align*}
      \frac{\diff \hat{\mathbf{z}}_{t,{{\bm{\phi}_0}}}}{\diff t} =\hat{\mathbf{z}}_{t,{{\bm{\phi}_0}}} +\mathbf{s}_{{{\bm{\phi}_0}}}(\hat{\mathbf{z}}_{t,{{\bm{\phi}_0}}}, T-t), \quad \hat{\mathbf{z}}_{0,{{\bm{\phi}_0}}}\sim p_{T,{{\bm{\phi}_0}}}
\end{align*}
and
\begin{align*}
      \frac{\diff \hat{\mathbf{z}}_{t}}{\diff t} =\hat{\mathbf{z}}_t +\nabla\log p_{T-t} (\hat{\mathbf{z}}_{t}), \quad 
      \hat{\mathbf{z}}_{0}\sim p_{T},
\end{align*}
with a coupling of $\hat{\mathbf{z}}_{0,{{\bm{\phi}_0}}}\sim p_{T,{{\bm{\phi}_0}}}$ and $\hat{\mathbf{z}}_{0}\sim p_{T}$ so that $W_2^2(p_{T,{{\bm{\phi}_0}}}, p_{T})=\mathbb{E}\norm{\hat{\mathbf{z}}_{0,{{\bm{\phi}_0}}} - \hat{\mathbf{z}}_0}_2^2$. We notice that
$W_2^2(p_{0,{{\bm{\phi}_0}}}, p_{\text{\text{data}}})  \leq \mathbb{E}\norm{\hat{\mathbf{z}}_{T,{{\bm{\phi}_0}}}- \hat{\mathbf{z}}_T}_2^2$. Thus, we need to obtain a upper bound of $\mathbb{E}\norm{\hat{\mathbf{z}}_{T,{{\bm{\phi}_0}}}- \hat{\mathbf{z}}_T}_2^2$. Let $u(t):=\mathbb{E}\norm{\hat{\mathbf{z}}_{t,{{\bm{\phi}_0}}}- \hat{\mathbf{z}}_t}_2^2$. Then
\begin{align}\label{eq:diff_u_t}
    \frac{\diff}{\diff t} u(t) =~& 2\mathbb{E}\inner{\hat{\mathbf{z}}_{t,{{\bm{\phi}_0}}}- \hat{\mathbf{z}}_t}{\frac{\diff}{\diff t} \big(\hat{\mathbf{z}}_{t,{{\bm{\phi}_0}}}- \hat{\mathbf{z}}_t\big)} \nonumber
    \\ =~& 2u(t) + 2 \mathbb{E}\Big[\inner{\hat{\mathbf{z}}_{t,{{\bm{\phi}_0}}}- \hat{\mathbf{z}}_t}{\mathbf{s}_{{{\bm{\phi}_0}}}(\hat{\mathbf{z}}_{t,{{\bm{\phi}_0}}}, T-t)-\nabla\log p_{T-t} (\hat{\mathbf{z}}_{t})}\Big] \nonumber
    \\ =~& 2u(t) + 2 \underset{(B.1)}{\underbrace{\mathbb{E}\Big[\inner{\hat{\mathbf{z}}_{t,{{\bm{\phi}_0}}}- \hat{\mathbf{z}}_t}{\mathbf{s}_{{{\bm{\phi}_0}}}(\hat{\mathbf{z}}_{t,{{\bm{\phi}_0}}}, T-t)- \nabla\log p_{T-t} (\hat{\mathbf{z}}_{t,{{\bm{\phi}_0}}})} \Big]}} \nonumber
    \\ ~&  + 2 \underset{(B.2)}{\underbrace{\mathbb{E}\Big[\inner{\hat{\mathbf{z}}_{t,{{\bm{\phi}_0}}}- \hat{\mathbf{z}}_t}{\nabla\log p_{T-t} (\hat{\mathbf{z}}_{t,{{\bm{\phi}_0}}})-\nabla\log p_{T-t} (\hat{\mathbf{z}}_{t})}\Big]}}. 
\end{align}
Let $\delta>0$, by applying Yang's inequality $ab=\big(\sqrt{2\delta}a \big) \big(\frac{b}{\sqrt{2\delta}}\big)\leq \delta a^2 + \frac{b^2}{4\delta}$ to (B.1) for nonnegative $a$ and $b$, and the Assumption~\ref{th:assumption}, it becomes
\begin{align}\label{eq:B_1}
    (B.1) \leq~& \delta u(t)+\frac{\epsilon_{\text{DM}}^2}{4\delta} .
\end{align}

We turn our attention to (B.2). Naively, (B.2) may be naively bounded above by $\text{Lip}\big(\nabla\log p_t(\cdot)\big) u(t)$, where $\text{Lip}\big(\nabla\log p_t(\cdot)\big)$ is the Lipschitz constant of $\nabla\log p_t(\cdot)$ in $\mathbf{x}$. However, we will now derive a sharper bound by incorporating assumptions on the data distribution.

We notice that $p_t(\mathbf{x}_t) = \int p_{t|0}(\mathbf{x}_t|\mathbf{x}_0)p_{\text{\text{data}}}(\mathbf{x}_0) \diff\mathbf{x}_0$, where $p_{t|0}(\mathbf{x}_t|\mathbf{x}_0)=\mathcal{N}\big(\mathbf{x}_t; e^{-t}\mathbf{x}_0, (1-e^{-2t})\mathbf{I} \big)$ is a transition kernel from $0$ to $t$ determined by the forward SDE. Therefore, expressing $p_t$ in convolution form, under Assumption~\ref{th:assumption}, and leveraging Lemma~\ref{th:strong_concave}, we deduce that  $\log p_{T-t}$ is a  $\gamma/\big(e^{-2(T-t)}+\gamma(1-e^{-2(T-t)}) \big)$-strongly concave distribution (see \citep{gao2023wasserstein}). Hence, 
\begin{align}\label{eq:B_2}
    (B.2)\leq -\frac{\gamma}{e^{-2(T-t)}+\gamma(1-e^{-2(T-t)})} u(t).
\end{align}
With the inequalities~\eqref{eq:B_1} and \eqref{eq:B_2}, we deduce from Eq.~\eqref{eq:diff_u_t} that
\begin{align*}
    u'(t) \leq a(t) u(t)+\frac{\epsilon_{\text{DM}}^2}{2\delta}, \quad\text{where}\quad a(t):=\Big(2+2\delta-\frac{2\gamma}{e^{-2(T-t)}+\gamma(1-e^{-2(T-t)})}\Big).
\end{align*}
By applying Grönwall's inequality, we obtain
\begin{align}\label{eq:gronwall}
    \mathbb{E}\norm{\hat{\mathbf{z}}_{T,{{\bm{\phi}_0}}}- \hat{\mathbf{z}}_T}_2^2& \leq e^{A(T)}\mathbb{E}\norm{\hat{\mathbf{z}}_{0,{{\bm{\phi}_0}}}- \hat{\mathbf{z}}_0}_2^2 + \frac{\epsilon^2_{\text{DM}}}{2\delta} \int_0^T e^{A(T)-A(t)}\diff t, \nonumber
    \\ & = e^{A(T)}W_2^2(p_{T,{{\bm{\phi}_0}}}, p_{T}) + \frac{\epsilon^2_{\text{DM}}}{2\delta} \int_0^T e^{A(T)-A(t)}\diff t.
\end{align}
where $A(t):=\int_0^t a(s)\diff s$. 

We aim to find an upper bound for inequality \eqref{eq:gronwall} that decays exponentially with respect to $T$.
In $a(t)$, $b(t):=\frac{\gamma}{e^{-2(T-t)}+\gamma(1-e^{-2(T-t)})}$ as a function of $t$ has the derivative as $\frac{2\gamma(\gamma-1)e^{-2(T-t)}}{\big(e^{-2(T-t)}+\gamma ( 1- e^{-2(T-t)})\big)^2}$. This implies when $\gamma \geq 1$, $b$'s minimum occurs at $b(0)=\frac{\gamma}{\gamma + e^{-2T}(1-\gamma)}$, which implies $a(t)\leq 2\big(1+\delta -b(0)\big)$ for all $t\in[0,T]$. Setting $\delta < \frac{e^{-2T}}{3-e^{-2T}}$, which implies $\frac{1}{2}>\frac{\delta}{(1+\delta) e^{-2T}-\delta}$, then $\gamma>\frac{3}{2}=1 +\frac{1}{2}>1+\frac{\delta}{(1+\delta) e^{-2T}-\delta}$ (notice that $(1+\delta)e^{-2T}-\delta > 2\delta$), we can deduce that 
\begin{align*}
    a(t) \leq 1+\delta - \frac{\gamma}{e^{-2T}+\gamma(1-e^{-2T})}<0.
\end{align*}
Let $h(\gamma, T):= \frac{\gamma}{e^{-2T}+\gamma(1-e^{-2T})} -(1+\delta)>0$. Then we establish that $a(t)\leq -h(\gamma, T)$, $A(T)\leq -h(\gamma, T) T$, and $A(T)-A(t)\leq -h(\gamma, T) t$ which implies $\int_{0}^{T}e^{A(T)-A(t)}\diff t \leq 1-e^{-h(\gamma, T)T}$. By applying the above bounds and inequality~\eqref{eq:gronwall}, (B) becomes
\begin{align}\label{eq:B_result}
    (B)&\leq \big(\mathbb{E}\norm{\hat{\mathbf{z}}_{T,{{\bm{\phi}_0}}}- \hat{\mathbf{z}}_T}_2^2 \big)^{\frac{1}{2}}\nonumber 
    \\ & \leq \bigg(e^{-h(\gamma, T) T}W_2^2(p_{T,{{\bm{\phi}_0}}}, p_{T}) + \frac{\epsilon^2_{\text{DM}}}{2\delta h(\gamma,T)} \big(1-e^{-h(\gamma, T)T}\big) \bigg)^{\frac{1}{2}} \nonumber 
    \\ & \leq e^{-\frac{1}{2}h(\gamma, T) T}W_2(p_{T,{{\bm{\phi}_0}}}, p_{T}) + \frac{\epsilon_{\text{DM}}}{\sqrt{2\delta h(\gamma,T)}} \big(1-e^{-h(\gamma, T)T}\big)^{\frac{1}{2}}.
\end{align}

Here, the last inequality is from a simple inequality $\sqrt{a+b}\leq\sqrt{a}+\sqrt{b}$ for nonnegative $a$ and $b$.

By combining inequalities \eqref{eq:A.1_result}, \eqref{eq:A.2_result}, and \eqref{eq:B_result}, we obtain
\begin{align*}
    W_2(p_{0,{\bm{\theta}}}, p_{\text{\text{data}}})  
     \leq &~
    e^{-\frac{T}{2}} \big(\mathbb{E}_{p_{\text{\text{data}}}(\mathbf{x}_0)}\norm{\mathbf{x}_0}_2^2\big)^{\frac{1}{2}}\Lambda  + \Lambda W_2\big(p_{T}, p_{T,{{\bm{\phi}_0}}} \big)
    \\  &+ \mathcal{L}_{\text{PaGoDA}}({\bm{\theta}};{{\bm{\phi}_0}})+\Big[\mathbb{E}_{p_{\text{\text{data}}}(\mathbf{x})p_{{\bm{\phi}_0}}(\mathbf{z}\vert\mathbf{x})}\big\Vert \mathbf{x}-G^{T\rightarrow0}(\mathbf{z}) \big\Vert_{2}^{2}\Big]^{\frac{1}{2}}
    \\  &+ e^{-\frac{1}{2}h(\gamma, T) T}W_2(p_{T,{{\bm{\phi}_0}}}, p_{T}) + \frac{\epsilon_{\text{DM}}}{\sqrt{2\delta h(\gamma,T)}} \big(1-e^{-h(\gamma, T)T}\big)^{\frac{1}{2}}
    \\ = &~\mathcal{L}_{\text{\text{rec}}}({\bm{\theta}};{{\bm{\phi}_0}}) +\Big[\mathbb{E}_{p_{\text{\text{data}}}(\mathbf{x})p_{{\bm{\phi}_0}}(\mathbf{z}\vert\mathbf{x})}\big\Vert \mathbf{x}-G^{T\rightarrow0}(\mathbf{z}) \big\Vert_{2}^{2}\Big]^{\frac{1}{2}}
    \\ &+ \big(\Lambda +e^{-\frac{1}{2}h(\gamma, T) T}\big) W_2\big(p_{T}, p_{T,{{\bm{\phi}_0}}} \big)
    \\ &+ \frac{\epsilon_{\text{DM}}}{\sqrt{2\delta h(\gamma,T)}} \big(1-e^{-h(\gamma, T)T}\big)^{\frac{1}{2}} + e^{-\frac{T}{2}} m\Lambda.
\end{align*}
This shows the first inequality in Theorem~\ref{th:w_2_convergence}.

Now, we show the second inequality in the statement of Theorem~\ref{th:w_2_convergence}. First, we establish an upper bound for $\Big[\mathbb{E}_{p_{\text{\text{data}}}(\mathbf{x})p_{{\bm{\phi}_0}}(\mathbf{z}\vert\mathbf{x})}\big\Vert \mathbf{x}-G^{T\rightarrow0}(\mathbf{z}) \big\Vert_{2}^{2}\Big]^{\frac{1}{2}}$  in terms of $\epsilon_{\text{DM}}$. Let $G^{0\rightarrow T}_{{\bm{\phi}_0}}$ denote the transition map defined by the empirical PF-ODE defined by the teacher $p_{{\bm{\phi}_0}}(\mathbf{x}\vert\mathbf{z})$, and $G^{0\rightarrow T}$ denote the ground truth transition map defined by the PF-ODE from $0$ to $T$. Then we have $\mathbf{x}=G^{T\rightarrow 0}(G^{0\rightarrow T}(\mathbf{x}))$ for all $\mathbf{x}\in\text{supp}(p_{\text{\text{data}}})$, and
\begin{align}\label{eq:W_2_looser_G_G_phi}
    \Big[\mathbb{E}_{p_{\text{\text{data}}(\mathbf{x})}p_{{\bm{\phi}_0}}(\mathbf{z}\vert\mathbf{x})}\big\Vert \mathbf{x}-G^{T\rightarrow0}(\mathbf{z}) \big\Vert_{2}^{2} \Big]^{1/2} 
    &= \Big[\mathbb{E}_{p_{\text{\text{data}}}(\mathbf{x})}\big\Vert G^{T\rightarrow 0}(G^{0\rightarrow T}(\mathbf{x}))-G^{T\rightarrow0}(G^{0\rightarrow T}_{{\bm{\phi}_0}}(\mathbf{x})) \big\Vert_{2}^{2} \Big]^{1/2} \nonumber \\
    &\leq \Lambda \Big[\mathbb{E}_{p_{\text{\text{data}}}(\mathbf{x})}\big\Vert G^{0\rightarrow T}(\mathbf{x})-G^{0\rightarrow T}_{{\bm{\phi}_0}}(\mathbf{x}) \big\Vert_{2}^{2} \Big]^{1/2}.
\end{align}
Here, we utilize the assumption that Assumption~\ref{th:assumption} (iii) also holds for $G^{T\rightarrow 0}$.

Consider the following two forward-time PF-ODEs on the interval $[0, T]$, both starting from $\mathbf{x}_{0} \sim p_{\text{\text{data}}}$:
\begin{align*}
      \frac{\diff \mathbf{x}_{t}}{\diff t} =-\mathbf{x}_t -\nabla\log p_{t} (\mathbf{x}_{t}), \quad 
      \frac{\diff \mathbf{x}_{t,{{\bm{\phi}_0}}}}{\diff t} =-\mathbf{x}_{t,{{\bm{\phi}_0}}} -\mathbf{s}_{{{\bm{\phi}_0}}}(\mathbf{x}_{t,{{\bm{\phi}_0}}}, t).
\end{align*}
By subtracting them and integrating from $0$ to $t$, we obtain
\begin{align*}
      \norm{\mathbf{x}_{t} - \mathbf{x}_{t,{{\bm{\phi}_0}}}}_2 &\leq  \underset{\mathbf{0}}{\underbrace{\mathbf{x}_{0} - \mathbf{x}_{0,{{\bm{\phi}_0}}}}} +\int_{0}^{t}\norm{ \big(\mathbf{x}_{\tau} - \mathbf{x}_{\tau,{{\bm{\phi}_0}}}\big)+\big(\nabla\log p_{\tau} (\mathbf{x}_{\tau}) - \mathbf{s}_{{{\bm{\phi}_0}}}(\mathbf{x}_{\tau,{{\bm{\phi}_0}}}, \tau)\big)}_2\diff t\\
      &\leq \int_{0}^{t}\norm{\mathbf{x}_{\tau} - \mathbf{x}_{\tau,{{\bm{\phi}_0}}}}_2\diff \tau +\epsilon_{\text{DM}}T.
\end{align*}
By applying  Grönwall's inequality,
\begin{align}\label{eq:W_2_looser_G_G_phi_gw}
    \norm{\mathbf{x}_{t} - \mathbf{x}_{t,{{\bm{\phi}_0}}}}_2 \leq Te^T \epsilon_{\text{DM}}.
\end{align}
Combining the above inequality with inequality~\eqref{eq:W_2_looser_G_G_phi}, it implies
\begin{align}\label{eq:W_2_looser_G_G_phi_1}
    \Big[\mathbb{E}_{p_{\text{\text{data}}(\mathbf{x})}p_{{\bm{\phi}_0}}(\mathbf{z}\vert\mathbf{x})}\big\Vert \mathbf{x}-G^{T\rightarrow0}(\mathbf{z}) \big\Vert_{2}^{2} \Big]^{1/2}\leq\Lambda T e^T \epsilon_{\text{DM}}.
\end{align}
Next, we derive an upper bound for $W_2(p_{T,{{\bm{\phi}_0}}}, p_{T})$ related to $\epsilon_{\text{DM}}$. Let $\pi(\hat{\mathbf{z}},\mathbf{z})$ be a coupling between $\hat{\mathbf{z}}\sim p_{T,{{\bm{\phi}_0}}}=G^{0\rightarrow T}_{{\bm{\phi}_0}}\sharp p_{\text{\text{data}}}$ and $\mathbf{z}\sim p_{T}=G^{0\rightarrow T}\sharp p_{\text{\text{data}}}$.
\begin{align}\label{eq:W_2_looser_G_G_phi_2}
    W_2^2(p_{T,{{\bm{\phi}_0}}}, p_{T})=W_2^2(G^{0\rightarrow T}_{{\bm{\phi}_0}}\sharp p_{\text{\text{data}}}, G^{0\rightarrow T}\sharp p_{\text{\text{data}}}) \leq\mathbb{E}_{\pi(\hat{\mathbf{z}},\mathbf{z})}\norm{\hat{\mathbf{z}}-\mathbf{z}}_2^2 \leq \big(Te^T\epsilon_{\text{DM}}\big)^2,
\end{align}
where the last inequality is derived from the inequality~\eqref{eq:W_2_looser_G_G_phi_gw}.

Therefore, with the first conclusion of Theorem~\ref{th:w_2_convergence} and inequalities~\eqref{eq:W_2_looser_G_G_phi_1} and \eqref{eq:W_2_looser_G_G_phi_2}, we derive
\begin{align*}
    W_2(p_{0,{\bm{\theta}}}, p_{\text{\text{data}}})  
 \lesssim \mathcal{L}_{\text{\text{rec}}}({\bm{\theta}};{{\bm{\phi}_0}}) +\epsilon_{\text{DM}}+ e^{-\frac{T}{2}} m\Lambda.
\end{align*}
\end{myproof}

The proof can be easily extended in two directions:
     (1) a more general (VP)-SDE: 
    \begin{align*}
        \diff\mathbf{x}_t = -f(t)\mathbf{x}_t\diff t + g(t) \diff\mathbf{w}_t
    \end{align*}
    with $\norm{f}_{L^{\infty}(t;[0,T])}$, $\norm{g}_{L^{\infty}(t;[0,T])}<\infty$, and
    (2) truncation at the least time $t=\delta$ (instead of $t=0$), with an additional argument based on \citep{lyu2023sampling}
    \begin{align*}
    W_2(p_{\delta}, p_{\text{\text{data}}})&\leq \Big(\mathbb{E}_{p_{\text{\text{data}}}(\mathbf{x}_0) } \mathbb{E}_{p_{\text{\text{prior}}}(\bm{\xi})}\norm{(1-e^{-\delta})\mathbf{x}_0 + \sqrt{1-e^{-2\delta}}\bm{\xi}}_2^2 \Big)^{\frac{1}{2}} \nonumber
    \\ &\leq \big((1-e^{-\delta})^2m^2 + (1-e^{-2\delta})D \big)^{\frac{1}{2}} \nonumber
    \\ &\lesssim  (\sqrt{D}\vee m )\sqrt{\delta},
    \end{align*}
where $p_\delta = G^{T\rightarrow \delta}\sharp p_{\text{\text{prior}}}$.

\subsubsection{$W_1$ Bound with PaGoDA's Reconstruction Loss}\label{subsec:W_1_conv_analysis}

\begin{assumpB}\label{th:assumption_rate}
    \begin{enumerate}[(i)]
        \item $m:=\mathbb{E}_{p_{\text{\text{data}}}(\mathbf{x})}\norm{\mathbf{x}}_2<\infty$;
        \item There is a $\epsilon_{\text{DSM}}>0$ so that $\sup_{\mathbf{x}, t}\norm{\mathbf{s}_{\phi}(\mathbf{x}, t) - \nabla\log 
        p_t(\mathbf{x})}_2^2\leq \epsilon_{\text{DM}}^2$;
        \item $G_{\bm{\theta}}^{T\rightarrow 0}$ is Lipschitz in $\mathbf{x}$: 
        \begin{align*}
            \Lambda:=\sup_{\mathbf{x}\neq\mathbf{y}} \frac{\norm{G_{\bm{\theta}}^{T\rightarrow 0}(\mathbf{x})-G_{\bm{\theta}}^{T\rightarrow 0}(\mathbf{y})}_2}{\norm{\mathbf{x}-\mathbf{y}}_2}<\infty,
        \end{align*}
        for all $\bm{\theta}$ and $T$.
        \item $\nabla\log p_t(\cdot)$ is Lipschitz in $\mathbf{x}$ with integrable Lipschitz constant: 
            \begin{align*}
            \Lambda_s (t):=\sup_{\mathbf{x}\neq\mathbf{y}} \frac{\norm{\nabla\log 
            p_t(\mathbf{x})-\nabla\log 
            p_t(\mathbf{y})}_2}{\norm{\mathbf{x}-\mathbf{y}}_2}<\infty,
            \end{align*}
        and $\Lambda_s$ is an $L^1$-integrable function on $(0,\infty)$.
    \end{enumerate}
\end{assumpB}

In the following proposition, we prove a variant of Theorem~\ref{th:w_2_convergence} which does not assume log-concavity of the data density (i.e., Assumption~\ref{th:assumption} (iv)).

\begin{theorem}[Variant of Theorem~\ref{th:w_2_convergence}]\label{th:w_1_convergence_rough}  Assume that Assumption~\ref{th:assumption_rate} holds. Let $\nu$ be either the oracle data distribution $p_{\text{\text{data}}}$ or an empirical distribution  $\hat{p}_{\text{\text{data}}, N}:=\frac{1}{N}\sum_{i=1}^{N}{\delta_{\mathbf{x}_i}}$, where $\mathbf{x}_i\sim p_{\text{\text{data}}}$ for $i=1,\cdots,N$. Let the PaGoDA's reconstruction loss starting from $\nu$ be defined as
\begin{align*}
    \mathcal{L}_{\text{\text{rec}}}(\bm{\theta}_{\nu};{\bm{\phi}_0}):=\mathbb{E}_{\nu(\mathbf{x})p_{\phi}(\mathbf{z}|\mathbf{x})}\big[\norm{\mathbf{x}-G^{T\rightarrow 0}_{\bm{\theta}_{\nu}}(\mathbf{z})}_2\big].
\end{align*}
Then we have
\begin{align*}
    W_1(p_{0,{\bm{\theta}}}, \nu) &~\leq   \mathcal{L}_{\text{\text{rec}}}({\bm{\theta}}_{\nu};{{\bm{\phi}_0}})+\mathbb{E}_{\nu(\mathbf{x})p_{{\bm{\phi}_0}}(\mathbf{z}\vert\mathbf{x})}\Big[\big\Vert \mathbf{x}-G^{T\rightarrow0}(\mathbf{z}) \big\Vert_{2}\Big]  +C_T T\epsilon_{\text{DM}} \nonumber
    \\&~+ \big(C_T+\Lambda) W_1(p_{T,{{\bm{\phi}_0}}}, p_{T}) + e^{-T} \big(\mathbb{E}_{p_{\text{\text{data}}}(\mathbf{x}_0)}\norm{\mathbf{x}_0}_2\big)\Lambda \nonumber
\end{align*}

In particular, if we assume Assumption~\ref{th:assumption} (iii) holds also for $G^{T\rightarrow 0}$, then for $T=\mathcal{O}\Big(\log\big(\frac{m\Lambda}{\epsilon_{\text{DM}}}\big)^2\Big)$ is sufficiently large, we have
\begin{align*}
    W_1(p_{0,{\bm{\theta}}}, p_{\text{\text{data}}}) \lesssim \mathcal{L}_{\text{\text{rec}}}({\bm{\theta}};{{\bm{\phi}_0}}) + \epsilon_{\text{DM}}.
\end{align*}
Here, we use $\lesssim$ to absorb the dependence on the constants $T$ and $\gamma$ into the inequality.
\end{theorem}

\begin{myproof}{Theorem}{\ref{th:w_1_convergence_rough}}

Define $p_{T, {{\bm{\phi}_0}}}:=G_{{\bm{\phi}_0}}^{0\rightarrow T}\sharp \nu$, and $p_{0, {{\bm{\phi}_0}}}:=G^{T\rightarrow0} \sharp p_{T, {{\bm{\phi}_0}}}$. From the triangle inequality, we have
\begin{align*}
    W_1(p_{0,{\bm{\theta}}}, \nu) \leq \underset{(A)}{\underbrace{W_1(p_{0,{\bm{\theta}}}, p_{0,{{\bm{\phi}_0}}}) }}+ \underset{(B)}{\underbrace{W_1(p_{0,{{\bm{\phi}_0}}}, \nu)}}.
\end{align*}
For (A), by following the similar argument as in Theorem~\ref{th:w_2_convergence}, we can obtain
\begin{align}\label{eq:A_result_cor}
    (A) &\leq e^{-T} \big(\mathbb{E}_{p_{\text{\text{data}}}(\mathbf{x}_0)}\norm{\mathbf{x}_0}_2\big)\Lambda  + \Lambda W_1\big(p_{T}, p_{T,{{\bm{\phi}_0}}} \big) + \mathcal{L}_{\text{PaGoDA}}({\bm{\theta}}_{\nu};{{\bm{\phi}_0}})+\mathbb{E}_{\nu(\mathbf{x})p_{{\bm{\phi}_0}}(\mathbf{z}\vert\mathbf{x})}\Big[\big\Vert \mathbf{x}-G^{T\rightarrow0}(\mathbf{z}) \big\Vert_{2}\Big]
\end{align}
For (B), by subtracting the following equations and integrating over $t$ from $0$ to $T$,
\begin{align*}
\begin{cases}
      \frac{\diff \hat{\mathbf{z}}_{t,{{\bm{\phi}_0}}}}{\diff t} &=\hat{\mathbf{z}}_{t,{{\bm{\phi}_0}}} +\mathbf{s}_{{{\bm{\phi}_0}}}(\hat{\mathbf{z}}_{t,{{\bm{\phi}_0}}}, T-t), \quad \hat{\mathbf{z}}_{0,{{\bm{\phi}_0}}}\sim p_{T,{{\bm{\phi}_0}}} \\
\frac{\diff \hat{\mathbf{z}}_{t}}{\diff t} &=\hat{\mathbf{z}}_t +\nabla\log p_{T-t} (\hat{\mathbf{z}}_{t}), \quad 
      \hat{\mathbf{z}}_{0}\sim p_{T},
\end{cases}
\end{align*}
we will obtain
\begin{align*}
    \hat{\mathbf{z}}_{T,{{\bm{\phi}_0}}}- \hat{\mathbf{z}}_T = \big(\hat{\mathbf{z}}_{0,{{\bm{\phi}_0}}}- \hat{\mathbf{z}}_0\big) + \int_{0}^{T} \big(\mathbf{s}_{{{\bm{\phi}_0}}}(\hat{\mathbf{z}}_{t,{{\bm{\phi}_0}}}, T-t)-\nabla\log p_{T-t} (\hat{\mathbf{z}}_{t})\big)\diff u.
\end{align*}
Now let $u(t):=\mathbb{E}\norm{\hat{\mathbf{z}}_{t,{{\bm{\phi}_0}}}- \hat{\mathbf{z}}_t}_2$. Then
\begin{align*}
    u(t) \leq~& u(0) + \mathbb{E}\int_{0}^{t} \norm{\mathbf{s}_{{{\bm{\phi}_0}}}(\hat{\mathbf{z}}_{\tau,{{\bm{\phi}_0}}}, T-\tau)-\nabla\log p_{T-\tau} (\hat{\mathbf{z}}_{\tau})}_2\diff \tau
    \\ \leq~&  u(0) + \int_{0}^{T} \mathbb{E}\norm{\mathbf{s}_{{{\bm{\phi}_0}}}(\hat{\mathbf{z}}_{\tau,{{\bm{\phi}_0}}}, T-\tau)-\nabla\log p_{T-\tau} (\hat{\mathbf{z}}_{\tau,{{\bm{\phi}_0}}})}_2\diff \tau 
    \\&+ \int_{0}^{t} \mathbb{E}\norm{\nabla\log p_{T-\tau} (\hat{\mathbf{z}}_{\tau,{{\bm{\phi}_0}}})-\nabla\log p_{T-\tau} (\hat{\mathbf{z}}_{\tau})}_2\diff \tau
    \\ \leq~& u(0) + T\epsilon_{\text{DM}} + \int_{0}^{t}\Lambda_s (\tau)u(\tau)\diff \tau,
\end{align*}
 where $\Lambda_s (t)$ is the Lipschitz constant of $\nabla\log p_t(\cdot)$ in $\mathbf{x}$. By applying integral form of Grönwall's inequality, we get
\begin{align}\label{eq:B_result_cor}
    (B)\leq\mathbb{E}\norm{\hat{\mathbf{z}}_{T,{{\bm{\phi}_0}}}- \hat{\mathbf{z}}_T}_2& \leq C_T\mathbb{E}\norm{\hat{\mathbf{z}}_{0,{{\bm{\phi}_0}}}- \hat{\mathbf{z}}_0}_2+C_T T\epsilon_{\text{DM}}=C_T W_1(p_{T,{{\bm{\phi}_0}}}, p_{T})+C_T T\epsilon_{\text{DM}}.
\end{align}
where $C_T:=\exp\big( \int_{0}^{T} \Lambda_s (t) \diff t \big)$ and the last equality follows from choosing a coupling of $\hat{\mathbf{z}}_{0,{{\bm{\phi}_0}}}\sim p_{T,{{\bm{\phi}_0}}}$ and $\hat{\mathbf{z}}_{0}\sim p_{T}$ so that $W_1(p_{T,{{\bm{\phi}_0}}}, p_{T})=\mathbb{E}\norm{\hat{\mathbf{z}}_{0,{{\bm{\phi}_0}}} - \hat{\mathbf{z}}_0}_2$.

By combining inequalities \eqref{eq:A_result_cor} and \eqref{eq:B_result_cor}, we obtain
\begin{align*}
    W_1(p_{0,{\bm{\theta}}}, \nu) &~\leq   \mathcal{L}_{\text{\text{rec}}}({\bm{\theta}}_{\nu};{{\bm{\phi}_0}})+\mathbb{E}_{\nu(\mathbf{x})p_{{\bm{\phi}_0}}(\mathbf{z}\vert\mathbf{x})}\Big[\big\Vert \mathbf{x}-G^{T\rightarrow0}(\mathbf{z}) \big\Vert_{2}\Big]  +C_T T\epsilon_{\text{DM}} \nonumber
    \\&~+ \big(C_T+\Lambda) W_1(p_{T,{{\bm{\phi}_0}}}, p_{T}) + e^{-T} \big(\mathbb{E}_{p_{\text{\text{data}}}(\mathbf{x}_0)}\norm{\mathbf{x}_0}_2\big)\Lambda. \nonumber
\end{align*}
A similar argument to Theorem~\ref{th:w_2_convergence} can be applied to obtain the second inequality in the statement of Theorem~\ref{th:w_1_convergence_rough}.
\end{myproof}

\subsection{Optimality analysis}

In this section, we compare the optimality of the learned distributions resulting from PaGoDA's training and distillation-based training loss, incorporating GAN~\cite{kim2023consistency, luhman2021knowledge}.

\paragraph{PaGoDA's Loss} We recall PaGoDA's training objective $\mathcal{L}_{\text{PaGoDA}}$ 
\begin{align*}
    \mathcal{L}_{\text{PaGoDA}}(G_{\bm{\theta}},D_{\bm{\psi}})=\mathcal{L}_{\text{\text{rec}}}(G_{\bm{\theta}})+ \lambda\mathcal{L}_{\text{adv}}(G_{\bm{\theta}},D_{\bm{\psi}})
\end{align*}
leverages the reconstruction loss
\begin{align*}
    \mathcal{L}_{\text{\text{rec}}}(G_{\bm{\theta}})=\mathbb{E}_{p_{\text{\text{data}}}(\mathbf{x})}\Big[\big\Vert\mathbf{x}-G_{\bm{\theta}}\big(E(\mathbf{x})\big)\big\Vert_{2}^{2}\Big],
\end{align*}
and adversarial loss
\begin{align*}
    \mathcal{L}_{\text{adv}}(G_{\bm{\theta}},D_{\bm{\psi}})=\mathbb{E}_{p_{\text{\text{data}}}(\mathbf{x})}\big[\log{D_{\bm{\psi}}(\mathbf{x})}\big]+\mathbb{E}_{p_{\text{\text{prior}}}(\mathbf{z})}\Big[\log{\Big(1-D_{\bm{\psi}}\big(G_{\bm{\theta}}(\mathbf{z})\big)\Big)}\Big].
\end{align*}

\paragraph{Knowledge Distillation Loss}
In the realm of knowledge distillation (KD) methods for DMs, approaches like  \emph{local consistency}~\cite{song2023consistency}, \emph{global consistency}~\cite{luhman2021knowledge}, or \emph{soft consistency}~\cite{kim2023consistency} are utilized to learn the noise-to-data trajectory of the teacher DM.  Let us consider the global consistency loss as a case study (similar arguments can apply to other distillation objectives), where the teacher's trajectory is obtained by solving its empirical PF-ODE from $T$ to $0$. The long jump along the trajectory is represented as $G^{T\rightarrow 0}_{\text{teacher}}(\mathbf{z})$, where $\mathbf{z}$ denotes the initial point (noise), $T$ signifies the initial time, and $0$ denotes the final time. The output of $G^{T\rightarrow 0}_{\text{teacher}}$ corresponds to the estimation of clean data, starting from $\mathbf{z}$.
\begin{align*}
    \mathcal{L}_{\text{KD}}(G_{\bm{\theta}}):=\mathbb{E}_{p_{\text{\text{prior}}}(\mathbf{z})}\Big[\big\Vert G^{T\rightarrow 0}_{\text{teacher}}(\mathbf{z})-G_{\bm{\theta}}\big(\mathbf{z}\big)\big\Vert_{2}^{2}\Big].
\end{align*}
In this context, we abuse the notation $G_{\bm{\theta}}(\mathbf{z})$ to denote the generator for KD.

The training of KD can also incorporate adversarial loss for enhanced performance \cite{kim2023consistency, sauer2024fast}. We represent the combined loss as:
\begin{align*}
    \mathcal{L}_{\text{KD+GAN}}(G_{\bm{\theta}},D_{\bm{\psi}}):=\mathcal{L}_{\text{KD}}(G_{\bm{\theta}})+\mathcal{L}_{\text{adv}}(G_{\bm{\theta}},D_{\bm{\psi}}).
\end{align*}

\begin{theorem}\label{th:optimality} Let $p_{{\bm{\phi}_0}}$ be the density determined the teacher DM. Suppose that GAN admits an optimal discriminator $D^*$. 

\begin{itemize}
    \item In PaGoDA, assume that the network parametrized generator class $\{G_{\bm{\theta}}\}$ is expressive enough so that it can simultaneously optimize both $\mathcal{L}_{\text{\text{rec}}}(G_{\bm{\theta}})$ and  $\mathcal{L}_{\text{adv}}(G_{\bm{\theta}}; D^*)$ with a same minimizer. Namely,  $\argmin_{\bm{\theta}}\{\mathcal{L}_{\text{\text{rec}}}(G_{\bm{\theta}})\}\cap \argmin_{\bm{\theta}}\{\mathcal{L}_{\text{adv}}(G_{\bm{\theta}}; D^*)\}\neq \emptyset$. Then 
\begin{align*}
    p_{\bm{\theta}^*, \text{PaGoDA}}:=G_{\bm{\theta}^*,\text{PaGoDA}}\sharp p_{\text{\text{prior}}} = p_{\text{\text{data}}}.
\end{align*}
\item In contrast, suppose that $p_{{\bm{\phi}_0}}\neq p_{\text{\text{data}}}$, then under similar conditions for KD+GAN where $\argmin_{\bm{\theta}}\{\mathcal{L}_{\text{KD}}(G_{\bm{\theta}})\}\cap \argmin_{\bm{\theta}}\{\mathcal{L}_{\text{adv}}(G_{\bm{\theta}}; D^*)\}\neq \emptyset$, there is no minimizer $\bm{\theta}^*$ so that $p_{\bm{\theta}^*, \text{KD+GAN}}:=G_{\bm{\theta}^*,\text{KD+GAN}}\sharp p_{\text{\text{prior}}} = p_{\text{\text{data}}}$. 
\end{itemize}

\end{theorem}

 The first part of the proof of the theorem follows from the following Lemma.

\begin{lemma}\label{th:argmin_id}
    If $\argmin_{\bm{\theta}}\{f(\bm{\theta})\}\cap \argmin_{\bm{\theta}}\{g(\bm{\theta})\}\neq \emptyset$, then $\argmin_{\bm{\theta}}\{f(\bm{\theta})+g(\bm{\theta})\}=\argmin_{\bm{\theta}}\{f(\bm{\theta})\}\cap \argmin_{\bm{\theta}}\{g(\bm{\theta})\}$.
\end{lemma}
\begin{proof}
    First, we prove the relationship $\argmin_{\bm{\theta}}\{f(\bm{\theta})+g(\bm{\theta})\}\supseteq\argmin_{\bm{\theta}}\{f(\bm{\theta})\} \cap \argmin_{\bm{\theta}}\{g(\bm{\theta})\}$. Indeed, it holds without additional assumption. Suppose that $\bm{\theta}^* \in \argmin_{\bm{\theta}}\{f(\bm{\theta})\} \cap \argmin_{\bm{\theta}}\{g(\bm{\theta})\}$. Then for any $\bm{\theta}$, we have
    $f(\bm{\theta})\geq f(\bm{\theta}^*)$ and $g(\bm{\theta})\geq g(\bm{\theta}^*)$, which implies  $f(\bm{\theta}) + g(\bm{\theta})\geq f(\bm{\theta}^*)+  g(\bm{\theta}^*)$. That is, $\bm{\theta}^* \in \argmin_{\bm{\theta}}\{f(\bm{\theta})+g(\bm{\theta})\}$.

On the other hand, suppose that $\bm{\theta}^* \in \argmin_{\bm{\theta}}\{f(\bm{\theta})+g(\bm{\theta})\}$.  We want to prove that $\bm{\theta}^* \in \argmin_{\bm{\theta}}\{f(\bm{\theta})\} \cap \argmin_{\bm{\theta}}\{g(\bm{\theta})\}$. Let $\bm{\theta}_{\cap}^* \in\argmin_{\bm{\theta}}\{f(\bm{\theta})\}\cap \argmin_{\bm{\theta}}\{g(\bm{\theta})\}$, where we notice that the existence of $\bm{\theta}_{\cap}^*$ is guaranteed by the assumption. In particular, we have $f(\bm{\theta}^*)\geq f(\bm{\theta}_{\cap}^*)$ and $g(\bm{\theta}^*)\geq g(\bm{\theta}_{\cap}^*)$. Then for any $\bm{\theta}$, we have
\begin{align*}
    \min_{\bm{\theta}}\{f(\bm{\theta})+g(\bm{\theta})\} =f(\bm{\theta}^*)+g(\bm{\theta}^*) \geq f(\bm{\theta}_{\cap}^*)+g(\bm{\theta}_{\cap}^*) \geq \min_{\bm{\theta}}\{f(\bm{\theta})+g(\bm{\theta})\}.
\end{align*}
Thus, $\min_{\bm{\theta}}\{f(\bm{\theta})+g(\bm{\theta})\} =f(\bm{\theta}^*)+g(\bm{\theta}^*) = f(\bm{\theta}_{\cap}^*)+g(\bm{\theta}_{\cap}^*) $ and
\begin{align*}
    \big[f(\bm{\theta}^*)-f(\bm{\theta}_{\cap}^*)\big] + \big[g(\bm{\theta}^*)-g(\bm{\theta}_{\cap}^*)\big] = 0.
\end{align*}
This implies $f(\bm{\theta}^*)=f(\bm{\theta}_{\cap}^*)=\min_{\bm{\theta}}\{f(\bm{\theta})\}$ and $g(\bm{\theta}^*)=g(\bm{\theta}_{\cap}^*)=\min_{\bm{\theta}}\{g(\bm{\theta})\}$, as the individual terms are nonnegative. Therefore, $\bm{\theta}^* \in\argmin_{\bm{\theta}}\{f(\bm{\theta})\}\cap \argmin_{\bm{\theta}}\{g(\bm{\theta})\}$, which concludes the proof.

\end{proof}
\begin{myproof}{Theorem}{\ref{th:optimality}}
With the lemma above, let $\bm{\theta}^*\in\argmin_{\bm{\theta}}\mathcal{L}_{\text{PaGoDA}}(G_{\bm{\theta}},D^*)$. Consequently, $\bm{\theta}^*$ should also simultaneously minimize both  $\mathcal{L}_{\text{\text{rec}}}$ and $\mathcal{L}_{\text{adv}}$.
Minimizing $\mathcal{L}_{\text{\text{rec}}}$ implies that $p_{\bm{\theta}^*, \text{PaGoDA}}=G_{\bm{\theta}^*,\text{PaGoDA}}\sharp p_{T,{\bm{\phi}_0}}$, where $p_{T,{\bm{\phi}_0}}$ represents the density derived from solving the teacher's empirical PF-ODE forward, starting from $p_{\text{\text{data}}}$. On the other hand, optimizing $\mathcal{L}_{\text{adv}}$ implies that $p_{\bm{\theta}^*, \text{PaGoDA}}= p_{\text{\text{data}}}$ by applying Theorem 1 in \cite{goodfellow2014generative}. This establishes the first part of the theorem.

In the second part, suppose on the contrary that there is a minimizer $\bm{\theta}^*$ of $\mathcal{L}_{\text{KD+GAN}}$ such that $p_{\bm{\theta}^*, \text{KD+GAN}}= p_{\text{\text{data}}}$. Again, by applying the above lemma, we infer that $\bm{\theta}^*$ should also minimize $\mathcal{L}_{\text{KD}}$ (and $\mathcal{L}_{\text{adv}}$). This implies that $p_{\bm{\theta}^*, \text{KD+GAN}}=p_{{\bm{\phi}_0}}$. However, this contradicts our assumption that  $p_{\text{\text{data}}}\neq p_{{\bm{\phi}_0}}$. Thus, such a minizer does not exist and the second part of the theorem is proven.

\end{myproof}

We remark that (1) optimality of $\mathcal{L}_{\text{GAN}}(\bm{\theta})$ may not be unique in $\bm{\theta}$, and that (2) the first part of the theorem can be directly extended to scenarios involving downsampling in the encoder .

\subsection{Stability Analysis}

\subsubsection{Preliminaries of Dynamical System}\label{subsec:pre_dynamical}
 To study its convergence and stability, we first introduce the prerequisites for Lyapunov stability~\citep{asner1970total,bhatia2002stability} in a general setup. Let $\mathcal{F}\colon\Xi\rightarrow\Xi$ be a continuously differentiable operator (that is, $\mathscr{C}^1$ operator), where $\Omega\subset\mathbb{R}^N$. We consider the discrete iteration dynamical system defined by
\begin{align*}
    \bm{\xi}_{k+1} = \mathcal{F}(\bm{\xi}_{k}) \quad \text{with } \bm{\xi}_{0}\in\Omega.
\end{align*}
Namely, $\bm{\xi}_{k+1}= \mathcal{F}^{(k)}(\bm{\xi}_{0}):=\underset{\text{k-copies}}{\underbrace{\mathcal{F}\circ\cdots\circ\mathcal{F}}}(\bm{\xi}_{0})$. A point $\bm{\xi}^*\in\Omega$  is called a \emph{fixed point} or \emph{equilibrium} (we use the terms interchangeably) of $\mathcal{F}$ if $\bm{\xi}^*=\mathcal{F}(\bm{\xi}^*)$. The stability and convergence analysis focuses on how the dynamical system $\mathcal{F}^{(k)}(\bm{\xi}_{0})$ approaches a fixed point as iterations $k$ are sufficiently large.

\begin{definition}{(Stability~\cite{bhatia2002stability})} Let $\bm{\xi}^*$ be an equilibrium of the $\mathscr{C}^1$ operator $\mathcal{F}\colon\Omega\rightarrow\Omega$. The equilibrium $\bm{\xi}^*$ is said to be
\begin{itemize}
    \item  \emph{stable} if for every $\epsilon>0$ there is a $\delta>0$ so that whenever $\norm{\bm{\xi}-\bm{\xi}^*}_2<\delta$, we have $\norm{\mathcal{F}^{(k)}(\bm{\xi})-\bm{\xi}^*}_2<\epsilon$ for all $k\in\mathbb{N}\cup\{0\}$.
    \item \emph{asymptotically stable} if $\bm{\xi}^*$ is stable, and there is a $\delta>0$ so that whenever $\norm{\bm{\xi}-\bm{\xi}^*}_2<\delta$, we have $\lim_{k\rightarrow\infty}\norm{\mathcal{F}^{(k)}(\bm{\xi})-\bm{\xi}^*}_2=0$.
    \item \emph{exponentially stable} if $\bm{\xi}^*$ is asymptotically stable, and there is a $\delta>0$ and $\alpha,\beta>0$ so that whenever $\norm{\bm{\xi}-\bm{\xi}^*}_2<\delta$, we have $\norm{\mathcal{F}^{(k)}(\bm{\xi})-\bm{\xi}^*}_2\leq \alpha\norm{\bm{\xi}-\bm{\xi}^*}_2 e^{-\beta k}$ for all $k\in\mathbb{N}\cup\{0\}$. The largest $\beta>0$ that satisfies the inequality for exponential stability is referred to as the \emph{rate of convergence}.
\end{itemize}
Let $\Gamma$ be a subset of the set of all equilibria. We say the dynamical system $\mathcal{F}^{(k)}$ \emph{locally converges on $\Gamma$} if $\mathcal{F}^{(k)}$ is exponentially stable at any point in $\Gamma$.
\end{definition}

The intuitions of the above stability notions are
\begin{itemize}
    \item A \emph{stable} equilibrium indicates that if an initialization is within some $\delta$-neighborhood of the equilibrium, the iterations starting from that initialization will always remain within an $\epsilon$-neighborhood of the equilibrium, for any arbitrarily chosen $\epsilon$. 
    \item An \emph{asymptotically stable} equilibrium indicates that iterations starting near the equilibrium not only remain close but ultimately converge to the equilibrium.
    \item An \emph{asymptotically stable} equilibrium indicates that  the iterations not only converge but do so at a rate no slower than the rate $e^{-\beta k}$ with respective to iteration step $k$.
\end{itemize}

Analyzing the eigenvalues of the Jacobian $\nabla_{\bm{\xi}}\mathcal{F}(\bm{\xi}^*)$ of the operator $\mathcal{F}$ at an equilibrium $\bm{\xi}^*$ is a crucial tool for studying stability. In principle \citep{asner1970total,bhatia2002stability}, if we can ensure that the Jacobian of $\mathcal{F}$ at some equilibrium has only eigenvalues with strictly negative real parts, then the dynamical system $\mathcal{F}^{(k)}$ is asymptotically stable at that equilibrium. In particular, we refer to a matrix as a \emph{Hurwitz matrix} if all its eigenvalues have strictly negative real parts. 

In the following lemma, we present a necessary condition to ensure that a special class of matrices will be Hurwitz.

\begin{lemma}{(Necessary condition for a Hurwitz matrix~\cite{mescheder2018training})}\label{th:lemma_eigen} Consider the following matrix $\mathcal{J}\in\mathbb{R}^{(N+M)\times(N+M)}$ with $P\in\mathbb{R}^{N\times N}$, $Q\in\mathbb{R}^{M\times M}$, and $B\in\mathbb{R}^{M\times N}$.
    \begin{align*} 
    \mathcal{J}=\begin{bmatrix}
 P& -B^T \\
B & Q
\end{bmatrix}.
\end{align*}
Suppose that $B$ is full rank. Then all eigenvalues of $\mathcal{J}$ have negative real part, if either (1) $P$ is negative definite and $Q$ is negative semi-definite, or (2) $P$ is negative semi-definite and $Q$ is negative definite.  
\end{lemma}

\subsubsection{Preliminaries for Analysis of PaGoDA Training}\label{subsec:pre_pagoda}

We consider PaGoDA's training, integrating reconstruction and adversarial losses with a weight $\eta>0$.
\begin{align}
    \mathcal{L}(\bm{\theta}, \bm{\psi}):=&~\mathbb{E}_{p_\text{\text{data}}(\mathbf{x})}\big[\eta\norm{\mathbf{x}-G_{\bm{\theta}}(E(\mathbf{x}))}_2^2+f(D_{\bm{\psi}}(\mathbf{x}))\big] + \mathbb{E}_{p_{G_{\bm{\theta}}}(\mathbf{x})}\big[f(-D_{\bm{\psi}}(\mathbf{x}))\big]\label{eq:PaGoDA_GAN_1} \\
     =&~\mathbb{E}_{p_\text{\text{data}}(\mathbf{x})}\big[\eta\norm{\mathbf{x}-G_{\bm{\theta}}(E(\mathbf{x}))}_2^2+f(D_{\bm{\psi}}(\mathbf{x}))\big] + \mathbb{E}_{p_\text{\text{prior}}(\mathbf{z})}\big[f(-D_{\bm{\psi}}(G_{\bm{\theta}}(\mathbf{z})))\big] \label{eq:PaGoDA_GAN_2}.
\end{align}
Here, $f\colon\mathbb{R}\rightarrow\mathbb{R}$ is a continuous differentiable function. In the vanilla GAN~\cite{goodfellow2014generative}, the $f$-function is taken as $f(u):=-\log\big(1+\exp(-u)\big)$, where $f'(u)=\exp(-u)/\big(1+\exp(-u)\big)>0$ and $f''(u)=-\exp(-u)/\big(1+\exp(-u)\big)<0$ for all $u\in\mathbb{R}$. We maintain the generality of $f$ and will prove the training stability of PaGoDA across a wide class of $f$.

The velocity field $\mathbf{v}(\bm{\theta}, \bm{\psi})$ corresponding to the gradient descent update is 
\begin{align*}
    \mathbf{v}(\bm{\theta}, \bm{\psi}):=\begin{bmatrix}
 -\nabla_{\bm{\theta}} \mathcal{L}(\bm{\theta}, \bm{\psi})\\
\nabla_{\bm{\psi}} \mathcal{L}(\bm{\theta}, \bm{\psi})
\end{bmatrix}.
\end{align*}

Gradient descent is a special case of fixed-point iteration. Now, we specify the operator $\mathcal{F}$ as an alternative gradient descent operator. That is, we consider $\mathcal{F}_h:=\mathcal{F}_{D,h}\circ\mathcal{F}_{G,h}$ with a learning rate $h>0$. Here,
\begin{align*}
    \mathcal{F}_{G,h}(\bm{\theta}, \bm{\psi}):= \begin{bmatrix}
 \bm{\theta}-h\nabla_{\bm{\theta}} \mathcal{L}(\bm{\theta}, \bm{\psi})\\
\bm{\psi} 
\end{bmatrix}
\quad \text{and}\quad
    \mathcal{F}_{D,h}(\bm{\theta}, \bm{\psi}):= \begin{bmatrix}
 \bm{\theta}\\
\bm{\psi}+h\nabla_{\bm{\psi}} \mathcal{L}(\bm{\theta}, \bm{\psi})
\end{bmatrix}.
\end{align*}


A point $(\bm{\theta}^*, \bm{\psi}^*)$ is called an \emph{equilibrium} of the system defined by $\mathbf{v}$ if $\mathbf{v}(\bm{\theta}^*, \bm{\psi}^*)=0$ (equivalently, $\mathcal{F}_h (\bm{\theta}^*, \bm{\psi}^*)= 0 $).
We can analyze the learning dynamic via the Jacobian matrix of $\mathbf{v}(\bm{\theta}, \bm{\psi})$ which is defined as the following:
\begin{align*}
    \mathcal{J}(\bm{\theta}, \bm{\psi}):=\begin{bmatrix}
 -\nabla_{\bm{\theta}}^2 \mathcal{L}(\bm{\theta}, \bm{\psi})& -\nabla_{\bm{\theta},\bm{\psi}}^2 \mathcal{L}(\bm{\theta}, \bm{\psi}) \\
\nabla_{\bm{\theta},\bm{\psi}}^2 \mathcal{L}(\bm{\theta}, \bm{\psi}) & \nabla_{\bm{\psi}}^2 \mathcal{L}(\bm{\theta}, \bm{\psi})
\end{bmatrix}.
\end{align*}

The following proposition relates Lemma~\ref{th:jacobian} to the stability of the gradient descent operator $\mathcal{F}_h$, serving as the main tool to prove the training stability of PaGoDA in Theorem~\ref{th:stable_pagoda}.

\begin{lemma}{(Locally stable on manifold -- modification of
\citep{mescheder2018training})}\label{th:stable_mfld} Suppose that the gradient descent operator $\mathcal{F}_h=\mathcal{F}_h(\bm{\upsilon}, \bm{\omega})$ is a $\mathscr{C}^1$ mapping. Let $(\bm{\upsilon}^*, \bm{\omega}^*)$ be an equilibrium (fixed point) of $\mathcal{F}_h$. Assume that there is a neighborhood $\Omega$ of $\bm{\omega}^*$ so that $\mathcal{F}_h$ admits equilibrium on  $\{\bm{\upsilon}^*\}\times\Omega $:
\begin{align*}
    \mathcal{F}_h(\bm{\upsilon}^*, \bm{\omega})=(\bm{\upsilon}^*, \bm{\omega})\quad \text{for all } \omega \in\Omega.
\end{align*}
If all the eigenvalues of $\mathcal{J}:=\nabla_{\bm{\bm{\upsilon}}}\mathcal{F}_h(\bm{\upsilon}^*, \bm{\omega}^*)$ have negative real parts, then for a sufficiently small learning rate $h$, the gradient descent iteration defined by $\mathcal{F}_h$ locally converges on $\Gamma:=\{(\bm{\upsilon}^*, \bm{\omega})\big\vert\omega \in\Omega\}$ with a rate of convergence $\abs{\lambda_{\text{max}}}$. Here, $\lambda_{\text{max}}$ denotes the eigenvalue of $\mathcal{J}$ with the largest absolute value. 
\end{lemma}

\begin{myproof}{Lemma}{\ref{th:stable_mfld}}
    This proposition is followed by Lemma~A.5. and Theorem~A.3. of \citep{mescheder2018training}.
\end{myproof}

\subsubsection{PaGoDA's Training is Stable}\label{subsec:pagoda_stable}

Proving PaGoDA's stability involves two steps: First, derive the components of First, deriving the components of $\mathcal{J}(\bm{\theta}^*, \bm{\psi}^*)$. Second, verify that these components satisfy Lemma~\ref{th:lemma_eigen}. After these, we can apply Lemma~\ref{th:stable_mfld} to conclude PaGoDA's training stability whenever the learning rate $h>0$ is sufficiently small.

\begin{assumpC}\label{th:assumption_stable_reduce}
    \begin{enumerate}[(i)]
        \item $E$ is not an identity map. 
        \item At $\bm{\theta}^*$, $p_{\bm{\theta}^*}=p_{\text{\text{data}}}$, and $\mathbf{x}=G_{\bm{\theta}^*}(E(\mathbf{x}))$ for a.e. $\mathbf{x}\in \text{supp}\big(p_{\text{\text{data}}}\big)$.  
        \item At $\bm{\psi}^*$, $D_{\bm{\psi}^*}(\mathbf{x})=0$ and $\nabla_{\mathbf{x}}D_{\bm{\psi}^*}(\mathbf{x})=0$ for $\mathbf{x}\in\text{supp}(p_{\text{\text{data}}})$. 
        
    \end{enumerate}
\end{assumpC}

\begin{lemma}\label{th:jacobian} Suppose that Assumption~\ref{th:assumption_stable_reduce} holds for an equilibrium $(\bm{\theta}^*, \bm{\psi}^*)$. Then the Jacobian at the equilibrium can be computed as
\begin{align*}
    \mathcal{J}(\bm{\theta}^*, \bm{\psi}^*)=\begin{bmatrix}
 K_{GG}& -K_{DG}^T \\
K_{DG} & K_{DD}
\end{bmatrix}.
\end{align*}
Here,  and
\begin{align*}
    K_{GG}=& -2\eta\mathbb{E}_{p_\text{\text{data}}(\mathbf{x})}\big[\nabla_{\bm{\theta}}G_{\bm{\theta}^*}(E(\mathbf{x}))^T\cdot\nabla_{\bm{\theta}}G_{\bm{\theta}^*}(E(\mathbf{x}))\big] \\ &+ f'(0)\mathbb{E}_{p_\text{\text{prior}}(\mathbf{z})}\big[\nabla_{\bm{\theta}}G_{\bm{\theta}^*}(\mathbf{z})^T\cdot\nabla^2_{\mathbf{x}}D_{\bm{\psi}^*}(G_{\bm{\theta}^*}(\mathbf{z}))\cdot\nabla_{\bm{\theta}}G_{\bm{\theta}^*}(\mathbf{z})\big]. \\
    K_{DG}&= -f'(0)\nabla_{\bm{\theta}} \mathbb{E}_{p_{G_{\bm{\theta}}}(\mathbf{x})}\big[ \nabla_{\bm{\psi}}D_{\bm{\psi}^*}(\mathbf{x})\big]\Big\vert_{\bm{\theta}=\bm{\theta}^*} \\
    K_{DD}&=2f''(0) \mathbb{E}_{p_\text{\text{data}}(\mathbf{x})}\big[\nabla_{\bm{\psi}}D_{\bm{\psi}^*}(\mathbf{x})\cdot\nabla_{\bm{\psi}}D_{\bm{\psi}^*}(\mathbf{x})^T\big].
\end{align*}

\end{lemma}

\begin{myproof}{Lemma}{\ref{th:jacobian}}
We first compute the gradients of $\mathcal{L}$ in terms of $\bm{\theta}$ and $\bm{\psi}$, where we utilize the formulations Eqs.~\eqref{eq:PaGoDA_GAN_1} and \eqref{eq:PaGoDA_GAN_2}, respectively.
    \begin{align}
    \nabla_{\bm{\theta}}\mathcal{L}(\bm{\theta}, \bm{\psi})=&-2\eta\mathbb{E}_{p_\text{\text{data}}(\mathbf{x})}\big[\inner{\mathbf{x}-G_{\bm{\theta}}(E(\mathbf{x}))}{\nabla_{\bm{\theta}}G_{\bm{\theta}}(E(\mathbf{x}))   } \big] \nonumber 
    \\&- \mathbb{E}_{p_\text{\text{prior}}(\mathbf{z})}\big[f'(-D_{\bm{\psi}}(G_{\bm{\theta}}(\mathbf{z})))\cdot\nabla_{\mathbf{x}}D_{\bm{\psi}}(G_{\bm{\theta}}(\mathbf{z}))\cdot\nabla_{\bm{\theta}}G_{\bm{\theta}}(\mathbf{z})\big] \label{eq:grad_theta_PaGoDA_GAN}.\\
    \nabla_{\bm{\psi}}\mathcal{L}(\bm{\theta}, \bm{\psi})=&~ \mathbb{E}_{p_\text{\text{data}}(\mathbf{x})}\big[f'(D_{\bm{\psi}}(\mathbf{x}))\nabla_{\bm{\psi}}D_{\bm{\psi}}(\mathbf{x}))\big] - \mathbb{E}_{p_{G_{\bm{\theta}}}(\mathbf{x})}\big[f'(-D_{\bm{\psi}}(\mathbf{x}))\nabla_{\bm{\psi}}D_{\bm{\psi}}(\mathbf{x}))\big] \label{eq:grad_psi_PaGoDA_GAN}.
\end{align}

    \begin{align*}
    \nabla^2_{\bm{\theta}}\mathcal{L}(\bm{\theta}, \bm{\psi})=&~2\eta\mathbb{E}_{p_\text{\text{data}}(\mathbf{x})}\big[\inner{\nabla_{\bm{\theta}}G_{\bm{\theta}}(E(\mathbf{x})) }{\nabla_{\bm{\theta}}G_{\bm{\theta}}(E(\mathbf{x})) } \big] - 2\eta\mathbb{E}_{p_\text{\text{data}}(\mathbf{x})}\big[\inner{\mathbf{x}-G_{\bm{\theta}}(E(\mathbf{x}))}{ \nabla^2_{\bm{\theta}}G_{\bm{\theta}}(E(\mathbf{x})) } \big] \nonumber 
    \\&+ \mathbb{E}_{p_\text{\text{prior}}(\mathbf{z})}\big[f''(-D_{\bm{\psi}}(G_{\bm{\theta}}(\mathbf{z})))\cdot\nabla_{\mathbf{x}}D_{\bm{\psi}}(G_{\bm{\theta}}(\mathbf{z}))\cdot\nabla_{\bm{\theta}}G_{\bm{\theta}}(\mathbf{z})\cdot\nabla_{\mathbf{x}}D_{\bm{\psi}}(G_{\bm{\theta}}(\mathbf{z}))\cdot\nabla_{\bm{\theta}}G_{\bm{\theta}}(\mathbf{z})\big]
    \\&- \mathbb{E}_{p_\text{\text{prior}}(\mathbf{z})}\big[f'(-D_{\bm{\psi}}(G_{\bm{\theta}}(\mathbf{z})))\cdot\nabla_{\bm{\theta}}G_{\bm{\theta}}(\mathbf{z})^T\cdot\nabla^2_{\mathbf{x}}D_{\bm{\psi}}(G_{\bm{\theta}}(\mathbf{z}))\cdot\nabla_{\bm{\theta}}G_{\bm{\theta}}(\mathbf{z})\big] 
    \\&- \mathbb{E}_{p_\text{\text{prior}}(\mathbf{z})}\big[f'(-D_{\bm{\psi}}(G_{\bm{\theta}}(\mathbf{z})))\cdot\nabla_{\mathbf{x}}D_{\bm{\psi}}(G_{\bm{\theta}}(\mathbf{z}))\cdot\nabla^2_{\bm{\theta}}G_{\bm{\theta}}(\mathbf{z})\big].
\end{align*}

According to Assumption~\ref{th:assumption_stable_reduce} (ii) and (iii), we have
\begin{align*}
    \nabla^2_{\bm{\theta}}\mathcal{L}(\bm{\theta}^*, \bm{\psi}^*)=&~2\eta\mathbb{E}_{p_\text{\text{data}}(\mathbf{x})}\big[\nabla_{\bm{\theta}}G_{\bm{\theta}^*}(E(\mathbf{x}))^T\cdot\nabla_{\bm{\theta}}G_{\bm{\theta}^*}(E(\mathbf{x}))\big] \\&- f'(0)\mathbb{E}_{p_\text{\text{prior}}(\mathbf{z})}\big[\nabla_{\bm{\theta}}G_{\bm{\theta}^*}(\mathbf{z})^T\cdot\nabla^2_{\mathbf{x}}D_{\bm{\psi}}(G_{\bm{\theta}^*}(\mathbf{z}))\cdot\nabla_{\bm{\theta}}G_{\bm{\theta}^*}(\mathbf{z})\big]. 
\end{align*}
Thus, we obtain
\begin{align*}
    K_{GG}=&~-\nabla^2_{\bm{\theta}}\mathcal{L}(\bm{\theta}^*, \bm{\psi}^*)\\
    =&~-2\eta\mathbb{E}_{p_\text{\text{data}}(\mathbf{x})}\big[\nabla_{\bm{\theta}}G_{\bm{\theta}^*}(E(\mathbf{x}))^T\cdot\nabla_{\bm{\theta}}G_{\bm{\theta}^*}(E(\mathbf{x}))\big] + f'(0)\mathbb{E}_{p_\text{\text{prior}}(\mathbf{z})}\big[\nabla_{\bm{\theta}}G_{\bm{\theta}^*}(\mathbf{z})^T\cdot\nabla^2_{\mathbf{x}}D_{\bm{\psi}}(G_{\bm{\theta}^*}(\mathbf{z}))\cdot\nabla_{\bm{\theta}}G_{\bm{\theta}^*}(\mathbf{z})\big]. 
\end{align*}

To compute $K_{DG}$, we first derive $\nabla_{\bm{\theta}}\mathcal{L}$ from Eq.~\eqref{eq:PaGoDA_GAN_1} as
\begin{align*}
    \nabla_{\bm{\theta}}\mathcal{L}(\bm{\theta}, \bm{\psi})=&~-2\eta\mathbb{E}_{p_\text{\text{data}}(\mathbf{x})}\big[\inner{\mathbf{x}-G_{\bm{\theta}}(E(\mathbf{x}))}{\nabla_{\bm{\theta}}G_{\bm{\theta}}(E(\mathbf{x})) }\big]+ \nabla_{\bm{\theta}} \mathbb{E}_{p_{G_{\bm{\theta}}}(\mathbf{x})}\big[f(-D_{\bm{\psi}}(\mathbf{x}))\big].
\end{align*}
Thus, we can compute
\begin{align*}
    \nabla^2_{\bm{\theta},\bm{\psi}}\mathcal{L}(\bm{\theta}, \bm{\psi})= -\nabla_{\bm{\theta}} \mathbb{E}_{p_{G_{\bm{\theta}}}(\mathbf{x})}\big[f'(-D_{\bm{\psi}}(\mathbf{x}))\cdot \nabla_{\bm{\psi}}D_{\bm{\psi}}(\mathbf{x})\big], 
\end{align*}
and hence,
\begin{align*}
    K_{DG}=\nabla^2_{\bm{\theta},\bm{\psi}}\mathcal{L}(\bm{\theta}^*, \bm{\psi}^*)= -f'(0)\nabla_{\bm{\theta}} \mathbb{E}_{p_{G_{\bm{\theta}}}(\mathbf{x})}\big[ \nabla_{\bm{\psi}}D_{\bm{\psi}^*}(\mathbf{x})\big]\Big\vert_{\bm{\theta}=\bm{\theta}^*}. 
\end{align*}

To compute $K_{DD}$, we can obtain from Eq.~\eqref{eq:PaGoDA_GAN_2} that
\begin{align*}
    \nabla^2_{\bm{\psi}}\mathcal{L}(\bm{\theta}, \bm{\psi})=&~ \mathbb{E}_{p_\text{\text{data}}(\mathbf{x})}\big[f''(D_{\bm{\psi}}(\mathbf{x}))\nabla_{\bm{\psi}}D_{\bm{\psi}}(\mathbf{x})\cdot\nabla_{\bm{\psi}}D_{\bm{\psi}}(\mathbf{x})^T\big] \\&+~ \mathbb{E}_{p_{G_{\bm{\theta}}}(\mathbf{x})}\big[f''(-D_{\bm{\psi}}(\mathbf{x}))
    \nabla_{\bm{\psi}}D_{\bm{\psi}}(\mathbf{x})\cdot\nabla_{\bm{\psi}}D_{\bm{\psi}}(\mathbf{x})^T\big]\\&+~\mathbb{E}_{p_\text{\text{data}}(\mathbf{x})}\big[f'(D_{\bm{\psi}}(\mathbf{x}))\nabla^2_{\bm{\psi}}D_{\bm{\psi}}(\mathbf{x}))\big]
 - \mathbb{E}_{p_{G_{\bm{\theta}}}(\mathbf{x})}\big[f'(-D_{\bm{\psi}}(\mathbf{x}))\nabla^2_{\bm{\psi}}D_{\bm{\psi}}(\mathbf{x}))\big] .
\end{align*}

Hence, by using Assumption~\ref{th:assumption_stable_reduce} (ii) and (iii), we get
\begin{align*}
    K_{DD}=\nabla^2_{\bm{\psi}}\mathcal{L}(\bm{\theta}^*, \bm{\psi}^*)=&~ 2f''(0) \mathbb{E}_{p_\text{\text{data}}(\mathbf{x})}\big[\nabla_{\bm{\psi}}D_{\bm{\psi}^*}(\mathbf{x})\cdot\nabla_{\bm{\psi}}D_{\bm{\psi}^*}(\mathbf{x})^T\big].
\end{align*}

\end{myproof}

We consider the following two sets
\begin{align*}
    \mathcal{M}_{G}&:=\big\{\bm{\theta}\big\vert p_{\bm{\theta}}=p_{\text{\text{data}}},\, \mathbf{x}=G_{\bm{\theta}}(E(\mathbf{x})) \text{ for a.e. } \mathbf{x}\in \text{supp}\big(p_{\text{\text{data}}}\big)\big\}\\
    \mathcal{M}_{D}&:=\big\{\bm{\psi}\big\vert S(\bm{\psi})=0\big\}, 
\end{align*}
where $S(\bm{\psi}):=\mathbb{E}_{p_{\text{\text{data}}}(\mathbf{x})}\big[\abs{D_{\bm{\psi}}(\mathbf{x})}^2+\norm{\nabla_{\mathbf{x}}D_{\bm{\psi}}(\mathbf{x})}_2^2\big]$. Also, we let  $\mathcal{T}_{\bm{\psi}^*}\mathcal{M}_D$ denote the tangent space of $\mathcal{M}_D$ at $\bm{\psi}^*$.

\begin{assumpC}\label{th:assumption_stable}
    \begin{enumerate}[(i)]
        \item The second continuously differentiable function $f\colon\mathbb{R}\rightarrow\mathbb{R}$ satisfies: $f'(0)> 0 $ and $f''(0)< 0$.
        \item There is a $\delta>0$ so that $\mathcal{M}_G\cap\mathbb{B}_{\delta}(\bm{\theta}^*)$ and $\mathcal{M}_D\cap\mathbb{B}_{\delta}(\bm{\psi}^*)$ are  $\mathscr{C}^1$ manifolds.
        \item $\nabla_{\bm{\theta}}G_{\bm{\theta}^*}(E(\mathbf{x}))^T\cdot\nabla_{\bm{\theta}}G_{\bm{\theta}^*}(E(\mathbf{x}))$ is positive definite, for all $\mathbf{x}\in \text{supp}\big(p_{\text{\text{data}}}\big)$.
        \item $\partial_{\textbf{w}}h(\bm{\psi}^*)\neq 0$ for any $\textbf{w}\notin\mathcal{T}_{\bm{\psi}^*}\mathcal{M}_D$, where $h(\bm{\psi}):=\nabla_{\bm{\theta}} \mathbb{E}_{p_{G_{\bm{\theta}}}(\mathbf{x})}\big[ D_{\bm{\psi}}(\mathbf{x})\big]\Big\vert_{\bm{\theta}=\bm{\theta}^*}$. 
        \item $\mathbf{w}^T\nabla^2_{\mathbf{x}}D_{\bm{\psi}^*}(\mathbf{x})\mathbf{w}\geq0$, for all $\mathbf{w}\notin\mathcal{T}_{\bm{\theta}^*}\mathcal{M}_G$ and $\mathbf{x}\in \text{supp}(p_{\text{\text{data}}})$.

        \begin{remark*}
            Two special cases are either (v-1) $\nabla^2_{\mathbf{x}}D_{\bm{\psi}^*}(\mathbf{x})=0$ for $\mathbf{x}\in\text{supp}(p_{\text{\text{data}}})$, or (v-2) $\mathbf{w}^T\nabla^2_{\mathbf{x}}D_{\bm{\psi}^*}(\mathbf{x})\mathbf{w}>0$, for all $\mathbf{w}\notin\mathcal{T}_{\bm{\theta}^*}\mathcal{M}_G$ and $\mathbf{x}\in \text{supp}(p_{\text{\text{data}}})$.
        \end{remark*}
    \end{enumerate}
\end{assumpC}

\begin{theorem}\label{th:stable_pagoda}
    Suppose that Assumptions~\ref{th:assumption_stable_reduce} and \ref{th:assumption_stable} hold for an equilibrium $(\bm{\theta}^*,\bm{\psi}^*)$ and $\eta>0$ is sufficiently large. Then the alternative gradient descent iteration $\mathcal{F}_h$ described in Section~\ref{subsec:pre_pagoda} is locally convergent on $\mathcal{M}_{G} \times \mathcal{M}_{D}$ for a sufficiently small learning rate $h>0$.    
\end{theorem}

\begin{myproof}{Theorem}{\ref{th:stable_pagoda}} The argument is motivated by \citep{mescheder2018training}.
We notice that $\mathcal{M}_{G} \times \mathcal{M}_{D}$ is a subset of all equilibria of the operators $\mathcal{F}_h$ (or $\mathbf{v}(\bm{\theta}, \bm{\psi})$). This is because that for any $(\bm{\theta},\bm{\psi})\in\mathcal{M}_{G} \times \mathcal{M}_{D}$, we have $p_{\bm{\theta}}=p_{\text{\text{data}}}$, $\mathbf{x}=G_{\bm{\theta}}(E(\mathbf{x}))$, $D_{\bm{\psi}}(\mathbf{x})=0$, and $\nabla_{\mathbf{x}}D_{\bm{\psi}}(\mathbf{x})=\mathbf{0}$ for $\mathbf{x}\in \text{supp}(p_{\text{\text{data}}})$. From Eqs.~\eqref{eq:grad_theta_PaGoDA_GAN} and \eqref{eq:grad_psi_PaGoDA_GAN}, we then can obtain $\nabla_{\bm{\theta}}\mathcal
{L}(\bm{\theta},\bm{\psi})=\nabla_{\bm{\psi}}\mathcal
{L}(\bm{\theta},\bm{\psi})=0$,  meaning $(\bm{\theta},\bm{\psi})$ is an equilibrium.

Now, we show that the alternating gradient descent converges locally on $\mathcal{M}_{G} \times \mathcal{M}_{D}$ by verifying Lemma~\ref{th:jacobian} is fulfilled, and hence, Lemma~\ref{th:stable_mfld} can be applied. Let $(\bm{\theta}^*, \bm{\psi}^*)\in\mathcal{M}_{G} \times \mathcal{M}_{D}$. There is a $\mathscr{C}^1$-diffeomorphism $\Psi$ that transforms a neighborhood of $(\bm{\theta}^*, \bm{\psi}^*)$ onto an open set in $\mathbb{R}^{(N+M)}$ due to Assumption~\ref{th:assumption_stable} (ii). More precisely, we can compute the relation of $\mathcal{F}_h$ and $\mathbf{v}$ after the $\Psi$-reparametrization. Let $\bm{\zeta}:=\Psi(\bm{\theta},\bm{\psi})$, and
\begin{align*}
    \mathcal{F}_h^{\Psi}(\bm{\zeta})&:=\Psi\circ\mathcal{F}_h\circ\Psi^{-1}(\bm{\zeta})\\
    \mathbf{v}^{\Psi}(\bm{\zeta})&:=\Psi'(\bm{\theta},\bm{\psi})\cdot\big(\mathbf{v}\circ \Psi^{-1}(\bm{\zeta})\big).
\end{align*}
Then
\begin{align*}
\nabla_{\zeta}\mathcal{F}_h^{\Psi}(\bm{\zeta}^*)&=\nabla_{\bm{\theta},\bm{\psi}}\Psi(\bm{\theta}^*,\bm{\psi}^*)\cdot\nabla_{\bm{\theta},\bm{\psi}}\mathcal{F}_h(\bm{\theta}^*,\bm{\psi}^*)\cdot\nabla_{\bm{\theta},\bm{\psi}}\Psi(\bm{\theta}^*,\bm{\psi}^*)^{-1}\\
\nabla_{\zeta}\mathbf{v}^{\Psi}(\bm{\zeta}^*)&=\nabla_{\bm{\theta},\bm{\psi}}\Psi(\bm{\theta}^*,\bm{\psi}^*)\cdot\nabla_{\bm{\theta},\bm{\psi}}\mathbf{v}(\bm{\theta}^*,\bm{\psi}^*)\cdot\nabla_{\bm{\theta},\bm{\psi}}\Psi(\bm{\theta}^*,\bm{\psi}^*)^{-1}.
\end{align*}

We remark that similar matrices have identical ranks and spectrum. Therefore, without loss of the generality, we can assume that $(\bm{\theta}^*,\bm{\psi}^*)=(\bm{0}_N, \bm{0}_M)\in\mathbb{R}^{N}\times\mathbb{R}^{M}$, and
\begin{align*}
    \mathcal{M}_G&=\mathcal{T}_{\bm{\theta}^*}\mathcal{M}_G=\{0\}^{N_G}\times\mathbb{R}^{N-N_G}\\
    \mathcal{M}_D&=\mathcal{T}_{\bm{\psi}^*}\mathcal{M}_D=\{0\}^{M_D}\times\mathbb{R}^{M-M_D}.
\end{align*}
We write the new parameterizations as $\bm{\theta}:=(\bm{\upsilon}_G, \bm{\omega}_G) \in \mathbb{R}^{N_G} \times \mathbb{R}^{N-N_G}$ and $\bm{\psi}:=(\bm{\upsilon}_D, \bm{\omega}_D) \in \mathbb{R}^{M_D} \times \mathbb{R}^{M-M_D}$.
For simplicity, we write $\mathbf{v}(\bm{\theta}, \bm{\psi}):=\mathbf{v}(\bm{\upsilon}_G, \bm{\omega}_G, \bm{\upsilon}_D, \bm{\omega}_D)$. To apply Lemma~\ref{th:stable_mfld}, we now aim to show that $\nabla_{(\bm{\upsilon}_G, \bm{\upsilon}_D)}\mathbf{v}(\bm{\theta}^*,\bm{\psi}^*)$ only admits eigenvalues with negative real parts. From Lemma~\ref{th:jacobian},
\begin{align*}
    \nabla_{(\bm{\upsilon}_G, \bm{\upsilon}_D)}\mathbf{v}(\bm{\theta}^*,\bm{\psi}^*)=\begin{bmatrix}
 \hat{K}_{GG}& -\hat{K}_{DG}^T \\
\hat{K}_{DG} & \hat{K}_{DD}
\end{bmatrix}.
\end{align*}
Here, $ \hat{K}_{GG}$, $ \hat{K}_{DG}$, and $ \hat{K}_{DD}$ represent submatrices of $ K_{GG}$, $ K_{DG}$, and $ K_{DD}$, respectively, with coordinates $(\bm{\upsilon}_G, \bm{\upsilon}_D)$, indicating the Jacobian of $\mathbf{v}$ with derivatives taken along the $\bm{\upsilon}_G$ and $\bm{\upsilon}_D$ directions.

First of all, we show that $K_{DD}$ is generally negative semi-definite. Let $\bm{\xi}\in\mathbb{R}^{(N+M)}$ be any vector. Then
\begin{align*}
    \bm{\xi}^TK_{DD}\bm{\xi} =~& 2f''(0) \mathbb{E}_{p_\text{\text{data}}(\mathbf{x})}\big[\bm{\xi}^T\nabla_{\bm{\psi}}D_{\bm{\psi}^*}(\mathbf{x})\cdot\nabla_{\bm{\psi}}D_{\bm{\psi}^*}(\mathbf{x})^T\bm{\xi}\big]\\=~&2f''(0) \mathbb{E}_{p_\text{\text{data}}(\mathbf{x})}\big[\big(\nabla_{\bm{\psi}}D_{\bm{\psi}^*}(\mathbf{x})^T\bm{\xi}\big)^T\cdot\nabla_{\bm{\psi}}D_{\bm{\psi}^*}(\mathbf{x})^T\bm{\xi}\big]\leq0,
\end{align*}
because $f''(0)<0$ from Assumption~\ref{th:assumption_stable} (i). Thus, for any $\hat{\bm{\xi}}_G\in\mathbb{R}^{N_G}$ and $\hat{\bm{\xi}}_D\in\mathbb{R}^{M_D}$ if we consider   $\hat{\bm{\xi}}:=(\hat{\bm{\xi}}_G,\hat{\bm{\xi}}_D)$ in $(\bm{\upsilon}_G,\bm{\upsilon}_D)$-coordinate, 
\begin{align*}
\hat{\bm{\xi}}^T\hat{K}_{DD}\hat{\bm{\xi}}=\bm{\xi}^TK_{DD}\bm{\xi}\leq 0,
\end{align*}
where $\bm{\xi}:=(\hat{\bm{\xi}}_G,\mathbf{0}_{N-N_G},\hat{\bm{\xi}}_D,\mathbf{0}_{M-M_D})\in\mathbb{R}^{(N+M)}$.

Next, we demonstrate that $\hat{K}_{DG}$ is full rank. We observe that $\hat{\bm{\xi}}_D\neq 0$ if and only if $\bm{\xi}\notin\mathcal{T}_{\bm{\psi}^*}\mathcal{M}_D$. Then, according to Assumption~\ref{th:assumption_stable} (iv), we deduce that if $\hat{\bm{\xi}}_D\neq 0$
\begin{align*}
    K_{DG}\bm{\xi}&= -f'(0)\nabla_{\bm{\theta}} \mathbb{E}_{p_{G_{\bm{\theta}^*}}(\mathbf{x})}\big[ \nabla_{\bm{\psi}}D_{\bm{\psi}^*}(\mathbf{x})\cdot\bm{\xi}\big]\Big\vert_{\bm{\theta}=\bm{\theta}^*} =-f'(0)\partial_{\bm{\xi}}h(\bm{\psi}^*)\neq 0.
\end{align*}
The elements of $K_{DG}\bm{\xi}$ corresponding to the $\bm{\upsilon}_D$-coordinates are represented by $\hat{K}_{DG}\hat{\bm{\xi}}_D$, while those corresponding to the $\bm{\omega}_D$-coordinates are $0$. Therefore, we conclude that $\hat{K}_{DG}\hat{\bm{\bm{\xi}}}_D \neq 0$. Consequently, by the rank–nullity theorem, $\hat{K}_{DG}$ is full-rank.

Finally, by using similar arguments by selecting $(\bm{\upsilon}_G,\bm{\upsilon}_D)$-coordinate, without loss of generality, we only need to show $K_{GG}$ is negative definite.
By applying Assumption~\ref{th:assumption_stable} (i) and (v), the following lemma concludes that if $\eta>0$ is sufficiently large, we can conclude the negative definiteness of $\nabla^2_{\bm{\theta}}\mathcal{L}(\bm{\theta}^*, \bm{\psi}^*)$ under Assumption~\ref{th:assumption_stable} (v-2). 
\begin{lemma}
    Let $\mathbf{A}$ be positive definite, and $\mathbf{B}$ be positive semi-definite. Then there is a $\eta_{\text{min}}>0$ so that $-\eta \mathbf{A}+\mathbf{B}$ is negative definite for all $\eta>\eta_{\text{min}}$. 
\end{lemma}
The lemma holds because, for positive (semi-) definite matrix $\mathbf{X}$, we generally have 
\begin{align*}
    \lambda_{\text{max}}(\mathbf{X})\norm{\mathbf{w}}^2\geq\mathbf{w}^T\mathbf{X}\mathbf{w}\geq \lambda_{\text{min}}(\mathbf{X})\norm{\mathbf{w}}^2,
\end{align*}
for all $\mathbf{w}$. Here, $\lambda_{\text{max}}(\mathbf{X})$ and $\lambda_{\text{min}}(\mathbf{X})$ denote the maximum and minimum eigenvalues of $\mathbf{X}$, respectively. Thus if select $\eta>\frac{\lambda_{\text{max}}(\mathbf{B})}{\lambda_{\text{min}}(\mathbf{A})}$, then for any $\mathbf{w}\neq \mathbf{0}$, we have
\begin{align*}
    \mathbf{w}^T(-\eta\mathbf{A}+\mathbf{B})\mathbf{w} = -\eta\mathbf{w}^T\mathbf{A}\mathbf{w}+\mathbf{w}^T\mathbf{B}\mathbf{w}
     \leq \big(-\eta\lambda_{\text{min}}(\mathbf{A}) + \lambda_{\text{max}}(\mathbf{B}) \big)\norm{\mathbf{w}}_2^2<0.
\end{align*}

By applying Lemma~\ref{th:lemma_eigen}, we know that $\nabla_{(\bm{\upsilon}_G, \bm{\upsilon}_D)}\mathbf{v}(\bm{\theta}^*,\bm{\psi}^*)$ only has eigenvalues with negative real parts. Therefore, with a sufficiently small learning rate $h>0$, Lemma~\ref{th:stable_mfld} guarantees the locally convergence of $\mathcal{F}_h$ on $\mathcal{M}_G\times\mathcal{M}_D$.

\end{myproof}

\subsubsection{Literature on Stability Analysis of Adversarial Training}

Studying the stability of GAN training from a dynamical systems perspective has been a popular approach \cite{mescheder2017numerics, nagarajan2017gradient, mescheder2018training, balduzzi2018mechanics, gemp2018global, wang2019solvable, qin2020training}. Generally, proving or disproving whether adversarial training is stable is challenging. However, \cite{mescheder2018training} provides an example (Dirac-GAN) showing that, in general, GANs are not stable unless additional conditions are imposed.

As a result, researchers have explored additional conditions to stabilize GAN training. Essentially, the goal is to impose extra regularizations on the GAN loss $\mathcal{L}_{\text{GAN}}(\bm{\theta},\bm{\psi}):=\mathbb{E}_{p_\text{\text{data}}(\mathbf{x})}\big[f(D_{\bm{\psi}}(\mathbf{x}))\big] + \mathbb{E}_{p_{G_{\bm{\theta}}}(\mathbf{x})}\big[f(-D_{\bm{\psi}}(\mathbf{x}))\big]$,
    or its velocity field $\mathbf{v}_{\text{GAN}}(\bm{\theta}, \bm{\psi}):=\begin{bmatrix}
 -\nabla_{\bm{\theta}} \mathcal{L}_{\text{GAN}}(\bm{\theta}, \bm{\psi})\\
\nabla_{\bm{\psi}} \mathcal{L}_{\text{GAN}}(\bm{\theta}, \bm{\psi})
\end{bmatrix}$
    to ensure that the resulting Jacobian  is Hurwitz. To elaborate further, we revisit the Jacobian $\mathcal{J}_{\text{GAN}}$ of the vanilla GAN, given by $\mathbf{v}_{\text{GAN}}(\bm{\theta}, \bm{\psi})$:    
    \begin{align*}
    \mathcal{J}_{\text{GAN}}(\bm{\theta}, \bm{\psi}):=\begin{bmatrix}
 -\nabla_{\bm{\theta}}^2 \mathcal{L}_{\text{GAN}}(\bm{\theta}, \bm{\psi})& -\nabla_{\bm{\theta},\bm{\psi}}^2 \mathcal{L}_{\text{GAN}}(\bm{\theta}, \bm{\psi}) \\
\nabla_{\bm{\theta},\bm{\psi}}^2 \mathcal{L}_{\text{GAN}}(\bm{\theta}, \bm{\psi}) & \nabla_{\bm{\psi}}^2 \mathcal{L}_{\text{GAN}}(\bm{\theta}, \bm{\psi})
\end{bmatrix}=\begin{bmatrix}
 K_{GG}& -K_{DG}^T \\
K_{DG} & K_{DD}
\end{bmatrix}.
\end{align*}
Here, we slightly abuse the notation from Section~\ref{subsec:pagoda_stable} by using $K_{ij}$, $i,j\in\{D, G\}$, to denote the corresponding components in $\mathcal{J}_{\text{GAN}}$. By similar argument of Lemma~\ref{th:jacobian}, we can  obtain (indeed, $\eta=0$ in Lemma~\ref{th:jacobian}) that
\begin{align*}
    K_{GG}&=  f'(0)\mathbb{E}_{p_\text{\text{prior}}(\mathbf{z})}\big[\nabla_{\bm{\theta}}G_{\bm{\theta}^*}(\mathbf{z})^T\cdot\nabla^2_{\mathbf{x}}D_{\bm{\psi}^*}(G_{\bm{\theta}^*}(\mathbf{z}))\cdot\nabla_{\bm{\theta}}G_{\bm{\theta}^*}(\mathbf{z})\big]. \\
    K_{DG}&= -f'(0)\nabla_{\bm{\theta}} \mathbb{E}_{p_{G_{\bm{\theta}}}(\mathbf{x})}\big[ \nabla_{\bm{\psi}}D_{\bm{\psi}^*}(\mathbf{x})\big]\Big\vert_{\bm{\theta}=\bm{\theta}^*} \\
    K_{DD}&=2f''(0) \mathbb{E}_{p_\text{\text{data}}(\mathbf{x})}\big[\nabla_{\bm{\psi}}D_{\bm{\psi}^*}(\mathbf{x})\cdot\nabla_{\bm{\psi}}D_{\bm{\psi}^*}(\mathbf{x})^T\big].
\end{align*}

Conceptually \citep{asner1970total,bhatia2002stability}, if we can ensure that the Jacobian at some equilibrium has only eigenvalues with strictly negative real parts, then the gradient descent iteration of $\mathcal{L}_{\text{GAN}}$ is asymptotically stable at that equilibrium. Therefore, the objective of many studies \cite{mescheder2017numerics, nagarajan2017gradient, mescheder2018training, qin2020training} is to find conditions to verify Lemma~\ref{th:lemma_eigen}. We focus on discussing the conditions for $K_{GG}$ and $K_{DD}$ to be negative (semi-)definite, as this distinguishes PaGoDA's Theorem~\ref{th:stable_pagoda} from the existing literature. 

Under Assumption~\ref{th:assumption_stable} (i) that $f''(0)<0$, it is worth noting that  $K_{DD}$ is generally negative semi-definite without additional conditions. Hence, studies \cite{mescheder2017numerics, nagarajan2017gradient, mescheder2018training} attempted to impose additional regularizers on $\mathcal{J}_{\text{GAN}}$ or $\mathbf{v}_{\text{GAN}}$ to ensure that either $K_{DD}$ is negative definite (as in \cite{nagarajan2017gradient, mescheder2018training}) or $K_{GG}$
  is negative definite (as in \cite{mescheder2018training}). In Table~\ref{tab:comparison_theory}, we provide a comparison of the various assumptions, at a high-level, drawn from the literature. 
  
  We emphasize that PaGoDA does not require $\nabla^2_{\mathbf{x}}D_{\bm{\psi}^*}(\mathbf{x})$ to be strictly positive definite, thanks to PaGoDA's reconstruction loss. Specifically, it accommodates the scenario where $\nabla^2_{\mathbf{x}}D_{\bm{\psi}^*}(\mathbf{x})= 0 $ on $\text{supp}(p_{\text{\text{data}}})$. It's noteworthy that this capability enables PaGoDA to address cases where the instability of GAN is demonstrated, as exemplified by examples provided by \cite{mescheder2018training}.

    \begin{table}[]
    \caption{Comparison of various assumptions on stability analysis.}\label{tab:comparison_theory}
    \centering
\begin{tabu} to 1\textwidth {X[l]|X[l]|X[l]}
\toprule
           Method & \multicolumn{1}{c|}{$K_{GG}$}                                                       & \multicolumn{1}{c}{$K_{DD}$}                                                                                                                                                                             \\ \midrule
 \cite{nagarajan2017gradient}'s Vanilla  GAN            & Both $p_{\text{\text{data}}}$ and $p_{\bm{\theta}}$ covers the whole space $\mathbb{R}^D$. & Additional technical assumptions (difficult to verify).                                                                                                                                                  \\ \midrule
\cite{mescheder2018training}'s Vanilla  GAN   &     
    \begin{itemize}[leftmargin=*]
        \item $D_{\bm{\psi}^*}(\mathbf{x})=\nabla_{\mathbf{x}}D_{\bm{\psi}^*}(\mathbf{x})=0$ on $\text{supp}(p_{\text{\text{data}}})$. 
        \item $\nabla^2_{\mathbf{x}}D_{\bm{\psi}^*}(\mathbf{x})$ positive definite.
    \end{itemize}
    This implies $K_{GG}$ is negative definite.

& \vspace{1.3cm} No further assumptions.                                                                                                                                                                         \\ \midrule
\cite{mescheder2018training}'s Regularized GAN &                            
    \begin{itemize}[leftmargin=*]  
        \item $D_{\bm{\psi}^*}(\mathbf{x})=\nabla_{\mathbf{x}}D_{\bm{\psi}^*}(\mathbf{x})=0$ on $\text{supp}(p_{\text{\text{data}}})$. 
        \item $\nabla^2_{\mathbf{x}}D_{\bm{\psi}^*}(\mathbf{x})=0$ on $\text{supp}(p_{\text{\text{data}}})$. 
    \end{itemize}
    This simply implies $K_{GG}=0$.

& \vspace{0.08cm} By introducing a regularizer to  modify the vector field $\mathbf{v}$ and obtaining a new vector field $\Tilde{\mathbf{v}}$, they can determine an $L_{DD}$ so that $\Tilde{K}_{DD}:= K_{DD} - L_{DD}$ is negative definite. Therefore, it is not vanilla GAN anymore. \\ \midrule
\rowcolor[HTML]{EFEFEF} 
PaGoDA      &        
        \begin{itemize}[leftmargin=*] 
        \item $D_{\bm{\psi}^*}(\mathbf{x})=\nabla_{\mathbf{x}}D_{\bm{\psi}^*}(\mathbf{x})=0$ on $\text{supp}(p_{\text{\text{data}}})$.
        \item $\nabla^2_{\mathbf{x}}D_{\bm{\psi}^*}(\mathbf{x})$ just need to be positive semi-definite on $\text{supp}(p_{\text{\text{data}}})$.
    \end{itemize} 
    Then with $\eta>0$ chosen to be sufficiently large in PaGoDA, $K_{GG}$ is negative definite. \vspace{0.35cm}
    &  \vspace{1.6cm} No further assumptions.                                                                                                                                                                     \\ \bottomrule
\end{tabu}
\end{table}

\clearpage
\newpage

\section{Limitations and Broader Impacts}\label{app:limitation}
\textbf{Limitations. } Algorithmically, the reconstruction loss is incompatible with the classifier-free guidance, which requires us to adopt the original distillation loss. However, as reconstruction loss directly uses the real data, it provides additional merit to decoder training, resulting in better performance as evidenced in the experiments. Theoretically, some theoretical assumptions of PaGoDA are challenging to verify in practice. For example, Theorems~\ref{th:w_2_convergence} and \ref{th:w_1_convergence_rough} require certain Lipschitz continuity of the score functions. This assumption is difficult to maintain at $t=0$ due to the potential concentration of the data manifold in a lower-dimensional space, causing singularity. However, by truncating the PF-ODE solving at $t=\delta$ (for some $\delta>0$), which is common in practice, this singularity is avoided, making the Lipschitz continuity assumption more feasible. In addition, Theorem~\ref{th:optimality}'s assumption of the existence of a common minimizer can be difficult to verify empirically. However, with proper neural network parametrization and effective optimization, this assumption becomes more feasible. At last, verifying Assumptions~\ref{th:assumption_stable_reduce} and \ref{th:assumption_stable} concerning the optimal properties of the generator and discriminator $(G_{\bm{\theta}},D_{\bm{\psi}})$ is challenging in practice. These assumptions, essential for general (Lyapunov) stability analysis, are difficult to validate empirically. However, they appear reasonable based on our experimental observations. Last, empirically, PaGoDA's T2I generation capability relies heavily on the scale and quality of the training dataset. 


\textbf{Broader Impacts. } PaGoDA, as a general media generative model, carries the risk of producing harmful or inappropriate content, such as deepfake images, graphic violence, or offensive material. To mitigate these risks, we avoid using the LAION dataset~\citep{schuhmann2022laion} in our model training, but robust content filtering and moderation mechanisms are essential to additionally prevent the generation of unethical or harmful media.

\end{document}